\documentclass[12pt]{article}
\usepackage[top=2.5cm, bottom=2.5cm, left=2.5cm, right=2.5cm]{geometry}
\usepackage{authblk}
\usepackage{natbib}
\providecommand{\keywords}[1]{\smallskip\noindent\textbf{Keywords:} #1}

\usepackage{afterpage}
\usepackage{graphicx}
\usepackage{multirow}
\usepackage{amsmath,amssymb,amsfonts}
\usepackage{amsthm}
\usepackage{mathrsfs}
\usepackage[title]{appendix}
\usepackage{xcolor}
\usepackage{textcomp}
\usepackage{manyfoot}
\usepackage{booktabs}
\usepackage{algorithm}
\usepackage{algorithmicx}
\usepackage{algpseudocode}
\usepackage{listings}
\usepackage[style=base]{subcaption}
\usepackage[hypcap=false]{caption}
\usepackage{todonotes}
\usepackage{mdframed}
\usepackage{tabularx}
\usepackage{float}
\usepackage{mdframed}
\usepackage{hyperref}
\newcommand{\R}{\mathbb{R}}

\begin{document}



\date{}
\title{Build on Priors: Vision--Language--Guided Neuro-Symbolic Imitation Learning for Data-Efficient Real-World Robot Manipulation}

\author[1,2]{Pierrick Lorang}
\author[2]{Johannes Huemer}
\author[1]{Timothy Duggan}
\author[2]{Kai Goebel}
\author[2]{Patrik Zips}
\author[1]{Matthias Scheutz}

\affil[1]{Department of Computer Science, Tufts University, Medford, MA, USA \\
  \texttt{\{pierrick.lorang, timothy.duggan, matthias.scheutz\}@tufts.edu}}
\affil[2]{Complex Dynamical Systems, Austrian Institute of Technology GmbH (AIT), Vienna, Austria \\
  \texttt{\{johannes.huemer, kai.goebel, patrik.zips\}@ait.ac.at}}

\maketitle

\abstract{
Enabling robots to learn long-horizon manipulation tasks from a handful of demonstrations remains a central challenge in robotics. Existing neuro-symbolic approaches often rely on hand-crafted symbolic abstractions, semantically labeled trajectories or large demonstration datasets, limiting their scalability and real-world applicability. We present a scalable neuro-symbolic framework that autonomously constructs symbolic planning domains and data-efficient control policies from as few as one to thirty unannotated skill demonstrations, without requiring manual domain engineering.
Our method segments demonstrations into skills and employs a Vision-Language Model (VLM) to classify skills and identify equivalent high-level states, enabling automatic construction of a state-transition graph. This graph is processed by an Answer Set Programming solver to synthesize a PDDL planning domain, which an oracle function exploits to isolate the minimal, task-relevant and target relative observation and action spaces for each skill policy. Policies are learned at the control reference level rather than at the raw actuator signal level, yielding a smoother and less noisy learning target. Known controllers can be leveraged for real-world data augmentation by projecting a single demonstration onto other objects in the scene, simultaneously enriching the graph construction process and the dataset for imitation learning. 
We validate our framework primarily on a real industrial forklift across statistically rigorous manipulation trials, and demonstrate cross-platform generality on a Kinova Gen3 robotic arm across two standard benchmarks.
Our results show that grounding control learning, VLM-driven abstraction, and automated planning synthesis into a unified pipeline constitutes a practical path toward scalable, data-efficient, expert-free and interpretable neuro-symbolic robotics.
Videos showcasing our approach applied on a forklift are available online\footnotemark[1].%
}

\keywords{Neuro-symbolic, Imitation Learning, Task and Motion Planning, Symbolic Planning, Skill Learning, Human-Robot Interaction, Real-World Robotics}
\footnotetext[1]{Video recordings of real-world forklift experiments:
  (1) statistical evaluation of load-navigate-unload cycle under randomized initial poses, trained with 30
  demonstrations (\url{https://youtu.be/syThLRDKRKg});
  (2) system trained to unload pallets onto an elevated truck bed from one single extra demonstration (\url{https://youtu.be/exjpYEDLn3U}).}

\section{Introduction}
Teaching robots to perform complex, long-horizon tasks remains a fundamental challenge at the intersection of robotics, planning, and machine learning. Imitation learning has emerged as a compelling paradigm for acquiring robot behaviors from demonstrations, sidestepping the need for manual reward engineering~\citep{Hussein_Gaber_Elyan_Jayne_2017, Osa_Pajarinen_Neumann_Bagnell_Abbeel_Peters_2018, Zare_Kebria_Khosravi_Nahavandi_2024}. Yet most imitation learning methods operate at the level of individual short-horizon skills, struggle with compounding distribution shifts over long execution horizons~\citep{Rajaraman_Yang_Jiao_Ramachandran_2020, Xu_Li_Yu_2020}, and require large demonstration datasets to generalize reliably.
These limitations become particularly acute in real-world deployments, where tasks span many sequential steps, the number of available demonstrations is small, and the system must generalize to object configurations or task structures not seen during training. Humans address long-horizon problems naturally by abstracting continuous
experience into symbolic structures---predicates, operators, and plans---that support flexible reasoning and reuse~\citep{Newell1976}. Hierarchical approaches such as Task and Motion Planning (TAMP)~\citep{Wolfe_Marthi_Russell_2010, Kaelbling_Lozano-Perez_2011, Garrett_Chitnis_Holladay_Kim_Silver_Kaelbling_Lozano-Perez_2021} mirror this strategy by decomposing problems into a symbolic planning layer and a continuous motion layer. However, they traditionally require hand-engineered symbolic domains, making them brittle and expensive to deploy in new settings.
A natural direction is to learn symbolic models and low-level controllers directly from data. Prior work has made progress on learning symbolic abstractions~\citep{Konidaris_Kaelbling_Lozano-Perez_2018, Bonet_Geffner_2020, kr2021rbrg, Ahmetoglu_Seker_Piater_Oztop_Ugur_2022, Silver_Chitnis_Kumar_McClinton_Lozano-Perez_Kaelbling_Tenenbaum_2022, Shah_Nagpal_Verma_Srivastava_2024, Umili_Antonioni_Riccio_Capobianco_Nardi_Giacomo} or low-level policies~\citep{Illanes_Yan_Icarte_McIlraith_2020, Kokel2021, Guan_Sreedharan_Kambhampati, Silver_Athalye_Tenenbaum_Lozano-Perez_Kaelbling_2023, goel2022rapidlearn, Lorang_Horvath_Kietreiber_Zips_Heitzinger_Scheutz_2024, Lorang_Goel_Shukla_Zips_Scheutz_2024, Lorang_Lu_Scheutz_2025} independently, and some recent methods attempt to learn both jointly~\citep{Silver_Athalye_Tenenbaum_Lozano-Perez_Kaelbling_2023}. This requirement not only demands significant domain expertise from engineers who must anticipate and encode the relevant symbolic vocabulary before deployment, but also renders the resulting systems brittle to novel environments where such prior knowledge is unavailable or incomplete.
A further challenge that has received limited attention is the \emph{data
efficiency} of the full pipeline. Learning reliable controllers and meaningful
symbolic abstractions simultaneously typically demands large demonstration
datasets~\citep{Konidaris_Kaelbling_Lozano-Perez_2018, Chitnis_Silver_Tenenbaum_Lozano-Perez_Kaelbling_2022}.
%
%
This becomes particularly acute in settings that require behavioral flexibility, such as autonomous forklifts operating across varied warehouse layouts or collaborative manipulators adapting to new object configurations, where the closed-world assumptions that make industrial data collection tractable begin to break down, and systems must generalize from only a handful of examples.
We address these challenges by proposing a unified neuro-symbolic framework that learns both a symbolic planning domain and low-level control policies from as few as one to thirty skill demonstrations per skill, without assuming any predefined symbolic vocabulary. Our key insight is that a Vision-Language Model (VLM) can replace the need for complete human annotations: it classifies demonstrated skills and identifies equivalent high-level states by comparing visual scene snapshots, enabling the automatic construction of a state-transition graph over the domain. 

\begin{figure*}[t]
\centering
\includegraphics[width=\textwidth]{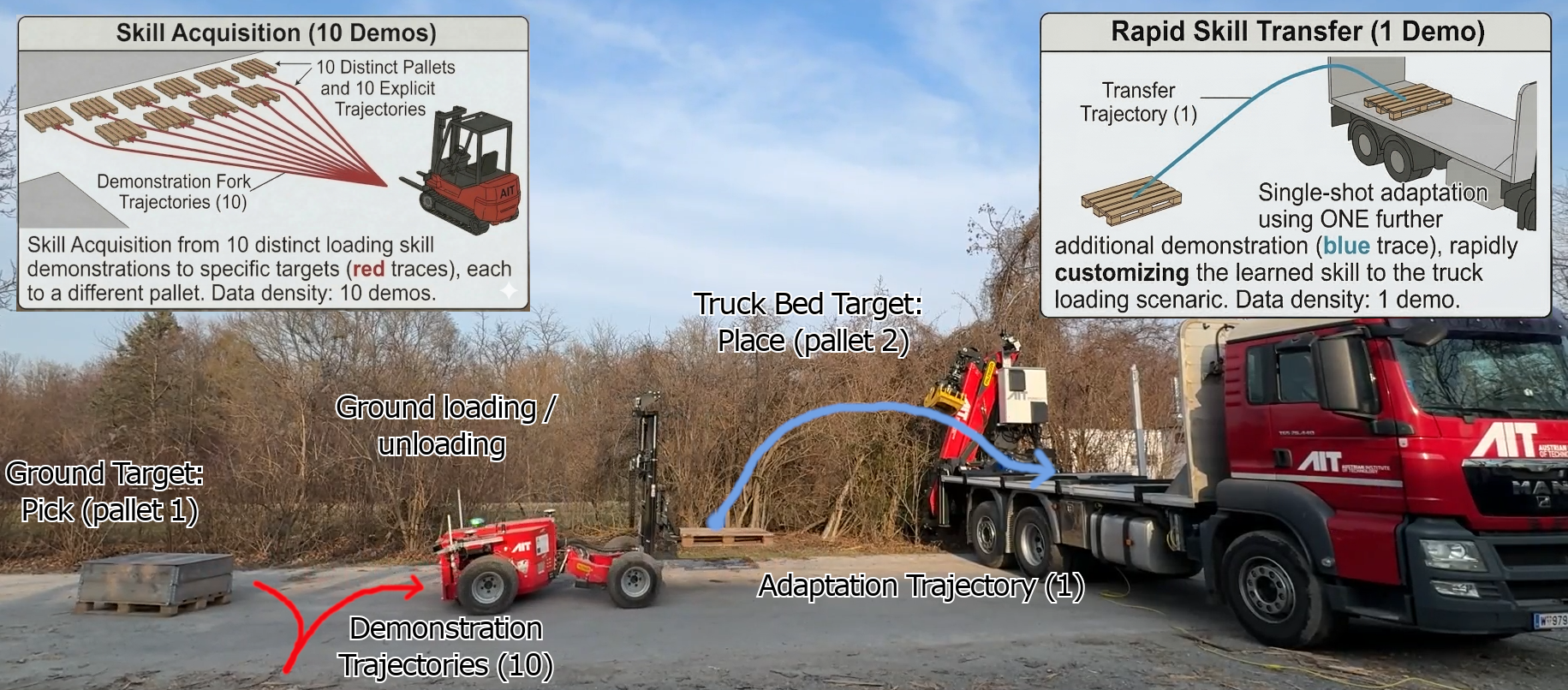}
\small \caption{Our forklift system loading two pallets from the ground and unloading them onto a truck. Such system builds on priors (perception, VLM for annotation and controls) to imitate long-horizon and complex task planning and acting from only 10 base skills demonstrations and 1 extra adaptation skill trajectory.}
\label{fig:forklift}
\end{figure*}
%
%
This graph is passed to an Answer Set Programming (ASP) solver~\citep{Bonet_Geffner_2020, kr2021rbrg} that synthesizes a consistent PDDL planning domain~\citep{mcdermott_pddl_1998}, capturing not only static spatial predicates, as clustering-based methods typically do~\citep{Shah_Nagpal_Verma_Srivastava_2024}, but also richer relational and temporal transitions such as waiting for a process to complete.
The extracted PDDL domain drives an oracle function that automatically identifies, for each planning operator, the minimal relevant observation and action space, inferring the relevant object observation from operator groundings in the training demonstrations as an attention mechanism, and pruning the action space by filtering out dimensions that remain unused across those same demonstrations. The oracle additionally computes relative end-effector pose with respect to the object of interest, providing a compact geometric signal that further reduces learning complexity. Policies are trained at the \emph{control reference level} rather than at the raw actuator signal level, yielding a smoother and less noisy imitation target. Controls notably enable real-world \emph{data augmentation}: given a single demonstration, a skill can be projected onto alternative objects present in the scene, generating additional synthetic demonstrations that simultaneously enrich the graph used for symbolic abstraction and expand the dataset for diffusion-based imitation learning~\citep{Mandlekar_2023_MimicGen}.
We validate our framework on a real industrial forklift system and further demonstrate its generality by deploying the same pipeline on a Kinova Gen3 robotic arm across standard manipulation benchmarks. In both settings, the full pipeline, from raw demonstrations to interpretable long-horizon task execution, requires minimal data, no symbolic supervision, and generalizes robustly to novel configurations.
This paper makes the following contributions:
\begin{itemize}
\item A VLM-driven graph construction pipeline that automates annotation, classifies skills and identifies equivalent high-level states from visual scene snapshots, requiring no predefined predicates, symbols, or manual engineering.
\item An ASP-based symbolic abstraction method that synthesizes a PDDL planning domain from the state-transition graph, supporting rich relational and temporal operator structures.
\item An oracle function that exploits the learned symbolic domain to isolate the minimal observation and action space per skill, including geometry-aware relative pose computation for unary operators.
\item Control-level imitation learning with real-world data augmentation, enabling a single demonstration to be projected onto multiple scene objects to multiply effective dataset size for both graph construction and policy training.
\item Experimental validation on a Kinova Gen3 arm and a real industrial forklift,
demonstrating data-efficient, generalizable, and interpretable long-horizon task
execution from as few as one demonstration per skill.%
\end{itemize}

\section{Related Work}

\subsection{Symbolic Planning and Neuro-Symbolic TAMP}

Classical approaches to Task and Motion Planning (TAMP) decompose long-horizon robot behavior into a symbolic task planner operating over discrete abstractions and a continuous motion planner executing the resulting plan~\citep{garrett2020online, Garrett_Chitnis_Holladay_Kim_Silver_Kaelbling_Lozano-Perez_2021}. A large body of work has studied how to acquire the symbolic components of such architectures from data rather than manual specification.

%
%
One stream of research focuses on learning action models from symbolic traces, either from existing datasets or through environmental interaction~\citep{Chitnis_Silver_Tenenbaum_Lozano-Perez_Kaelbling_2022, Kumar_McClinton_Chitnis_Silver_Lozano-Pérez_Kaelbling_2023, Chitnis_Silver_Tenenbaum_Lozano-Perez_Kaelbling_2022, Konidaris_Kaelbling_Lozano-Perez_2018, Umili_Antonioni_Riccio_Capobianco_Nardi_Giacomo}.
Foundational work by~\citet{Konidaris_Kaelbling_Lozano-Perez_2018} established a principled link between low-level skills and abstract symbolic representations, showing that precondition and effect symbols are both necessary and sufficient for high-level planning. Building on this,~\citet{Chitnis_Silver_Tenenbaum_Lozano-Perez_Kaelbling_2022} introduced Neuro-Symbolic Relational Transition Models (NSRTs), which are data-efficient to learn and generalize over objects, and later extended this line to jointly learn symbolic predicates alongside operators~\citep{Kumar_McClinton_Chitnis_Silver_Lozano-Pérez_Kaelbling_2023}, removing the need for hand-specified state abstractions. 
%
%
A common assumption across all of these works is that symbolic predicates are either given or derived from supervised symbolic traces, as they do not emerge from raw, unlabeled continuous demonstrations.

An orthogonal direction reduces the data requirements for symbolic domain acquisition by framing the problem as combinatorial search.~\citet{Bonet_Geffner_2020} showed that first-order symbolic representations for planning can be inferred from the structure of the state space via a SAT-based search, working from unlabeled state graphs rather than expert-annotated trajectories. This was later extended to Answer Set Programming (ASP) encodings~\citep{kr2021rbrg}, enabling more expressive domain learning with minimal data. Prior work has also applied these solvers to learn continuous features that improve TAMP efficiency from a few example plans with embedded continuous data~\citep{Curtis_Silver_Tenenbaum_Lozano-Pérez_Kaelbling_2022}; however, these approaches assume a predefined symbolic domain and solved plans as input.

%
%
In contrast to methods requiring predefined predicate vocabulary or expert-annotated symbolic traces, our framework requires only a human-provided skill name vocabulary and brief domain hints, automatically constructing the full predicate and operator structure via VLM-driven annotation and ASP synthesis. Unlike prior ASP-based approaches, we co-learn continuous controllers alongside the symbolic domain from a small number of demonstrations.

\subsection{Reinforcement and Imitation Learning for Long-Horizon Tasks}

A parallel line of research addresses long-horizon task execution through learning-based methods without explicit symbolic representations. Imitation learning approaches directly clone behavior from demonstrations: early work on skill sequencing~\citep{Manschitz_Kober_Gienger_Peters_2014, Le_Jiang_Agarwal_Dudik_Yue_Hal_Daume_2018, Pertsch_Lee_Wu_Lim_2021, Tanwani_Yan_Lee_Calinon_Goldberg_2021, Zhu_Stone_Zhu_2022, Teng_Chen_Ai_Zhou_Xuanyuan_Hu_2023} enables the imitation of multi-step behaviors but yields opaque policies that are difficult to repurpose for novel task structures. More recently, generative policy architectures have substantially improved the fidelity and dexterity of learned behaviors: Diffusion Policy~\citep{chi2023diffusionpolicy} frames visuomotor control as a denoising diffusion process and captures complex multimodal action distributions, while Action Chunking with Transformers (ACT)~\citep{Zhao_Kumar_Levine_Finn_2023} predicts short action sequences to reduce compounding errors, achieving high success rates on fine manipulation from as few as fifty demonstrations. These methods were designed for single-skill or short-horizon settings and achieve impressive performance within that scope. Scaling such approaches to long-horizon compositional tasks, however, remains an open challenge that they were not designed to address: reusing sub-behaviors compositionally and adapting to novel task structures without retraining are complementary problems that our framework targets.

Reinforcement learning approaches have similarly been guided by pre-established symbolic domains to structure exploration~\citep{Icarte2020RewardME, sarathy2021spotter, goel2022rapidlearn, peorl-Yang, Gehring_Asai_Chitnis_Silver_Kaelbling_Sohrabi_Katz_2022, Cheng_Xu_2023}. Reward machines~\citep{Icarte2020RewardME} decompose the reward function according to task structure, and symbolic oracles have been used to provide guidance or filter agent actions~\citep{Illanes_Yan_Icarte_McIlraith_2020, Mitchener_Tuckey_Crosby_Russo_2022}. More recent hybrid frameworks~\citep{sarathy2021spotter, goel2022rapidlearn, Lorang_Goel_Shukla_Zips_Scheutz_2024, Lorang_Lu_Scheutz_2025} spawn RL agents to bridge planning gaps when novel situations arise. However, these approaches still rely on manually crafted symbolic representations and spend significant environment interaction time exploring to discover missing operators. Moreover, they evaluate primarily in discrete action domains, or in continuous settings where continuous controllers are assumed to exist rather than co-learned.

Our approach differs fundamentally: we co-learn both the symbolic domain and the continuous controllers from a small set of raw demonstrations, with no manual symbolic engineering and no RL exploration.

\subsection{Learning Symbolic Abstractions from Raw Data}

Extracting symbolic abstractions directly from raw trajectories is a more demanding challenge that has received comparatively less attention. \citet{Shah_Nagpal_Verma_Srivastava_2024} proposed extracting relational symbols from raw demonstrations, but require fifty or more demonstrations to produce a functional planning domain, and the resulting operators are still refined using manually coded controllers. Segmentation methods~\citep{Loula_Allen_Silver_Tenenbaum_2020} can ease this process by identifying sub-task boundaries, but do not resolve the underlying reliance on extensive data.~\citet{Keller_Tanneberg_Peters_2025} cluster continuous observations to induce predicates, also requiring substantial data before stable abstractions emerge.

The approach most closely related to ours is that of~\citet{Lorang_Lu_Huemer_Zips_Scheutz_2025}, which co-learns symbolic domains and low-level controllers from very few demonstrations. Our work builds on this paradigm and extends it with real-world data augmentation as well as automated annotation, demonstrating that the same pipeline, from raw demonstrations to interpretable long-horizon task execution, can be successfully deployed on a real forklift and generalized across manipulation benchmarks while requiring way less human input.

\subsection{Relation to Vision-Language-Action Models}
Vision-Language-Action models
(VLAs)~\citep{brohan2023rt2, black2024pi0, kim2024openvla} represent
a compelling alternative paradigm: by fine-tuning large pretrained
vision-language models on robot trajectory data, they acquire
policies that can be conditioned on natural language instructions and
generalize across diverse tasks and object categories within their
training distribution. Their expressiveness and language
grounding make them attractive for real-world deployment. However,
several fundamental limitations make end-to-end VLAs poorly suited to
the setting we address.

%
%
\textbf{Data requirements.} VLAs are pretrained on datasets comprising hundreds of thousands to millions of demonstration trajectories~\citep{brohan2023rt2, open_x_embodiment_rt_x_2023}, and fine-tuning for new tasks typically requires hundreds of task-specific demonstrations~\citep{kim2024openvla}. While broad pretraining can reduce fine-tuning requirements in well-represented domains, this advantage diminishes for specialized industrial settings such as autonomous forklift operation, which are absent from current VLA pretraining datasets and where our framework's data efficiency is most relevant.
Our framework, by contrast, is
designed explicitly for the few-demonstration regime ($1$--$30$ skill
demonstrations), making it applicable in industrial and specialized
settings where large-scale data collection is impractical.

\textbf{Lack of interpretability and replanning.} VLAs are black-box
function approximators: they produce actions directly from observations
without exposing an intermediate symbolic representation of task
state. Hence, there is no interpretable plan that an operator can
inspect, no natural point at which to inject task-level constraints,
and no mechanism for replanning if execution deviates from the
intended task structure. Our framework produces an explicit PDDL
plan before execution begins, which is human-interpretable and can be verified or overridden prior to deployment.

%

%
\textbf{Compositional generalization.} VLAs generalize by interpolation within their training distribution: they can adapt to new object instances or scene appearances, but composing skills in sequences not represented in training data remains an open challenge they were not designed to address~\citep{black2024pi0}. Symbolic planning, by contrast, supports combinatorial generalization by construction: once a set of operators has been learned, any task expressible as a sequence of those operators can be planned and executed, even if that specific sequence was never demonstrated, contingent on the correctness of the learned PDDL domain, which is the primary technical challenge our framework addresses.

\textbf{Complementarity.} VLMs and VLAs are distinct: VLMs provide semantic understanding and annotation, while VLAs are end-to-end controllers mapping observations to actions. Our framework uses VLMs internally for skill classification, state equivalence judgment, and open-vocabulary detection, and the architectural comparison here is therefore between our neuro-symbolic pipeline as a whole and VLA-based end-to-end control, not between VLMs and VLAs. These approaches are complementary: the diffusion skill policies could be replaced by VLA-style controllers for tasks requiring richer visual conditioning, while retaining the symbolic planning layer for task-level generalization and interpretability. We leave this integration as a direction for future work.
\section{Preliminaries: Neuro-Symbolic Planning-Diffusion Models}

\subsection{Symbolic Planning} Symbolic planning builds upon a formal
domain description $\sigma = \langle \mathcal{E}, \mathcal{F},
\mathcal{S}, \mathcal{O}\rangle$, where $\mathcal{E}$ is a set of
entities, $\mathcal{F}$ a set of boolean or numerical predicates over
entities, $\mathcal{S}$ a set of symbolic states formed by grounded
predicates, and $\mathcal{O}$ a set of operators. Each operator $o \in
\mathcal{O}$ is defined by preconditions $\psi$ and effects $\omega$
over predicates. A grounded operator $\hat{o}$ binds objects to
parameters and can be applied if its preconditions hold, updating the
state according to its effects. A planning task $T=(\mathcal{E},
\mathcal{F}, \mathcal{O}, s_0, s_g)$ seeks a plan
$\mathcal{P}=[o_1,\ldots,o_{|\mathcal{P}|}]$ that transitions from
initial state $s_0$ to goal state $s_g$~\citep{mcdermott_pddl_1998}.

\subsection{Diffusion Imitation Learning (IL)}  Imitation learning (IL)
aims to learn a policy $\pi(\tilde{s})$ from expert demonstrations
$\{(\tilde{s}_t, a_t, \tilde{s}_{t+1})\}_{t=0}^{T}$, where
$\tilde{s}_t$ denotes a continuous state, $a_t$ the expert action, and
$\tilde{s}_{t+1}$ the resulting state. The policy is trained by
minimizing the mean squared error between predicted and expert
actions: $\mathcal{L}(\pi) = \frac{1}{T} \sum_{t=0}^{T} \|
\pi(\tilde{s}_t) - a_t \|^2 .$ Unlike reinforcement learning, IL
bypasses exploration and reward engineering, providing a more
data-efficient way to acquire complex behaviors directly from
demonstrations.

Diffusion policies~\citep{chi2023diffusionpolicy} are imitation
learning methods for continuous control that adapt diffusion models
from generative modeling to policy learning. Expert actions are
perturbed with Gaussian noise during training, and a denoising network
$\epsilon_{\theta}$ is optimized to recover the original actions
conditioned on the state. At inference, the learned reverse process
iteratively refines noisy samples to generate actions, yielding a
stochastic policy $p_{\theta}(a_t \mid s_t)$. Formally, let $\tilde{s}_t \in \mathbb{R}^d$ denote the observation
at timestep $t$ and let $a_t = \Delta p_t \in \mathbb{R}^6$ denote
the end-effector displacement action. A skill policy $\pi_i$ defines
a conditional distribution $p_\theta(a_t \mid \tilde{s}_t)$ over
displacement actions. We model this distribution using a diffusion
policy~\citep{chi2023diffusionpolicy}, which parameterizes the reverse
diffusion process as a denoising network $\epsilon_\theta$ and
generates actions by iteratively refining samples from a Gaussian
prior:
\begin{equation}
\begin{split}
a_t^{(0)} &\sim \mathcal{N}(0, I), \\
a_t^{(k-1)} &= \frac{1}{\sqrt{\alpha_k}}
\left(
a_t^{(k)}
- \frac{1-\alpha_k}{\sqrt{1-\bar{\alpha}_k}}
\epsilon_\theta\!\left(a_t^{(k)}, \tilde{s}_t, k\right)
\right) \\
&\quad + \sqrt{1-\alpha_k}\,\epsilon,
\end{split}
\label{eq:diffusion}
\end{equation}
where $\{\alpha_k\}$ is the noise schedule and $\epsilon \sim
\mathcal{N}(0, I)$. 
%
%
The diffusion formulation is particularly suited to control reference imitation because it captures multi-modal action distributions without mode averaging, a critical property when a skill can be initiated from several valid approach configurations.

Following the options framework~\citep{SUTTON1999181}, each operator
$o_i \in \mathcal{O}$ can be decomposed into $J_i$ sequential
action-step sub-policies $\{\pi_{i,1}, \ldots, \pi_{i,J_i}\}$,
each active until a termination condition $\beta_{i,j} :
\mathcal{H}_t \rightarrow [0,1]$ triggers. The full operator policy
is a learned finite automaton over sub-policies:
\begin{equation}
    \pi_i = \langle \pi_{i,1}, \beta_{i,1},\,
    \pi_{i,2}, \beta_{i,2},\, \ldots,\,
    \pi_{i,J_i}, \beta_{i,J_i} \rangle,
    \label{eq:skill_automaton}
\end{equation}
where execution of $o_i$ terminates when $\beta_{i,J_i}$ triggers,
signaling that the symbolic effects $\omega_i$ have been achieved
in the continuous world.

\subsection{Neuro-symbolic Models} Neuro-symbolic models
combine symbolic reasoning with neural control. A planner solves a
STRIPS task $T = \langle \mathcal{E}, \mathcal{F}, \mathcal{O}, s_0,
s_g \rangle$ to produce a plan
$\mathcal{P}=[o_1,\ldots,o_{|\mathcal{P}|}]$, where each operator
$o_i$ is refined into a neural skill $\pi_i \in \Pi$. Each skill
$\pi_i$ interacts with the environment to realize the operator's
effects $\omega_i$, transitioning the system from a state $s$ to a new
state $s'$. This layered approach enables flexible execution in
continuous spaces while maintaining high-level task abstraction.

\subsection{Control Reference Level Policies}
%
%
A key design choice in hierarchical robot learning is the
\emph{level of abstraction} at which imitation learning is performed.
Raw actuator signals, such as joint torques or motor currents, are noisy,
hardware-specific, and sensitive to the mechanical properties of the
platform. Instead, we operate at the \emph{control reference level}:
policies output desired end-effector displacements $\Delta p \in
\mathbb{R}^6$ in Cartesian space, which are tracked by a low-level
controller (e.g., a Cartesian impedance or velocity controller)
running at higher frequency. This separation between task-space
imitation and joint-space tracking is consistent with classical
robot control theory~\citep{siciliano2009robotics} and yields a
smoother, less noisy imitation target that is transferable across
platforms sharing similar kinematic structure~\footnote{Note that such controls could also be learned separately as primitive-level policies.}. 

\subsection{Foundation Models}
%
Foundation models~\citep{bommasani2021opportunities} are large-scale neural networks pretrained on vast corpora of multimodal data, including images, video, and text, that acquire broad world knowledge and demonstrate strong zero-shot generalization across a wide variety of downstream tasks. In robotics, two families of foundation models are of particular relevance to our framework.
%

%
\textbf{Vision-Language Models (VLMs).} A VLM is parameterized as $f_\theta : (\mathcal{I}, \mathcal{Q}) \rightarrow \mathcal{R}$, where $\mathcal{I}$ denotes an image (or set of images), $\mathcal{Q}$ a natural-language query, and $\mathcal{R}$ a natural-language (or structured) response. Through pretraining on internet-scale image--text pairs, VLMs encode rich semantic knowledge about object categories, spatial relations, and action concepts. In the robotics context, VLMs have been used for task planning from human demonstrations~\citep{wake2023gpt4v}, labeling of demonstration datasets~\citep{Shridhar_2022_DIAL}, and zero-shot skill classification~\citep{ViLa_2023}. These models are particularly attractive for our framework because they require no task-specific fine-tuning for scene understanding: given a pair of pre- and post-skill images and a human-provided vocabulary of skill labels, a VLM can identify which skill was executed in most cases, substantially reducing the need for manual annotation. For ambiguous cases where the VLM confidence falls below a threshold, a human fallback mechanism ensures label correctness.
%

%
\textbf{Open-Vocabulary Object Detectors.} Object detection models such as OWLv2~\citep{Minderer_Gritsenko_Houlsby_2024} are distinct from VLMs: rather than producing a global response to a query, they localize object instances as bounding boxes given a text prompt, without retraining on new categories. Formally, OWLv2 computes $\text{OWL}(I, p) = \{(b_i, s_i)\}$, where $I$ is an image, $p$ a text prompt, $b_i$ a predicted bounding box, and $s_i$ the associated confidence score. In our framework, three distinct vision models serve separate purposes: a VLM for skill annotation and state equivalence judgment, OWLv2 for open-vocabulary object detection, and on platforms requiring embedded inference, a compact YOLOv8~\citep{yolov8_ultralytics} student distilled from OWLv2. 

\section{Problem Definition}
\label{sec:prob_def}

\subsection{Setting and Assumptions}
We consider a robot operating in a semi-structured environment
containing a set of manipulable objects $\mathcal{E} = \{e_1, \ldots,
e_N\}$. The robot observes the world through a continuous observation
space $\tilde{\mathcal{S}} \subseteq \mathbb{R}^d$, comprising RGB
images from cameras and proprioceptive readings, and acts through a continuous action space $\mathcal{A} \subseteq \mathbb{R}^m$ of end-effector displacement commands. The
true underlying world state admits a symbolic abstraction: there
exists a discrete state space $\mathcal{S}$ defined over a set of
boolean predicates $\mathcal{F}$ grounded on $\mathcal{E}$, and a
set of operators $\mathcal{O}$ whose application transitions the world
between symbolic states. Neither $\mathcal{F}$ nor $\mathcal{O}$ are
known a priori; they must be inferred from data.

\subsection{Demonstration Data}
%
%
The learner is provided with a small demonstration dataset
$\mathcal{D} = \{\tau^{(1)}, \ldots, \tau^{(n)}\}$, where each
trajectory $\tau^{(k)} = \{(\tilde{s}_t, a_t)\}_{t=0}^{T_k}$ is a
sequence of continuous observation-action pairs generated either by a
human operator or by automated data augmentation (Section~\ref{sec:augmentation}),
performing a single skill.
The total number of demonstrations
per skill is assumed to be small: $|\mathcal{D}_{o_i}| \in [1, 30]$
for each operator $o_i$. No symbolic annotation is provided alongside
the demonstrations: there are no predicate labels, no state
segmentation markers, and no specification of planning operators.
The only input is a semantic human-provided vocabulary of skill names
$\Lambda = \{l_1, \ldots, l_K\}$, consistent with the level of supervision that would be required to finetune a VLA model. These labels are not assigned to trajectories.

\begin{figure*}[t]
    \centering
    \includegraphics[width=\textwidth]{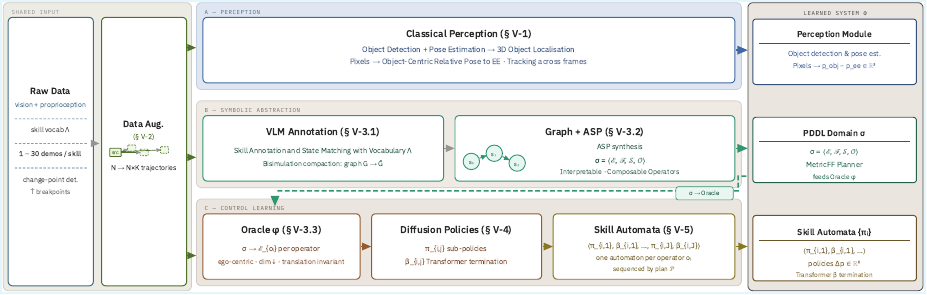}
    \caption{
\textbf{Training Pipeline.}
Two shared input columns feed three processing lanes, each producing one component of the learned system $\Phi$ (right).
\textbf{Shared input:}
Raw demonstrations (vision, proprioception, vocabulary $\Lambda$, change-point breakpoints $\hat{T}$) are augmented by projecting waypoint skeletons onto all scene objects ($N \rightarrow N \times K$ trajectories). Augmented data feeds all three lanes.
\textbf{Lane A (object detection):}
OWLv2 distills a compact YOLOv8 student online. Detections are lifted to 3D and combined with proprioception to form ego-centric trajectories $\tau = \{ p_{\text{obj}} - p_{\text{ee}}, a_t \}$, tracked across time. Delivers the \emph{YOLOv8 Detector}.
\textbf{Lane B (symbolic abstraction):}
A VLM classifies each skill segment by comparing before/after scene images against $\Lambda$, building graph $G$ that an ASP solver converts into PDDL domain $\sigma = \langle \mathcal{E}, \mathcal{F}, \mathcal{S}, \mathcal{O} \rangle$. Delivers \emph{PDDL Domain $\sigma$}, and feeds $\sigma$ back (dashed) to Lane C.
\textbf{Lane C (control learning—sequential after Lane B):}
The Oracle reads $\sigma$ to construct minimal, ego-centric observations $\tilde{s}^{o_i}_t$ per operator. Diffusion sub-policies with Transformer termination predictors $\beta_{i,j}$ are assembled into per-operator skill automata. Delivers \emph{Skill Automata $\{\pi_i\}$}.
}
\label{fig:overview}
\end{figure*}

\subsection{Long-Horizon Task Execution}
At execution time, the user specifies a task as a pair $(s_0, s_g)$, where
$s_0 \in \mathcal{S}$ is the initial symbolic state and $s_g \in
\mathcal{S}$ is the desired goal state. Task specification can be done visually, by selecting snapshots mapped to $(s_0, s_g)$. The system must produce a
closed-loop execution that drives the robot from $s_0$ to $s_g$ by
sequencing learned skill policies. 
Success requires both correct
high-level planning---selecting the right operator sequence---and
reliable low-level execution of each skill under real-world perceptual noise.

\subsection{Objective}
The goal is to learn, from $\mathcal{D}$ and vocabulary $\Lambda$
alone, a system $\Phi = (\sigma, \{\pi_i\}_{i=1}^{|\mathcal{O}|})$
consisting of a symbolic planning domain $\sigma$ and a set of neural
skill policies, such that $\Phi$ solves long-horizon tasks $(s_0,
s_g)$ at execution time with high success rate, generalizing across all
three levels defined above, and without any further human supervision
or domain engineering. Another central objective of our framework is generalization beyond the
specific object configurations seen during training. We distinguish
three levels of generalization, each handled by a distinct component
of our pipeline: \textbf{Intra-task generalization} refers to the ability to execute a
known operator sequence when objects appear at positions not seen
during training.
\textbf{Inter-task generalization} refers to the ability to solve
tasks with goal states $s_g$ that require operator sequences not
present in any single demonstration.
\textbf{Cross-object generalization} refers to the ability to apply a
known skill to object instances not present during training.
\section{Neuro-symbolic Model}

%
%
We utilize prior work demonstrating that neuro-symbolic models can be
learned from few
demonstrations~\citep{Lorang_Lu_Huemer_Zips_Scheutz_2025} to construct our neuro-symbolic model (NSM) which combines PDDL-based symbolic
planning with low-level controls. Figure~\ref{fig:overview} provides an overview of this approach. When performing a task, the model
first detects the object position, then generates a high-level plan using the symbolic planner, and
executes each step by using a trained diffusion policy. All diffusion policies operate on relative position observations, specifically object position expressed
with respect to the end effector, and output end-effector displacement
actions. The framework only receives images and proprioceptive information as input during execution. Both the
high-level planning domain and the low-level controls are learned
directly from demonstrations. Like VLAs, the NSM learns to detect,
plan, and act entirely from observational data,
without task-specific hand-crafted controllers or manually specified symbolic domains.

\subsection{Perception}
\subsubsection{From Pixels to Pose}
Prior work~\citep{Lorang_Lu_Huemer_Zips_Scheutz_2025} used object-based
input to the framework, leaving a ``perception gap'' to be
addressed. The proposed framework requires any perception stack to expose a
single, platform-agnostic interface for object interaction: a 6-DoF pose estimate per tracked
entity expressed in a frame relative to the sensor. This interface can be realized
by any combination of sensors and algorithms appropriate for the
deployment platform. As a concrete instantiation, we reuse the
perception modules from the ADAPT system~\citep{Huemer2025ADAPT}, which
fuses stereo-camera detections, wheel-encoder odometry, RTK-GNSS, and
LiDAR measurements across three tightly coupled stages described below.

\paragraph{Pallet detection and 6-DoF pose estimation}
A convolutional encoder-decoder network~\citep{Beleznai_2025_PalletDetection} is trained entirely on
synthetic stereo image pairs to detect pallets via a part-constellation
model: keypoints and edge segments vote probabilistically for pallet
centers, and a depth-informed PnP solver lifts the 2D detections into
a 6-DoF camera-frame pose estimate
$\hat{T}^{\text{cam}}_e = (\hat{R}_e, \hat{w}_e) \in SE(3)$
for each pallet instance $e$.

\paragraph{Joint localization and pallet mapping}
Raw detections $\hat{T}^{\text{cam}}_e$ are fused with wheel-encoder
odometry and RTK-GNSS measurements in an iSAM2 factor graph that
jointly optimizes the vehicle pose sequence $\{X_t\}$ and the global
pallet poses $\{P_e\}$ in real time. Pallet detections enter the
graph as binary pose factors between $X_t$ and $P_e$, with diagonal
Gaussian covariance whose translational and rotational uncertainties
grow linearly with sensor range to reflect the distance-dependent
degradation of stereo accuracy. 
Mahalanobis-distance data association~\citep{Mahalanobis_2018_Reprint} links detections across views, while pallet
bookkeeping maintains object permanence when pallets leave the sensor
frustum, yielding a centimeter-accurate, incrementally updated map of
all pallet poses in the mission area.

%
%
\paragraph{Truck loading edge detection}
Prior to each unloading cycle, the target truck platform is
localized once from LiDAR point clouds aggregated over time.
After height filtering and edge-candidate classification, specifically points
whose local neighbourhood contains both a vertical and a
ground-parallel normal direction, a RANSAC line fit extracts
the loading edge and its front face, defining a reference frame
$T^{\text{truck}} \in SE(3)$ from which configurable pallet slot
targets $\{T^{\text{slot}}_k\}$ are derived for downstream
placement planning.

\subsubsection{Mapping Detections to Planning Objects}
\label{sec:mapping}

To establish the mapping between tracking IDs and symbolic entities,
we compute relational features (e.g., spatial proximity, relative
positions) for both detected objects and ground-truth entities
$\mathcal{E}$, then solve a correspondence problem that maximizes
relational similarity. For instance, if the symbolic state contains
``$\text{on}(\text{cube1}, \text{cube2}) \wedge
\text{left-of}(\text{cube3}, \text{cube1})$'' and the tracker detects
three objects with IDs $\{\text{blue\_cube\_0}$,
$\text{red\_cube\_0}$, $\text{green\_cube\_0}\}$, we compute the same
relational predicates for each detection set and match the
configuration that exhibits the highest structural similarity,
producing the mapping $\rho(\text{blue\_cube\_0})$ $=\text{cube1}$,
$\rho(\text{red\_cube\_0})=\text{cube2}$,
$\rho(\text{green\_cube\_0})=\text{cube3}$. This bijection $\rho:
\mathcal{T}_t \to \mathcal{E}$ grounds symbolic references in the
visual stream and resolves scene ambiguities that arise when visually
similar objects occupy symmetric configurations.  I.e., without
relational grounding, two red cubes in a mirrored arrangement, for
example, would be indistinguishable to appearance-based tracking
alone.  By leveraging the symbolic state $s_t$ and its predicates, the
system can determine which physical object corresponds to which
symbolic entity based on their roles in the task. This ensures that
when the planner generates an action $o_i(e_j)$ for a specific entity
$e_j \in \mathcal{E}$, the Oracle function $\phi$
(Section~\ref{sec:oracle}) will select the correct corresponding
physical object from $\mathcal{T}_t$ via $\rho^{-1}(e_j)$, allowing
the low-level policy $\pi_i$ to execute the manipulation on the
intended target.

\subsection{Real-World Data Augmentation using Controls}
\label{sec:augmentation}
A central bottleneck of data-efficient imitation learning is that a
single demonstration covers only one object configuration: the
trajectory from start state $\tilde{s}_0$ to goal state $\tilde{s}_g$
is entangled with the specific object poses encountered in that
episode, and does not generalize to novel arrangements without
additional data. For manipulation skills, where the end-effector
trajectory is tightly coupled to object pose, control theory offers a
principled remedy: given a target waypoint or reference trajectory, a
controller can exploit a precise model of the physical embodiment, its
kinematic and dynamic constraints, and its sensor characteristics to
drive the system toward that target reliably across varying initial
conditions. The gap between the two is therefore not one of
capability, but of interface: imitation learning excels at extracting
{\em what} to do from raw experience but remains brittle to perceptual
variation, while classical control is highly competent at {\em how} to
reach a specified target but requires that target to be symbolically
and spatially well-defined. Bridging this gap---translating
demonstrated intent into controller-ready references that remain valid
across object configurations---is precisely the challenge this section
addresses.

To multiply the effective dataset size without additional human
effort, we exploit the structure of the learned control system to
project a single recorded demonstration onto other objects present in
the scene.

\subsubsection{Trajectory projection onto new objects}

For a manipulation skill, let the recorded demonstration consist of $K$ waypoints
$\mathcal{W}^{\text{src}} = \{T_k^{\text{src}}\}_{k=1}^{K} \subset SE(3)$,
where each $T_k^{\text{src}} = (R_k^{\text{src}},\, w_k^{\text{src}})$
encodes the end-effector orientation $R_k^{\text{src}} \in SO(3)$ and
position $w_k^{\text{src}} \in \mathbb{R}^3$ at step $k$.
Given the source ($src$) and target ($tgt$) object poses at the start and end of the
manipulation segment,
$T^{\text{src}}_{\text{start}},\, T^{\text{src}}_{\text{end}},\,
 T^{\text{tgt}}_{\text{start}},\, T^{\text{tgt}}_{\text{end}} \in SE(3)$,
the trajectory is first rigidly re-based to the target object frame via
\begin{equation}
    \begin{aligned}
    T_{\text{align}} = T^{\text{tgt}}_{\text{start}}\bigl(T^{\text{src}}_{\text{start}}\bigr)^{-1}, \\
    T_k^{\text{align}} = T_{\text{align}}\cdot T_k^{\text{src}}, \quad k=1,\dots,K.
    \label{eq:rebase}
    \end{aligned}
\end{equation}
This re-basing aligns the trajectory origin to the target configuration,
but provides no guarantee that the projected end pose
$T_K^{\text{align}}$ coincides with $T^{\text{tgt}}_{\text{end}}$,
since the relative transformation between start and end may differ
between source and target configurations.
The residual $SE(3)$ error is computed as
\begin{equation}
    T^{\text{res}} = T^{\text{tgt}}_{\text{end}}\bigl(T_K^{\text{align}}\bigr)^{-1},
    \qquad T^{\text{res}} = (R^{\text{res}},\, p^{\text{res}}),
    \label{eq:residual}
\end{equation}
where $p^{\text{res}} \in \mathbb{R}^3$ and $R^{\text{res}} \in SO(3)$
are its translational and rotational components, respectively.
To correct the trajectory smoothly, the residual is distributed along
the waypoint sequence using the normalized arc-length parameter
$\alpha_k \in [0,1]$, defined as
$\alpha_k = \ell_k / \ell_K$ where
$\ell_k = \sum_{j=2}^{k}\|w_j^{\text{align}} - w_{j-1}^{\text{align}}\|$
is the cumulative Euclidean path length up to waypoint $k$.
The augmented waypoints are then
\begin{equation}
  \begin{aligned}
    w_k^{\text{aug}} &= w_k^{\text{align}} + \alpha_k\, p^{\text{res}}, \\
    R_k^{\text{aug}} &= \mathrm{Slerp}(\alpha_k;\,I,\,R^{\text{res}})\cdot R_k^{\text{align}}.
  \end{aligned}
  \label{eq:augment}
\end{equation}
where $\mathrm{Slerp}(\alpha;\,I,\,\R^{\text{res}})$ denotes spherical linear
interpolation in $SO(3)$ from the identity $I$ to $\R^{\text{res}}$,
ensuring a geodesic, minimum-torque correction of orientation.
The augmented waypoint set
$\mathcal{W}^{\text{aug}} = \{(R_k^{\text{aug}},\,w_k^{\text{aug}})\}_{k=1}^{K}$
is finally passed to the motion planner, which replans a collision-free
trajectory through these references.
In the spirit of MimicGen~\citep{Mandlekar_2023_MimicGen}, this procedure
adapts a single demonstration to new scene configurations without
additional human effort; unlike MimicGen, however, our method operates
entirely at the level of abstracted control references and is deployed
directly on a real robot without simulation. In the limiting case $K=2$,
the projection reduces to remapping the start and goal poses to the target
configuration, with the motion planner responsible for generating the
connecting trajectory, as used for navigation-based skills.

\subsubsection{Dual benefit for graph construction and policy learning} The augmented trajectories serve two complementary roles in the overall pipeline. First, each augmented trajectory contributes an additional edge instance to the state-transition graph $G$: after VLM annotation (Section~\ref{sec:annotation}), it increases the statistical support per edge type, enabling the ASP solver to infer a more consistent and compact PDDL domain. Second, the augmented demonstrations provide additional training pairs $(\tilde{s}_t, a_t)$ for the diffusion policy of operator $o_i$, directly expanding the effective dataset size.

Despite the augmented trajectories being generated by the motion planner rather than a human, training a diffusion policy on top of them remains essential. The diffusion model learns a richer multi-modal action distribution by combining planner-generated and human-demonstrated trajectories, generalizing more robustly to positional perturbations at execution time~\citep{chi2023diffusionpolicy}. Furthermore, the diffusion policy operates in closed-loop over relative observations at inference time, whereas the planner is open-loop with respect to the object detector; the imitation learning layer therefore absorbs residual localization errors and environmental perturbations that the planner cannot anticipate.

\subsection{Automated Symbolic Abstractions for Reasoning.}
\subsubsection{Annotating Trajectories with Foundation Models}
\label{sec:annotation}
Each raw demonstration trajectory $\tau \in \mathcal{D}$ represents a continuous segment of robot execution during which a single skill is performed. To build the state-transition graph $G = \langle V, E, L \rangle$ required by the ASP-based symbolic abstraction stage, we must (i) identify the high-level state $n$ before and $n'$ after each skill execution, and (ii) assign a skill label $l$ to each directed edge $(n, l, n')$. Our approach replaces expert symbolic annotation with a VLM-driven pipeline, requiring only a small human-provided vocabulary $\Lambda = \{l_1, \ldots, l_K\}$ of skill names, e.g., \texttt{unload pallet}, \texttt{load pallet} and \texttt{navigate} for the forklift domain, which are at the same level of supervision as a prompt given to a VLA to describe a task.

\paragraph{State representation} Each node $n \in V$ corresponds to a visual scene snapshot, concretely an image $I$ captured at the transition between two consecutive skill executions. Nodes are identified up to bisimulation equivalence: two snapshots $n$ and $n'$ are merged if a VLM judges the depicted scene configurations to be semantically equivalent given short domain specific hints (e.g., ``The definition of a state is primarily set by the number of pallets in the scene, the agent location and whether a pallet is currently loaded''). Formally, we query the VLM with a prompt of the form:
\begin{quote}
\emph{``You are an expert forklift operator assessing whether two snapshots represent
the SAME high-level world state.
You are given: Image A – the snapshot for node A  (video: \{node\_a.video\_id\}). Image B – the snapshot for node B  (video: \{node\_b.video\_id\})
Focus on: \{hints\}, ignore everything else.
Minor pixel-level differences are acceptable — judge at the TASK level.''}
\end{quote}
together with the image pair. If the VLM returns a positive equivalence judgment, the corresponding graph nodes are merged into a single abstract state. This bisimulation compaction~\citep{Lorang_Lu_Huemer_Zips_Scheutz_2025} reduces the number of distinct nodes in $G$ and increases the number of skill instances available per graph edge, improving the quality of the PDDL domain subsequently inferred by the ASP solver.

\paragraph{Skill classification} To label each graph edge, we extract uniformly distributed frames from the demonstration segment $\tau$, and issue a classification prompt to the VLM:
\begin{quote}
\emph{``You are an expert forklift operator analyzing a short manipulation sequence.
You are given: 10 video frames sampled uniformly across the clip. The corresponding control trajectory.
Trajectory summary:
\{action\_trajectory\}
TASK:
1. Identify the single dominant manipulation skill performed across the entire sequence.
2. Choose the most precise label possible among the established labels: $\{l_1, \ldots, l_K\}$? Respond with the skill name only.''}
\end{quote}

The VLM responds with a label $\hat{l} \in \Lambda$, which is assigned
to the graph edge $(n, \hat{l}, n')$. This approach is inspired by
recent work on VLM-based skill recognition and task planning from
demonstration videos~\citep{wake2023gpt4v, SeeDo_2024}, and eliminates
the manual trajectory labeling that most prior neuro-symbolic
approaches implicitly require. In addition to such prompts, we query
the VLM to output a confidence metric $c(\tau)$. When $c(\tau) <
\tau_c$, the trajectory is flagged and routed to a human
annotator. This mechanism decouples the scalability of the pipeline
from VLM reliability: in easy, visually distinct tasks the system
operates fully autonomously, while in ambiguous cases it requests the
minimal human intervention necessary to preserve label correctness. In
practice, we find that the fraction of flagged trajectories is small,
confirming that VLMs provide reliable zero-shot skill classification
when pre- and post-skill images exhibit clear visual differences.

\subsubsection{Symbolic Abstraction}
From raw demonstration trajectories $\mathcal{D}$, we extract node
transitions $\tau^{node} = (n, l, n')$, where $n$ and $n'$ are
high-level states and $l$ is the VLM-assigned label, for instance
\texttt{stack}. The nodes all correspond to higher-level states where
$n$ and $n'$ correspond to the beginning and the end of the skill
demonstration, respectively. These transitions form a graph $G=\langle
V,E,L\rangle$ whose nodes represent abstract states and edges
represent skills.  To compact the structure, we compute a minimal
bisimulation $\bar{G}$, removing redundant states while preserving
equivalence. Using an ASP-based solver~\citep{Bonet_Geffner_2020,
  kr2021rbrg}, we infer a symbolic domain $\sigma = \langle
\mathcal{E}, \mathcal{F}, \mathcal{S}, \mathcal{O}\rangle$ in PDDL
form.  This yields a symbolic abstraction of the demonstrations
suitable for classical planning, as well as a complete description of
every node (state) in the graph.

\subsubsection{Oracle-Guided Observation Filtering}
\label{sec:oracle}
A key mechanism enabling sample efficiency in our framework is the
\emph{Oracle} function $\phi$ which filters the full scene observation
down to only the information relevant to the operator currently being
executed. Formally, given the PDDL domain $\sigma = \langle
\mathcal{E}, \mathcal{F}, \mathcal{S}, \mathcal{O} \rangle$ and the
symbolic plan $\mathcal{P} = [o_1, \ldots, o_{|\mathcal{P}|}]$, the
Oracle extracts from each operator $o_i$ the minimal entity set:
\begin{equation}
    \mathcal{E}_{o_i} = \bigl\{ e \in \mathcal{E} \;\big|\;
    e \text{ appears in } \psi_{o_i} \cup \omega_{o_i} \bigr\},
    \label{eq:oracle_entities}
\end{equation}
where $\psi_{o_i}$ and $\omega_{o_i}$ are the preconditions and
effects of $o_i$, respectively. At execution timestep $t$, the tracking
system provides 3D position estimates $\mathcal{P}_t$ for all detected
objects. The Oracle resolves symbolic entities to physical detections
via the bijection $\rho : \mathcal{T}_t \rightarrow \mathcal{E}$ and
constructs a filtered, ego-centric observation:
\begin{equation}
    \tilde{s}^{o_i}_t = \Bigl\{\, p^{\text{rel}}_e =
    p_{\rho^{-1}(e)} - p_{\text{ee}} \;\Big|\;
    e \in \mathcal{E}_{o_i} \Bigr\},
    \label{eq:oracle_obs}
\end{equation}
where $p_{\text{ee}} \in \mathbb{R}^3$ is the current end-effector
position. For unary operators (i.e., $|\mathcal{E}_{o_i}| = 1$), the
Oracle additionally computes the full relative 6-DoF transform between
the end effector and the target object, providing a compact geometric
signal that encodes both position and orientation offsets. The
observation $\tilde{s}^{o_i}_t$ is then passed as input to the
corresponding diffusion policy $\pi_i$.

This design has two important consequences for generalization. First,
expressing object positions relative to the end effector makes the
observation space translation-invariant: a policy trained on one
absolute scene configuration transfers directly to new configurations
as long as the relative geometry is similar, which is guaranteed when
operating on the filtered entity set $\mathcal{E}_{o_i}$. Second,
the dimensionality of $\tilde{s}^{o_i}_t$ scales with $|\mathcal{E}_{o_i}|$
rather than with the total number of scene objects $|\mathcal{E}|$,
dramatically reducing the input space and the amount of demonstration
data required to train a reliable policy for each operator.

\begin{figure*}[t]
    \centering
    \includegraphics[width=\textwidth]{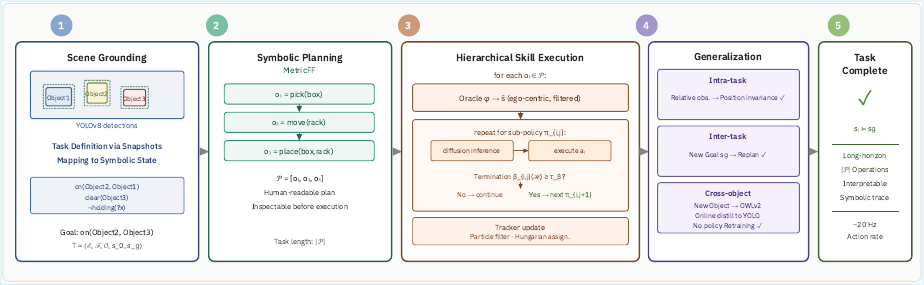}
    \caption{
    {\em Execution Pipeline.}
    At execution time, the system proceeds through five stages.
    \textbf{(1) Scene grounding:}
    YOLOv8 detects task-relevant objects, 3D positions are estimated, and relational predicates are evaluated to construct the initial symbolic state $s_0$. The user specifies a goal $s_g$ as a partial state.
    \textbf{(2) Symbolic planning:}
    MetricFF solves the PDDL instance $\mathcal{T} = \langle \mathcal{E}, \mathcal{F}, \mathcal{O}, s_0, s_g \rangle$ in milliseconds, producing a human-readable, inspectable plan $\mathcal{P}$.
    \textbf{(3) Hierarchical skill execution:}
    For each operator $o_i$, the Oracle $\phi$ resolves symbolic entity bindings to physical detections and constructs the ego-centric observation. Diffusion sub-policies execute closed-loop at $\sim$20\,Hz; a Transformer termination predictor $\beta_{i,j}$ gates transitions between action steps. The tracker updates the symbolic state after each operator completes.
    \textbf{(4) Generalization:}
    Three levels of generalization are handled by distinct mechanisms—relative observations for intra-task generalization, replanning for inter-task generalization, and on-the-fly OWLv2 distillation for cross-object transfer.
    \textbf{(5) Termination:}
    Execution terminates when $s_t \models s_g$.
    }
    \label{fig:execution}
\end{figure*}

\subsection{Neural Plasticity for Skill Learning}

\subsubsection{Training Controls}
Given the symbolic domain $\sigma$ and the demonstration dataset
$\mathcal{D}$, we instantiate and train the operator policy automaton
of Equation~\ref{eq:skill_automaton} for each operator $o_i \in
\mathcal{O}$ as follows.

\paragraph{Sub-policy decomposition} The change-point breakpoints
$\hat{\mathcal{T}}$ extracted from each operator demonstration segment the trajectory into $J_i$ phases corresponding to behaviorally distinct
stages of the manipulation (e.g., approach, contact, retraction).
Each phase defines the training data for one sub-policy $\pi_{i,j}$:
the demonstration segments between consecutive breakpoints
$[t_{j-1}, t_j]$ provide the observation-action pairs
$\{(\tilde{s}^{o_i}_t, a_t)\}$ used for diffusion policy training. The same change-point breakpoints
$\hat{\mathcal{T}} = \{t_1, \ldots, t_K\}$ additionally provide the
supervision signal for termination learning. For each sub-policy
$\pi_{i,j}$, the positive class (termination) is defined by a window
of $\delta$ timesteps centered on each detected breakpoint $t_k$,
and the negative class by all remaining timesteps. This creates a
natural and annotation-free labeling of termination events directly
from the motion signal, with no additional human supervision required.


\paragraph{Diffusion policy training} Each sub-policy $\pi_{i,j}$ is
trained by minimizing the standard diffusion denoising objective over
its assigned demonstration segments:
\begin{equation}
    \mathcal{L}_{\pi} = \mathbb{E}_{({\tilde{s}}_t, a_t),\, k,\,
    \epsilon}\!\left[\left\|\epsilon -
    \epsilon_\theta\!\left(\sqrt{\bar{\alpha}_k}\, a_t +
    \sqrt{1-\bar{\alpha}_k}\,\epsilon,\;
    \tilde{s}^{o_i}_t,\; k\right)\right\|^2\right],
    \label{eq:diffusion_loss}
\end{equation}
where $k \sim \mathcal{U}(1, K)$ is a uniformly sampled diffusion
timestep and $\epsilon \sim \mathcal{N}(0, I)$. Training data for
each sub-policy is augmented with the projected demonstrations
generated by the data augmentation procedure
(Section~\ref{sec:augmentation}), multiplying the effective dataset
size without additional human effort.

\subsubsection{Termination Condition Learning}
\label{sec:termination}
Each skill operator $o_i$ is decomposed into a sequence of action-step
sub-policies $\{\pi_{i,1}, \ldots, \pi_{i,J_i}\}$, inspired by the
options framework~\citep{SUTTON1999181}. Each sub-policy $\pi_{i,j}$ is
associated with a learned termination condition $\beta_{i,j} :
\mathcal{H}_t \rightarrow [0,1]$, which predicts the probability that
the current action step has been completed and execution should
transition to $\pi_{i,j+1}$. Termination is triggered when
$\beta_{i,j}(\mathcal{H}_t) \geq \tau_\beta$ for a threshold
$\tau_\beta$.

\paragraph{Transformer termination predictor} Termination prediction
requires reasoning over a temporal context window rather than a single
observation snapshot, since the completion of a contact-rich action
step (e.g., a grasp closing or a pallet insertion) manifests itself as a
pattern in recent observations rather than an instantaneous state
change. We, therefore, model $\beta_{i,j}$ as a
Transformer~\citep{vaswani2017attention} sequence classifier operating
over a sliding history window $\mathcal{H}_t = (\tilde{s}_{t-W},
\ldots, \tilde{s}_t) \in \mathbb{R}^{W \times d}$ of the last $W$
filtered observations.

The model first projects observations into a latent space of dimension
$d_h$ and adds a learned positional encoding:
\begin{equation}
    \mathbf{H}^{(0)} = \mathbf{X} \mathbf{W}_{\text{in}} + \mathbf{P},
    \quad \mathbf{X} \in \mathbb{R}^{W \times d},\;
    \mathbf{W}_{\text{in}} \in \mathbb{R}^{d \times d_h},\;
    \mathbf{P} \in \mathbb{R}^{W \times d_h},
    \label{eq:proj}
\end{equation}
where $\mathbf{P}$ is a learned parameter matrix. The projected
sequence is processed by an $L$-layer Transformer encoder with $N_h$
attention heads and feedforward dimension $4d_h$:
\begin{equation}
    \mathbf{H}^{(\ell)} = \text{TransformerLayer}\!\left(\mathbf{H}^{(\ell-1)}\right),
    \quad \ell = 1, \ldots, L.
    \label{eq:transformer}
\end{equation}
The representation of the final token $\mathbf{h}_W =
\mathbf{H}^{(L)}_{W,:} \in \mathbb{R}^{d_h}$ aggregates the full
causal context and is passed through a two-layer MLP prediction head:
\begin{equation}
    \beta_{i,j}(\mathcal{H}_t) = \sigma\!\left(\mathbf{W}_2\,
    \text{ReLU}\!\left(\mathbf{W}_1 \mathbf{h}_W + \mathbf{b}_1\right)
    + \mathbf{b}_2\right) \in [0,1],
    \label{eq:termination}
\end{equation}
where $\sigma$ denotes the sigmoid activation. The model is trained
with binary cross-entropy loss:
\begin{equation}
    \mathcal{L}_{\beta} = -\frac{1}{T}\sum_{t=1}^{T}
    \left[y_t \log \beta_{i,j}(\mathcal{H}_t) +
    (1 - y_t)\log\bigl(1 - \beta_{i,j}(\mathcal{H}_t)\bigr)\right],
    \label{eq:bce}
\end{equation}
%
%
where $y_t = 1$ if timestep $t$ falls within the $\delta$-window of a
breakpoint and $y_t = 0$ otherwise. Class imbalance, since termination
events are sparse relative to execution timesteps, is addressed by
positive-class reweighting with weight $\gamma = T / (2
|\hat{\mathcal{T}}| \delta)$.

The Transformer architecture is preferred over recurrent alternatives
(e.g., LSTM) for this task because the self-attention mechanism can
directly relate the current observation to any prior timestep in the
window, capturing long-range dependencies in contact dynamics without
the vanishing gradient limitations of sequential state
updates~\citep{vaswani2017attention}. In practice, we use $L=3$ layers,
$N_h=4$ attention heads, $d_h=128$, and a history window of $W=50$
timesteps.

\subsection{Execution of the Framework}
At execution time, the system operates as a closed-loop hierarchical
executor that interleaves symbolic planning with continuous skill
execution. Figure~\ref{fig:execution} illustrates the full execution
pipeline. We describe each stage in turn.

\subsubsection{Task specification and grounding}
A task is defined by an initial state $s_0 \in \mathcal{S}$ and a goal
state $s_g \in \mathcal{S}$ expressed over the predicate vocabulary
$\mathcal{F}$ of the learned domain $\sigma$.
In practice, the user selects a pair of snapshots from the set of
symbolic states observed during training; since each snapshot is
directly associated with a set of grounded predicates, planning is
immediate upon this selection.
If no suitable snapshot exists for $s_0$ or $s_g$, the user must
manually compose the state by specifying the relevant predicates from
$\mathcal{F}$, which requires interpreting the ASP-synthesized PDDL
domain $\sigma$ directly. 
Together, these form the PDDL
planning instance:
\begin{equation}
    T = \langle \mathcal{E}, \mathcal{F}, \mathcal{O}, s_0, s_g \rangle.
    \label{eq:planning_instance}
\end{equation}

\subsubsection{Symbolic planning} The planning instance $T$ is passed to
MetricFF~\citep{hoffmann2003metric}, a classical forward-chaining
heuristic planner, which searches for a sequence of grounded operators
$\mathcal{P} = [o_1, \ldots, o_{|\mathcal{P}|}]$ that transitions
$s_0$ to a state satisfying $s_g$. Each grounded operator
$\hat{o}_i = o_i(e^1_i, \ldots, e^{k_i}_i)$ binds specific entities
from $\mathcal{E}$ to the operator parameters. Planning operates
entirely in the discrete symbolic space and completes in milliseconds
for the task horizons considered, making it negligible in the overall
execution budget. The resulting plan $\mathcal{P}$ is human-readable
and can be inspected or overridden by an operator before execution
begins, providing a natural point of human oversight.

\subsubsection{Hierarchical skill execution} Execution proceeds by
iterating over the plan $\mathcal{P}$ and invoking each operator
policy $\pi_i$ in sequence. For each operator $o_i$, the Oracle
$\phi$ resolves the bound entities $\{e^1_i, \ldots, e^{k_i}_i\}$
to their current physical detections via $\rho^{-1}$ and constructs
the filtered ego-centric observation $\tilde{s}^{o_i}_t$
(Equation~\ref{eq:oracle_obs}). The operator policy automaton
$\pi_i = \langle \pi_{i,1}, \beta_{i,1}, \ldots, \pi_{i,J_i},
\beta_{i,J_i} \rangle$ then executes its sub-policies sequentially:
at each timestep $t$, the active sub-policy $\pi_{i,j}$ samples a
Cartesian displacement action via the reverse diffusion process
(Equation~\ref{eq:diffusion}), which is forwarded to the low-level
Cartesian controller. The Transformer termination predictor
$\beta_{i,j}$ evaluates the observation history $\mathcal{H}_t$
at every timestep; when $\beta_{i,j}(\mathcal{H}_t) \geq \tau_\beta$,
execution advances to $\pi_{i,j+1}$. When the final sub-policy
$\pi_{i,J_i}$ terminates, operator $o_i$ is considered complete and
the tracker updates the symbolic state $s_t$ by applying the effects
$\omega_i$. The full execution loop is summarized in
Algorithm~\ref{alg:execution}.

\begin{algorithm}[t]
\caption{Framework Execution at execution time}
\label{alg:execution}
\begin{algorithmic}[1]
\Require Planning domain $\sigma$, policies $\{\pi_i\}$,
         initial state $s_0$, goal $s_g$
\State Compute plan $\mathcal{P} = [o_1, \ldots, o_{|\mathcal{P}|}]$
       via MetricFF on $T = \langle \mathcal{E}, \mathcal{F},
       \mathcal{O}, s_0, s_g \rangle$
\For{each operator $o_i \in \mathcal{P}$}
    \State Resolve entities: $\mathcal{E}_{o_i} \leftarrow
           \phi(o_i, \mathcal{T}_t)$ \Comment{Oracle filtering}
    \For{each sub-policy $\pi_{i,j}$, $j = 1, \ldots, J_i$}
        \Repeat
            \State Observe $\tilde{s}^{o_i}_t \leftarrow$
                   ego-centric filtered observation
            \State Sample $a_t \sim p_\theta(\cdot \mid
                   \tilde{s}^{o_i}_t)$ \Comment{Diffusion policy}
            \State Execute $a_t$ on robot
            \State Update tracker $\mathcal{T}_{t+1}$,
                   $t \leftarrow t+1$
        \Until{$\beta_{i,j}(\mathcal{H}_t) \geq \tau_\beta$}
               \Comment{Termination condition}
    \EndFor
    \State Apply effects: $s_t \leftarrow s_t \oplus \omega_i$
\EndFor
\State \Return success if $s_t \models s_g$, failure otherwise
\end{algorithmic}
\end{algorithm}

\subsubsection{Generalization at execution time} The hierarchical structure
of Algorithm~\ref{alg:execution} provides generalization across all
three levels defined in Section~\ref{sec:prob_def}. Intra-task
generalization is handled implicitly: because $\tilde{s}^{o_i}_t$ is
recomputed at every timestep from live detector estimates, the policy
always operates on the current relative geometry rather than any
memorized absolute position, naturally accommodating object
displacements due to prior skill executions or environmental
perturbations. Inter-task generalization is achieved by replanning:
for a new goal $s_g'$ not present in any training episode, MetricFF
composes the available operators into a new sequence
$\mathcal{P}'$ without requiring any additional demonstrations,
as long as each individual operator in $\mathcal{P}'$ has been
learned. Cross-object generalization follows naturally from PDDL operators being lifted: at execution time, any object of the correct type can be substituted into an operator, regardless of whether that specific instance appeared in training demonstrations, provided the perception module can supply reliable pose estimates for it~\footnote{Integrating a completely new object within the planning domain requires a semantic name label being detected by an open-world detector and an object-type classification for planning.}.

\section{Evaluations}

We evaluate our Neuro-Symbolic model on two robots in different settings which we will
describe in the following.

\subsection{Robotic Systems}

\subsubsection{Autonomous Outdoor Forklift}

\begin{figure*}[t]
    \centering 
    \includegraphics[width=\textwidth]{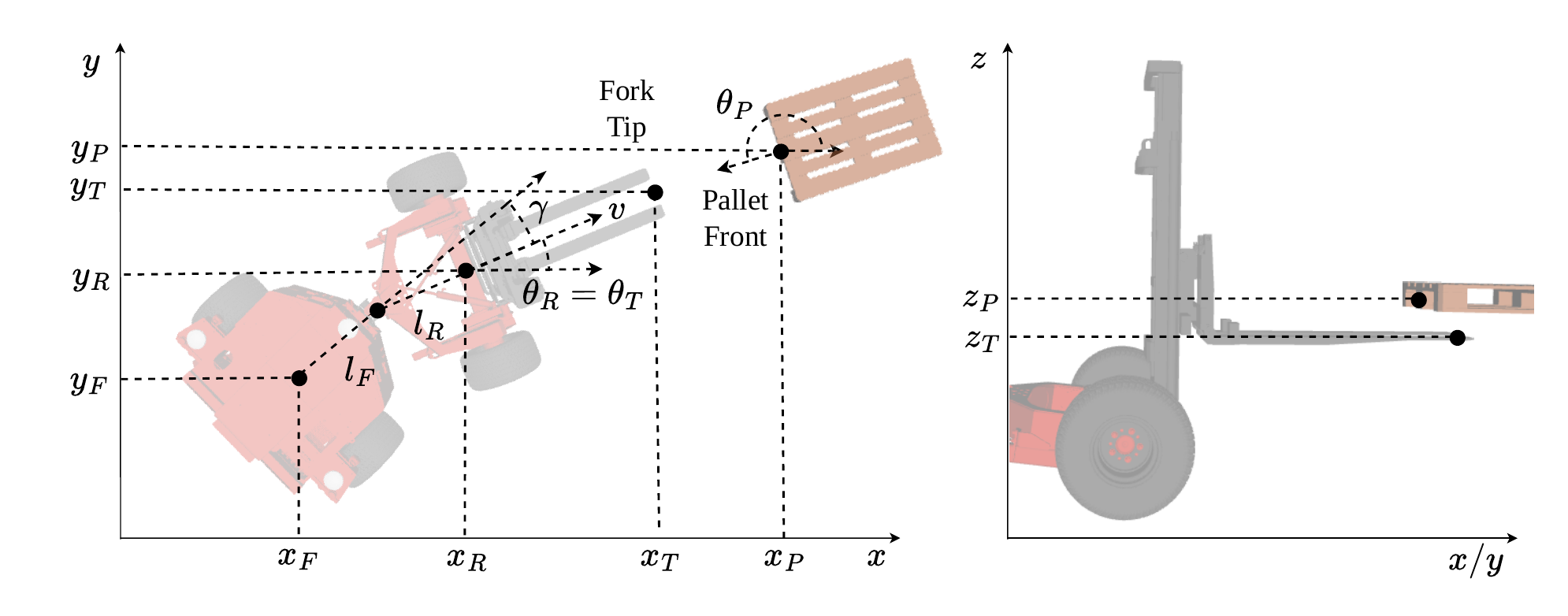}
    \vspace{-20pt}
    \caption{\textbf{Pallet alignment problem with respect to ADAPT kinematics.} Relevant states include: rear body pose $\mathbf{q}_R = [x_R, y_R, \theta_R]^\mathrm{T}$, fork tip pose $\mathbf{q}_T = [x_T, y_T, \theta_T, z_T]^\mathrm{T}$, and target pallet pose $\mathbf{q}_P = [x_P, y_P, \theta_P, z_P]^\mathrm{T}$, along with the articulation angle $\gamma$ and control inputs $v$ and $\dot{\gamma}$.}
    \label{fig:vehicle_control_diagram}
\end{figure*}

The first evaluation system is the ADAPT (Autonomous Dynamic
All-terrain Pallet Transporter) autonomous outdoor forklift for pallet
handling in unstructured environments~\citep{Huemer2025ADAPT} (see
Figure~\ref{fig:forklift}).  The vehicle is hydraulically actuated and
consists of a two-part articulated chassis where the front body houses
the motor and the rear body carries the fork lift mechanism.  The
system bundles five core capabilities required for fully autonomous
pallet transport, which are all explained in detail
in~\citep{Huemer2025ADAPT}. We provide a brief overview for
completeness:
\begin{itemize}
    \item \textbf{Navigation:} Collision-free, bi-directional path planning with a Hybrid A* planner adapted for articulated vehicles, combined with path tracking.
    \item \textbf{Localization and Mapping:} A factor-graph-based joint vehicle localization and pallet mapping framework providing centimeter-accurate pose estimates.
    \item \textbf{Pallet Detection:} A synthetically trained, geometry-based neural network pipeline for pallet detection and 6-DOF pose estimation.
    \item \textbf{Manipulation:} Precise pallet pick-up and placement via cascaded vehicle base and fork actuation control loops.
    \item \textbf{Obstacle Avoidance:} Real-time LiDAR-based terrain mapping and dynamic obstacle detection.
\end{itemize}
The aspect most relevant to this work is the task of picking up and placing pallets, where vehicle base control and fork positioning control play a crucial role, as they are actively used for data augmentation. Both are illustrated in Fig.~\ref{fig:vehicle_control_diagram} and explained as follows.

\paragraph{Vehicle Base Control}
ADAPT's chassis consists of two rigid bodies connected by a center articulation joint.
The rear body pose in the 2D plane is $\mathbf{q}_R = [x_R, y_R, \theta_R]^\mathrm{T}$, and the articulation angle between the two bodies is $\gamma$.
The kinematic model relating the state derivatives to the control inputs velocity $v$ and steering rate $\dot{\gamma}$ is:
\begin{equation}
\begin{bmatrix}
\dot{\mathbf{q}}_R \\
\dot{\gamma}
\end{bmatrix}
=
\begin{bmatrix}
    \dot{x}_R \\
    \dot{y}_R \\
    \dot{\theta}_R \\
    \dot{\gamma}
\end{bmatrix}
=
\begin{bmatrix}
\cos{(\theta_R)} & 0 \\
\sin{(\theta_R)} & 0 \\
\frac{\sin{(\gamma)}}{l_R \cos{(\gamma)} + l_F} & \frac{l_F}{l_R \cos{(\gamma)} + l_F} \\
0 & 1
\end{bmatrix}
\begin{bmatrix}
v \\
\dot{\gamma}
\end{bmatrix},
\label{eq:adapt_kinematics}
\end{equation}
where $l_F$ and $l_R$ are the distances from the articulation point to the front and rear axles, respectively.
The vehicle base is controlled by tracking the pallet pose $\mathbf{q}^d = [x_P, y_P, \theta_P]^\mathrm{T}$ as reference, so that the vehicle drives up to and aligns with the pallet front.
The pose error expressed in the vehicle body-fixed frame is:
\begin{equation}
\begin{bmatrix}
e_x \\
e_y \\
e_\theta
\end{bmatrix}
=
\begin{bmatrix}
\cos{(\theta_P)} & \sin{(\theta_P)} & 0 \\
-\sin{(\theta_P)} & \cos{(\theta_P)} & 0 \\
0 & 0 & 1
\end{bmatrix}
\begin{bmatrix}
x_R - x_P \\
y_R - y_P \\
\theta_R - \theta_P
\end{bmatrix},
\label{eq:ptc_error_system}
\end{equation}
comprising the longitudinal error $e_x$, lateral error $e_y$, and heading error $e_\theta$.
Minimising $e_y$ and $e_\theta$ yields the desired steering rate
\begin{equation}
\dot{\gamma}^d = -\frac{K_1 v (l_F + l_R)}{l_R}\,e_y - \frac{K_2 (l_F + l_R)}{l_R}\,e_\theta - \frac{v}{l_R}\,\gamma,
\label{eq:gamma_dot_desired}
\end{equation}
derived via Lyapunov-based analysis, where $K_1$ and $K_2$ are control gains.
The desired velocity is
\begin{equation}
v^d = \max\!\left(v_R - k_{v}(\dot{\gamma}^d)^{2},\; v_{\min}\right),
\label{eq:v_desired}
\end{equation}
reducing speed proportionally to curvature to maintain tracking accuracy, where $v_R$ is the nominal reference velocity and $v_{\min}$ a lower speed bound.

\paragraph{Fork Positioning Control}
Pallet pick-up requires the fork tips $\mathbf{q}_T = [x_T, y_T, \theta_T, z_T]^\mathrm{T}$ to be precisely aligned with the pallet front face $\mathbf{q}_P = [x_P, y_P, \theta_P, z_P]^\mathrm{T}$.
The pallet position is transformed into the fork tip coordinate frame:
\begin{equation}
\begin{bmatrix}
x_P' \\
y_P'
\end{bmatrix}
=
\begin{bmatrix}
\cos{(e_\theta)} & -\sin{(e_\theta)} \\
\sin{(e_\theta)} & \cos{(e_\theta)}
\end{bmatrix}
\begin{bmatrix}
x_P - x_T \\
y_P - y_T
\end{bmatrix},
\label{eq:adapt_ftt}
\end{equation}
where $e_\theta = \theta_P - \theta_T$ is the heading misalignment, and $x_P'$, $y_P'$ are the longitudinal and lateral distances to the pallet expressed in the fork tip frame.
Based on the transformation output, the reference signals for the fork actuators are derived directly: the fork shift reference $s^d = y_P'$ drives lateral alignment, the lift mast height reference $l^d = z_P$ corrects the vertical offset, and the tilt reference $\beta^d$ keeps the forks parallel to the ground.
The heading error $e_\theta$ is reduced by the vehicle base control loop described above.

\paragraph{Computational profile} The dominant cost at runtime is diffusion policy inference, which requires
$K$ denoising steps per action; in practice we use
$K=10$, yielding action inference at approximately $20\,Hz$. Perception and planning run concurrently on the same hardware, with symbolic planning introducing negligible overhead relative to policy inference. All components run on-board on an NVIDIA L4 GPU (Nuvo-9166GC embedded IPC), confirming the system's viability for edge deployment without any cloud offloading. 

\subsection{Evaluation Settings}
\label{sec:experiments}

We evaluate the full framework across three distinct experimental
settings designed to probe complementary aspects of the system: data
efficiency and manipulation precision on the real industrial forklift,
adaptive skill extension via single-demonstration transfer, and
cross-platform domain independence on the Kinova Gen3 robotic arm.
 

\subsubsection{Statistical Evaluation of Forklift Pallet Manipulation}
\label{sec:exp_forklift_stat}
 
To rigorously quantify the manipulation performance of the
framework under realistic variability, we conducted a large-scale
statistical evaluation of the full load-navigate-unload cycle on
the ADAPT platform.
Rather than fixing the initial vehicle pose relative to the pallet,
we reset the forklift autonomously between trials by exploiting the
onboard navigation module and the fork actuation control loops.
Concretely, at the start of each trial, a target reset pose
$\mathbf{q}_{\text{reset}} = [x, y, \theta]^\top$ is sampled
from a truncated Gaussian distribution
\begin{equation}
    \mathbf{q}_{\text{reset}} \sim
    \mathcal{N}(\boldsymbol{\mu}_{\text{nominal}},\,\Sigma_{\text{reset}})
    \;\big|_{\,\mathbf{q} \in \mathcal{R}},
    \label{eq:reset_dist}
\end{equation}
where $\boldsymbol{\mu}_{\text{nominal}}$ is the nominal approach
pose used during training, $\Sigma_{\text{reset}} =
\mathrm{diag}(\sigma_x^2, \sigma_y^2, \sigma_\theta^2)$ encodes
positional and angular uncertainty, and $\mathcal{R}$ is a
rectangular feasibility region that excludes poses that would
result in irreparable initial collisions or sensor occlusions.
We set $\sigma_x = \sigma_y = 0.4$\,m and
$\sigma_\theta = 15^\circ$, reflecting the realistic range of
positioning errors encountered in unstructured outdoor environments.
The navigation module drives the vehicle autonomously to
$\mathbf{q}_{\text{reset}}$ before each trial, without any manual
repositioning, making the evaluation fully unsupervised.
 
We report results under two complementary success criteria:
\emph{strong} success requires task completion without either a strong
collision (contact force exceeding a threshold) or a significant
pallet displacement ($>$\,10\,cm lateral shift at deposit);
\emph{soft} success permits minor contact events and marginal pallet
displacements ($<$\,10\,cm), reflecting industrial-grade tolerance
requirements.  Figure~\ref{fig:forklift_results} summarizes the
aggregate success rates as a function of the number of demonstrations
per skill.  To further characterize the performance dependence on the
initial pose, Figure~\ref{fig:forklift_reset_distribution} shows the
distribution of successes and failures over the forklift reset region,
where each episode is represented by an arrow encoding the initial
$(x_{\text{init}}, y_{\text{init}}, \psi_{\text{init}})$ pose relative
to the pallet.

Across the 30- and 10-demonstration settings, no strong spatial
correlation is observed between initial pose and outcome, indicating
that the policy generalizes robustly across the sampled reset
distribution. In the 5-demonstration regime, sensitivity increases
notably along the longitudinal axis $x$, with a coherent success
region spanning approximately 50\,cm, beyond which performance
degrades.  Failure modes are primarily attributed to policy training quality, as demonstration count is the dominant factor governing robustness. Nevertheless, we note that in some instances, perception noise such as lighting conditions adversely affects object detection and task success. We do not observe any obvious correlation between the pallet geometric pose and the agent performance.

 
\subsubsection{Long-Horizon Task Execution from Minimal Demonstrations}
\label{sec:exp_longhorizon}
 
To stress-test the data efficiency of the symbolic planning
component, we evaluated the framework on a compound long-horizon
task involving \emph{two pallets} and \emph{two distinct drop-off
zones}, a setting that requires the planner to sequence eight high-level operators ($\texttt{navigate}$,
$\texttt{load\_pallet}$, $\texttt{navigate}$,
$\texttt{unload\_pallet}$ $\times\,2$ per pallet) in the correct
order.
Importantly, no single training demonstration ever depicted a
two-pallet, two-zone task; rather the system was trained exclusively from
demonstrations of individual skills, each demonstrated at most
$n \in \{1, 5, 10, 30\}$ times.
At evaluation time, the user specifies the task by selecting the
initial and goal snapshots from the set of symbolic states observed
during training; MetricFF then synthesizes the correct multi-step
plan in less than one second.
 
This experiment directly validates \emph{inter-task
generalization}: the ability to compose learned operators into
novel sequences not represented in any single demonstration
trajectory.
The system successfully completed the two-pallet task in
$9/10$ trials with 10 base demonstrations per skill, confirming that the symbolic planner can extrapolate compositionally well beyond the
demonstrated task horizon.

\begin{figure}[t]
    \centering
    \begin{subfigure}{0.49\textwidth}%
        \centering
        \includegraphics[width=\linewidth]{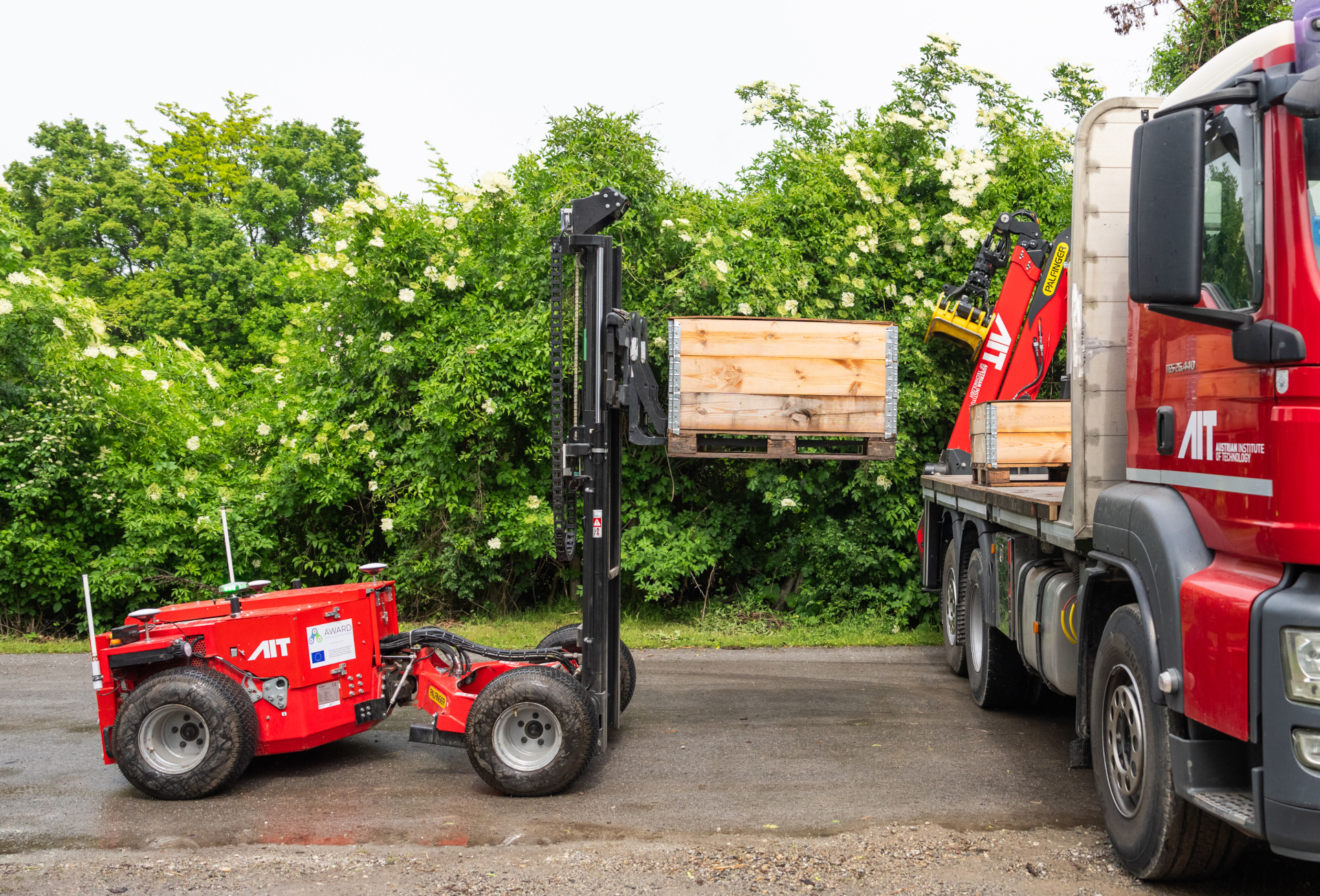}%
    \end{subfigure}%
    \hfill
\caption{
Main evaluation domain: an automated forklift for managing multiple pallets in outdoor scenario. The pallets can be stored on the ground or on a truck platform. \linebreak
Image credit: © AIT/tm-photography.}
\end{figure}
 
\subsubsection{Single-Demonstration Adaptation to a Novel Drop-Off Location}
\label{sec:exp_adaptation}
 
A central practical requirement in warehouse and construction-site
deployments is the ability to extend a deployed system to new task
configurations without full retraining.
We demonstrate this by introducing a \emph{truck-bed drop-off}
location that sits $z \approx 1.7$\,m above ground, substantially
outside the vertical range covered by the original training
demonstrations, which were all performed at ground level.

A single demonstration of the new unload task with $z$ height elevation (the height of the truck loading area) $\texttt{unload\_pallet\_on\_truck}$ skill was provided. The human
also classified such skill as complementary with the existing base
\texttt{unload} skill. The Oracle reuses the operator definition to identify the minimal observation space for the new skill, subtracts the dimensions already covered by existing skills, and returns only the residual, here, the relative
$z$ position between the end-effector and the drop-off location. A diffusion policy for this skill is then trained from the single demonstration. The ASP solver does not need to re-runs in such
scenario, as no additional abstraction is required. A new location is
simply added as one additional entity in the PDDL problem definition.
At evaluation time, the planner seamlessly integrates the new operator
into existing multi-step plans, for example, routing pallet transport
to the truck rather than a ground zone, without any modification to
the previously trained sub-policies.
 
This result highlights a key advantage of the neuro-symbolic
architecture: because skills are compositional building blocks
managed by a symbolic planner, extending the task repertoire
requires adding a single leaf to the operator graph rather than
retraining an end-to-end model from scratch.
The new operator is available for planning
immediately after the diffusion policy has been trained on the
(augmented) single demonstration.

\begin{figure*}[t]
    \centering
    \begin{subfigure}{0.42\textwidth}%
        \centering
        \includegraphics[width=\linewidth]{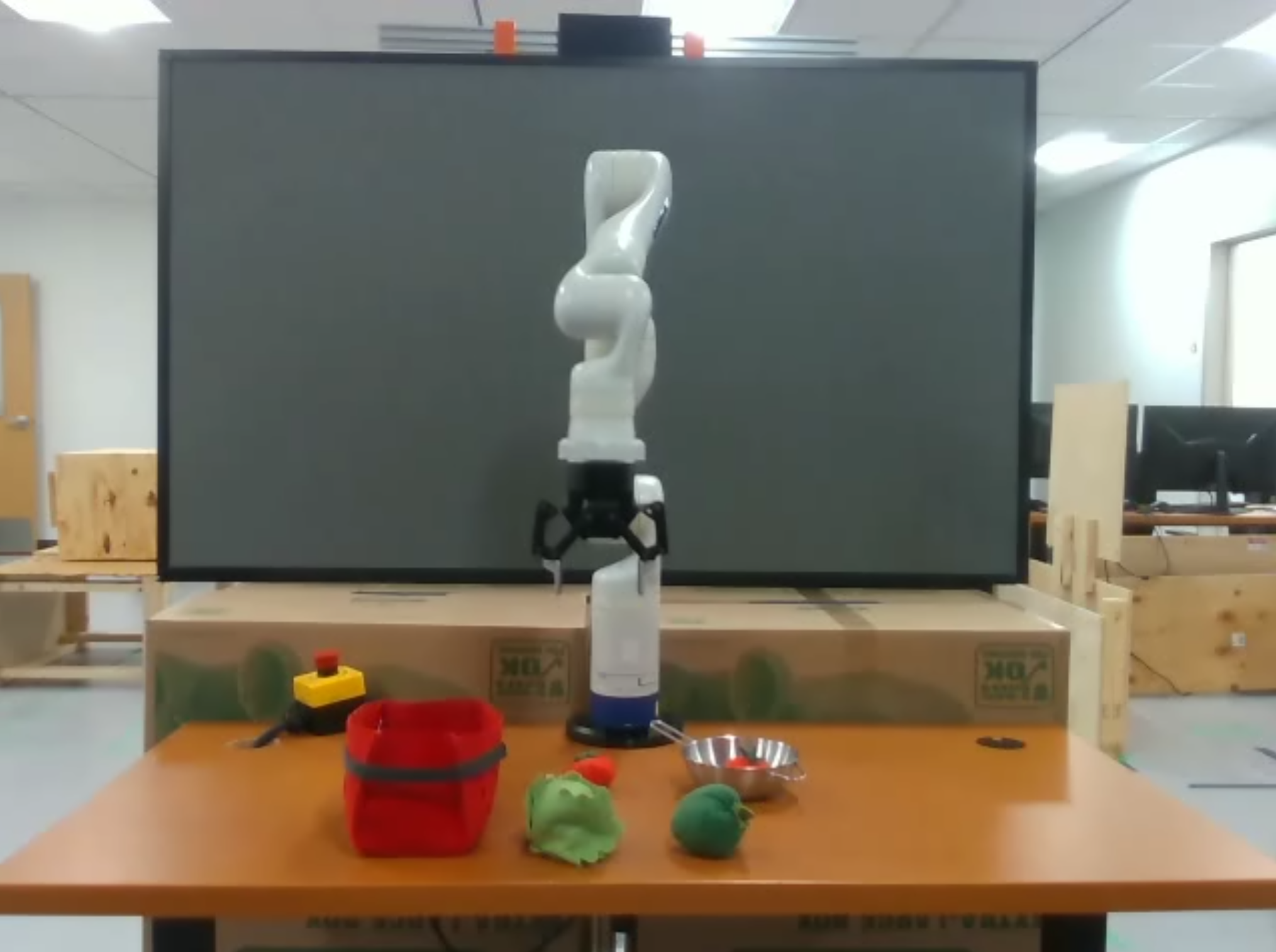}%
        \caption{Kinova Kitchen}
    \end{subfigure}%
    \hfill
    \begin{subfigure}{0.42\textwidth}%
        \centering
        \includegraphics[width=\linewidth]{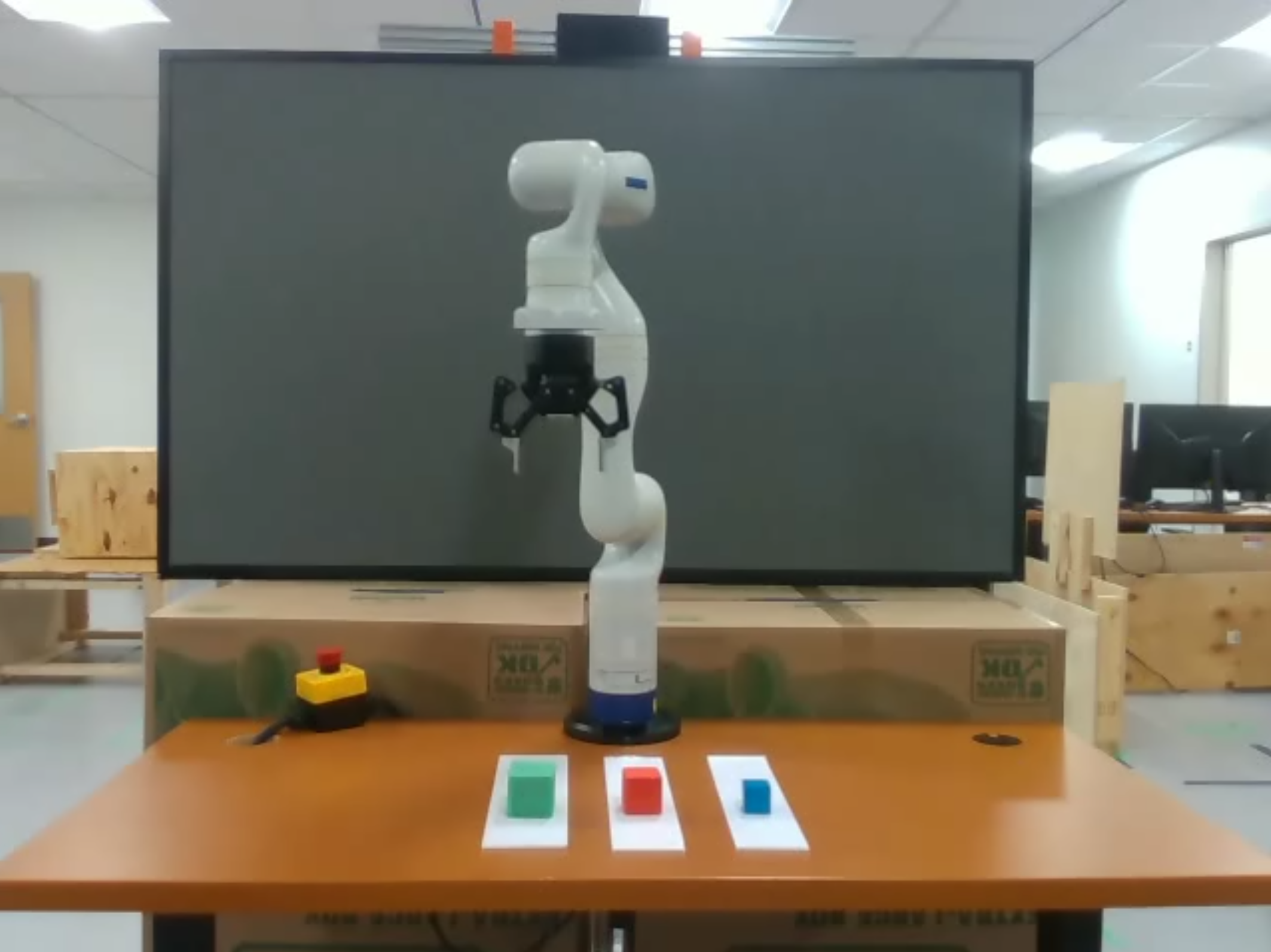}%
        \caption{Kinova Stack}
    \end{subfigure}%
    \hfill
\caption{{\em Validation domains:} We validate our approach on a
  different embodiment (Kinova robotic arm) across two domains:
  \texttt{Kitchen}, where the arm must organize objects on a kitchen
  table, and \texttt{Stacking}, where it must stack cubes in the order
  of their size.  The architecture mirrors the forklift domain, with
  the perception stack replaced by OWLv2 for object detection and
  depth data for 3D position estimation. From a \emph{single} human
  demonstration, the agent extracts waypoints, projects them onto new
  object configurations, and executes the corresponding controls to
  complete each task end-to-end. We further validate a knowledge
  distillation pathway, OWLv2 supervising a lightweight YOLOv8
  model, and demonstrated controls supervising compact diffusion
  policy networks, confirming that the architecture can be deployed
  on resource-constrained hardware without sacrificing task
  performance.}
\end{figure*}

\subsubsection{Cross-Platform Validation on a Kinova Gen3 Manipulator}
\label{sec:exp_kinova}

To demonstrate platform independence, we deployed the same symbolic
abstraction and learning pipeline unchanged on a Kinova Gen3 7-DOF
arm with a Robotiq 2F-85 gripper, across two standard benchmarks:
\emph{block stacking} and \emph{kitchen manipulation}. The sole
platform-specific adaptation is the perception stack, where pallet
detection is replaced by OWLv2 detections distilled into a YOLOv8
model. The same code path, prompts, and hyperparameters used for the
forklift produce valid PDDL domains for tabletop manipulation without
modification, confirming that VLM-driven annotation and ASP-based
domain synthesis are genuinely domain-agnostic. Both environments are
evaluated over 20 goal configurations.



\paragraph{End-Effector Control}
The arm is controlled in Cartesian space via a differential inverse
kinematics controller. Given a desired end-effector pose
$T^d = (R^d, w^d) \in SE(3)$, the task-space error is
\begin{equation}
    e = \begin{bmatrix} w^d - w \\ \mathrm{Log}(R^{\top} R^d) \end{bmatrix} \in \mathbb{R}^6,
    \label{eq:kinova_error}
\end{equation}
where $w \in \mathbb{R}^3$ is the current end-effector position,
$R \in SO(3)$ its orientation, and $\mathrm{Log}(\cdot)$ the
$SO(3)$ logarithmic map returning the rotation vector of the
relative orientation error expressed in the current body frame.
The desired joint velocities are computed via the damped least-squares
pseudoinverse of the geometric Jacobian $J \in \mathbb{R}^{6 \times 7}$:
\begin{equation}
    \dot{q}^d = J^{\dagger}(\lambda)\, K\, e,
    \qquad J^{\dagger}(\lambda) = J^\top\!\left(JJ^\top + \lambda^2 I\right)^{-1},
    \label{eq:kinova_ik}
\end{equation}
where $K \in \mathbb{R}^{6\times6}$ is a diagonal proportional gain matrix and
$\lambda > 0$ a damping factor ensuring numerical stability near singularities.

\paragraph{Grasp Execution}
Grasp and release actions are triggered symbolically by the task
planner and executed open-loop by the Robotiq gripper controller,
with grasp width set as a function of the target object type as
specified in the PDDL domain.

\paragraph{Perception pipeline}
On the Kinova platform, we replaced the pallet-specific detector with
the open-vocabulary OWLv2
model~\citep{Minderer_Gritsenko_Houlsby_2024}, prompted at runtime with
the target object names (e.g., \texttt{red cube}, \texttt{mug},
\texttt{drawer handle}).  OWLv2 produces axis-aligned bounding boxes
with associated confidence scores; 3D object positions are recovered
by back-projecting the box centroid through the calibrated depth image
from a wrist-mounted RGB-D sensor, yielding metric estimates
$\hat{\mathbf{p}}_e \in \mathbb{R}^3$ for each detected entity.  A
Hungarian-algorithm-based multi-object tracker maintains consistent
track identities across frames, resolving the correspondence between
YOLOv8 detection IDs and symbolic entities $\mathcal{E}$ via the
relational matching procedure described in Section~\ref{sec:mapping}. This fully
perception-agnostic setup required no additional annotation beyond
the skill vocabulary $\Lambda$.
 
\paragraph{Manipulation benchmarks}
The \emph{block stacking} benchmark presents the robot with three
colored cubes in randomized configurations and requires building a
target tower arrangement --- a long-horizon task (up to six chained
operators) demanding both intra-task generalization (unseen initial
positions) and inter-task generalization (novel tower orderings not
seen during training). A single \texttt{pick\_and\_place} demonstration
was provided, augmented to 20 trajectories via the pipeline described
in Section~\ref{sec:augmentation}. The \emph{kitchen manipulation}
benchmark increases perceptual and geometric complexity, requiring the
robot to manipulate objects with greater intra-category shape variation
(pan, basket, vegetables), and the planner to sequence heterogeneous
skills over a longer horizon. We reused the same controls learned on the \emph{block stacking} tasks accross the  \emph{kitchen manipulation} tasks.

\paragraph{Computational profile} On the kinova arm, we use $K=25$ diffusion inferences per action step, yielding action inference at approximately $25\,Hz$. Perception (Owlv2 or Yolov8~\citep{Minderer_Gritsenko_Houlsby_2024,yolov8_ultralytics}) and planning still run concurrently on the same hardware (Nvidia RTX4090). Object detection runs asynchronously at the YOLOv8 inference rate
($>30\,Hz$), decoupled from the policy inference loop via the
particle filter tracker which bridges any detection latency.
\section{Results}
\label{sec:results}



\begin{figure}[t]
    \centering
    \includegraphics[width=0.75\linewidth]{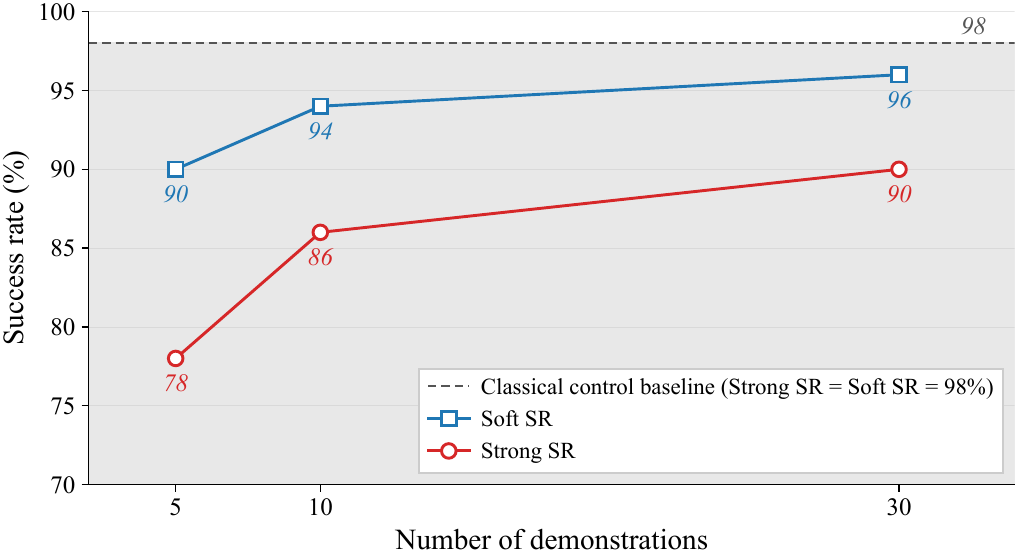}
    \caption{
    Success rate of the neurosymbolic approach as a function of the number of demonstrations for real-world forklift pallet management. The task consists of loading a pallet, navigating to a target area, and unloading it under randomized relative position to the pallet. We report two criteria: Strong SR (no significant collisions or pallet displacement) and Soft SR (task completed but minor collisions or pallet pushing allowed). The dashed line indicates the classical control baseline, which achieves 98\% under both criteria. Performance improves monotonically with additional demonstrations, with Soft SR approaching the baseline at 30 demonstrations.}
    \label{fig:forklift_results}
\end{figure}

\subsection{Trajectories to Plans}
\subsubsection{VLM Graph Construction}

\begin{figure}[t]
    \centering
    \begin{subfigure}[t]{0.48\linewidth}
        \centering
        \includegraphics[height=4.5cm, width=\linewidth, keepaspectratio]{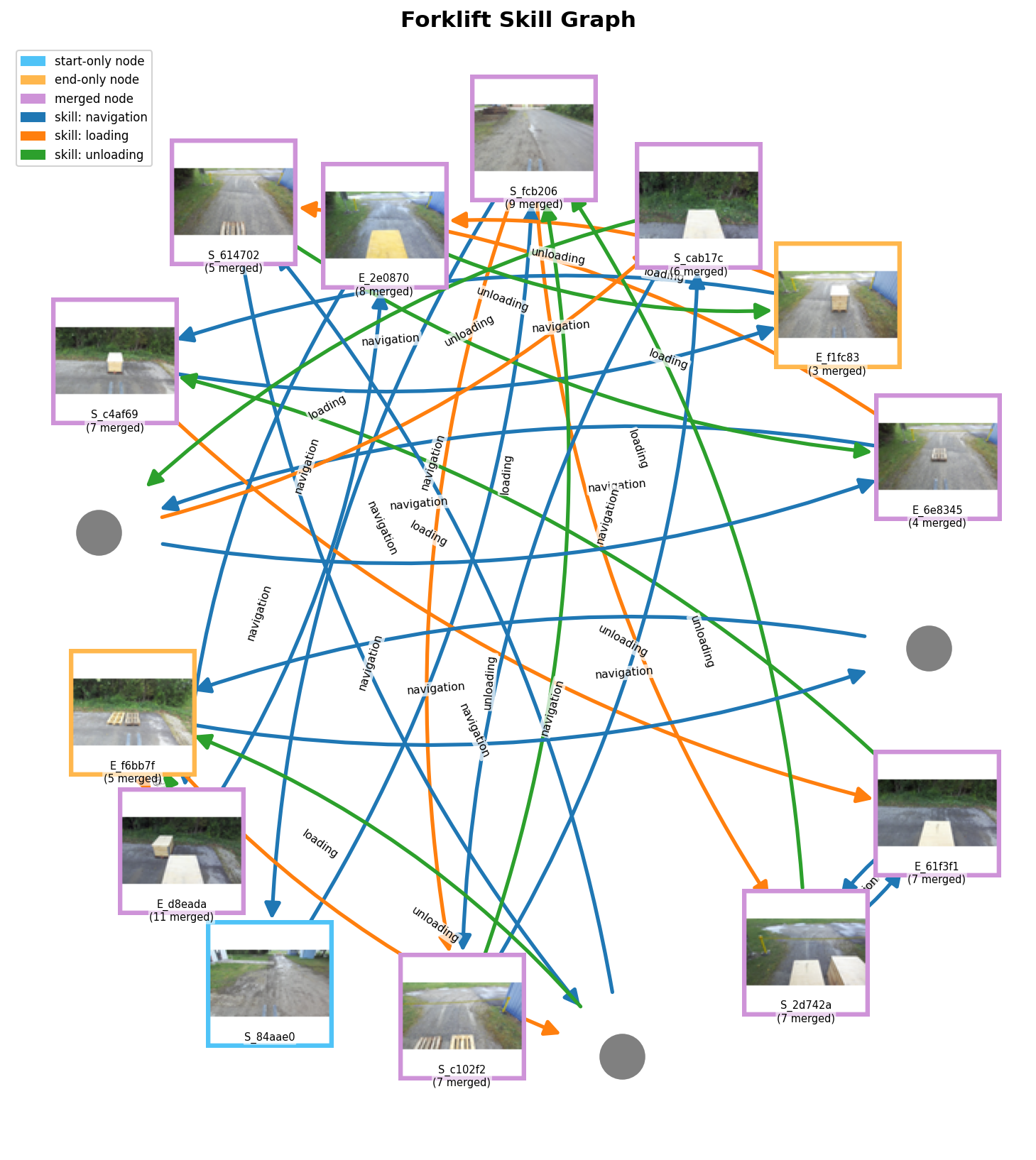}
        \caption{VLM Graph Construction Result}
        \label{fig:graph1}
    \end{subfigure}
    \hfill
    \begin{subfigure}[t]{0.48\linewidth}
        \centering
        \includegraphics[height=4.5cm, width=\linewidth, keepaspectratio]{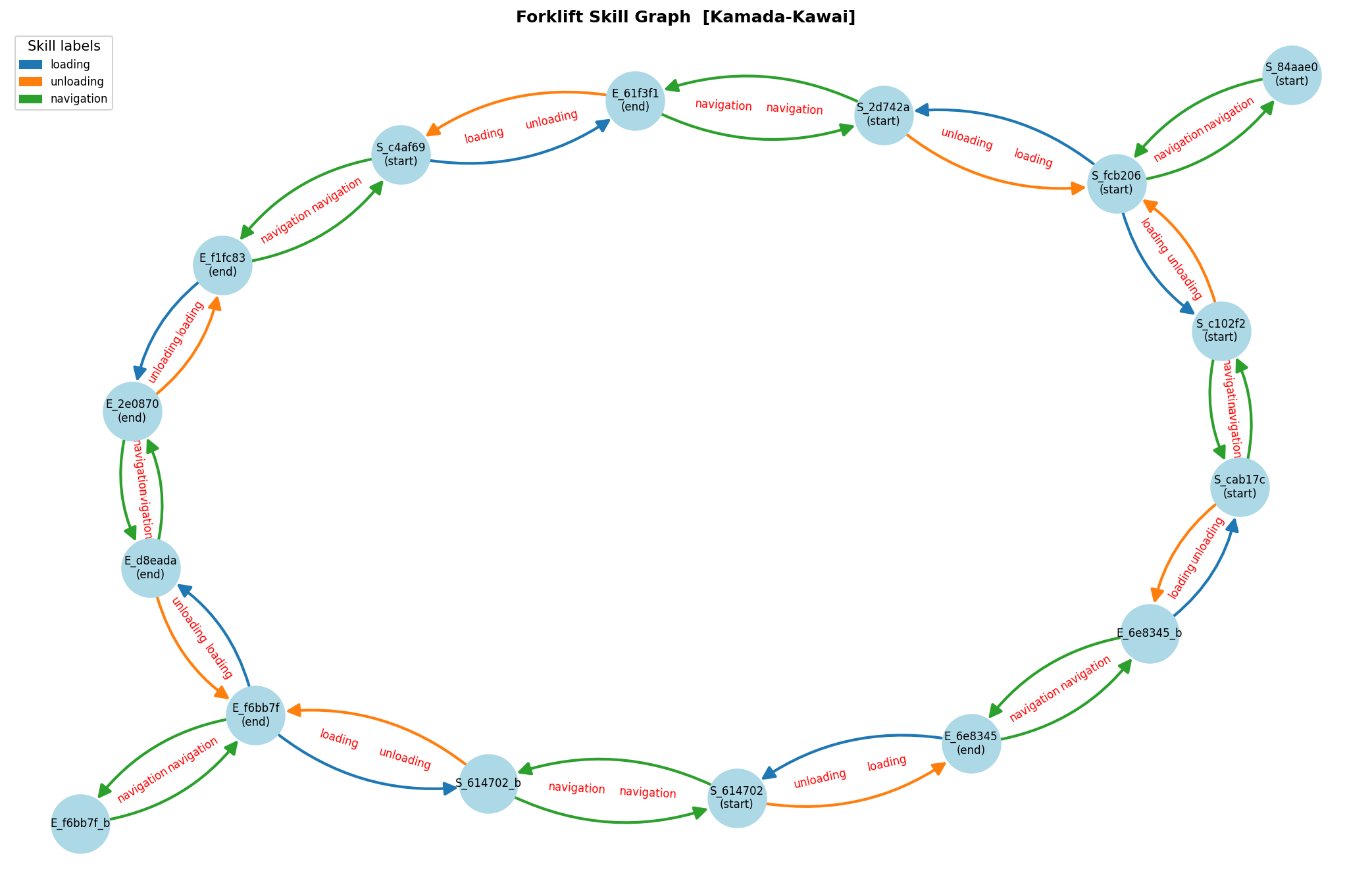}
        \caption{Graph bisimulation fed to the ASP Solver}
        \label{fig:graph2}
    \end{subfigure}
    \caption{VLM-constructed state-transition graph with three added nodes for
    visual completeness (left) and its bisimulation reduction (right), used as
    input to the ASP solver for automated action model induction. A bigger version of both graphs is attached in Appendix.~\ref{sec:appendix}.}
    \label{fig:side_by_side}
\end{figure}

The graph shown in Fig.~\ref{fig:graph1} is constructed by feeding demonstration
trajectories to a VLM, which clusters visually similar high-level states into
shared nodes and annotates the resulting edges with the corresponding skill
labels (\textit{navigate}, \textit{load}, \textit{unload}). Each skill was
demonstrated 5 times.
The resulting graph is then reduced via bisimulation
(Fig.~\ref{fig:graph2}), collapsing behaviorally equivalent nodes into canonical
representatives before being passed to the ASP solver.

We employ Gemini~3~Flash~\cite{google2025gemini3flash} as our VLM-based graph
constructor. Given a pair of observations from two distinct demonstrations, the
model is prompted to output a binary label (\texttt{true}/\texttt{false})
indicating whether the two observations correspond to the same abstract state,
and thus should be merged into a single node. Edges between nodes are then
annotated with a single-word skill label (\texttt{navigate}, \texttt{load}, or
\texttt{unload}), derived from the skill identifier associated with the
transition in the demonstration trajectory. The graph is built incrementally
over 5 demonstrations per skill: each new state, i.e., start or end snapshot of the demonstrated skill trajectory, is compared against all
existing nodes, and either merged into a matching node or instantiated as a new
one, with the corresponding labelled edge added. As reported in
Table~\ref{tab:vlm_accuracy}, the VLM achieves high accuracy on edge annotation
($98\%$) but more modest accuracy on node matching ($85\%$), confirming that
human supervision remains necessary to ensure graph correctness with current
VLMs.

\begin{table}[h]
    \centering
    \caption{VLM graph construction accuracy using Gemini~3~Flash.}
    \label{tab:vlm_accuracy}
    \begin{tabular}{lc}
        \toprule
        \textbf{Task} & \textbf{Accuracy} \\
        \midrule
        Node matching (state equivalence, boolean) & $85\%$ \\
        Edge annotation (skill label, one word)    & $98\%$ \\
        \bottomrule
    \end{tabular}
\end{table}

In Fig.~\ref{fig:graph1}, 3 nodes were manually injected by a human operator for full graph visualization, nevertheless a complete and noise-free graph is \textit{not} a prerequisite for
the pipeline to yield a functional planning domain: the ASP can, to some extent, operate over partial or imperfect graphs and still produce an
executable domain (see results published by~\citet{Bonet_Geffner_2020}). However, missing nodes or edges may cause the solver to
generalize over fewer observed transitions, potentially resulting in action
models that are correct but overly specific, requiring more parameters,
additional preconditions, or failing to unify effects that a fully observed
graph would have merged. The comparison between full and partial graph outputs
is analyzed in the next paragraph.

\subsubsection{Post-Hoc PDDL output Interpretation} The output of the ASP Solver does not come
with semantics attached; a human needs to interpret the gross ASP
outputs reported in Appendix.~\ref{sec:appendix}. The following action models are the interpreted results of the ASP computation
over the graph structure of the state space of the forklift domain, using either the full or partial observation graphs. The action models for the Kinova domain are reused from~\citet{Lorang_Lu_Huemer_Zips_Scheutz_2025}.

\noindent\makebox[\columnwidth]{\rule{\columnwidth}{0.6pt}}

\begin{center}
    \textbf{Forklift Multiple Pallets Storage Domain}
\end{center}

\noindent\makebox[\columnwidth]{\rule{\columnwidth}{0.6pt}}

\footnotesize
\setlength{\baselineskip}{1.3em}

\begin{mdframed}[linewidth=0.8pt, innerleftmargin=6pt, innerrightmargin=6pt,
                 innertopmargin=4pt, innerbottommargin=4pt]
\textbf{Full Domain Graph}

\smallskip
\noindent\underline{MOVE: $a_1(x_1, x_2)$}\\
\hspace*{1em} Pre: \texttt{(free\_location $x_1$), (forklift\_at $x_2$)}\\
\hspace*{1em} Eff: \texttt{(not (free\_location $x_1$)), (not (forklift\_at $x_2$)),}\\
\hspace*{4.2em} \texttt{(forklift\_at $x_1$), (free\_location $x_2$)}

\smallskip
\noindent\underline{UNLOAD: $a_2(x_1, x_2)$}\\
\hspace*{1em} Pre: \texttt{(forklift\_at $x_2$), (loaded\_pallet $x_1$)}\\
\hspace*{1em} Eff: \texttt{(not (loaded\_pallet $x_1$)), (free\_forklift), (at $x_1$ $x_2$)}

\smallskip
\noindent\underline{LOAD: $a_3(x_1, x_2)$}\\
\hspace*{1em} Pre: \texttt{(free\_forklift), (forklift\_at $x_2$), (at $x_1$ $x_2$)}\\
\hspace*{1em} Eff: \texttt{(not (free\_forklift)), (not (at $x_1$ $x_2$)), (loaded\_pallet $x_1$)}

\smallskip
\noindent\textbf{Objects:} $o_1, o_2, o_3, o_4$ \textit{(2 pallets $\times$ 2 locations)}
\end{mdframed}

\medskip






\clearpage
\begin{mdframed}[linewidth=0.8pt, innerleftmargin=6pt, innerrightmargin=6pt,
                 innertopmargin=4pt, innerbottommargin=4pt]
\textbf{One Node, Two Edges Missing}\\
\smallskip
\noindent\underline{MOVE: $a_1(x_1, x_2, x_3)$}\\ \quad
    {\footnotesize Static: $\neg(x_1{=}x_2),\ \neg(x_1{=}x_3)$}\\
\hspace*{1em} Pre: \texttt{(at $x_2$ $x_1$), (at $x_2$ $x_2$), (connected $x_3$ $x_1$)}\\
\hspace*{1em} Eff: \texttt{(not (at $x_2$ $x_1$)), (not (at $x_2$ $x_2$)),}\\
\hspace*{4.2em} \texttt{(at $x_1$ $x_1$), (at $x_1$ $x_2$)}\\
\smallskip
\noindent\underline{UNLOAD: $a_2(x_1, x_2)$}\\ \quad
    {\footnotesize Static: $\neg(x_1{=}x_2)$}\\
\hspace*{1em} Pre: \texttt{(loaded $x_1$), (at $x_1$ $x_1$), (at $x_2$ $x_2$), (connected $x_2$ $x_2$)}\\
\hspace*{1em} Eff: \texttt{(not (loaded $x_1$)), (not (at $x_2$ $x_2$)),}\\
\hspace*{4.2em} \texttt{(free\_forklift), (at $x_1$ $x_2$)}\\
\smallskip
\noindent\underline{LOAD: $a_3(x_1, x_2)$}\\ \quad
    {\footnotesize Static: $\neg(x_1{=}x_2)$}\\
\hspace*{1em} Pre: \texttt{(free\_forklift), (at $x_2$ $x_1$), (at $x_2$ $x_2$), (connected $x_1$ $x_1$)}\\
\hspace*{1em} Eff: \texttt{(not (free\_forklift)), (not (at $x_2$ $x_1$)),}\\
\hspace*{4.2em} \texttt{(loaded $x_2$), (at $x_1$ $x_1$)}\\
\smallskip
\noindent\textbf{Objects:} $o_1, o_2, o_3, o_4$
\textit{(2 pallets $\times$ 2 locations)}
\end{mdframed}

\medskip

\begin{mdframed}[linewidth=0.8pt, innerleftmargin=6pt, innerrightmargin=6pt,
                     innertopmargin=4pt, innerbottommargin=4pt]
    \textbf{Two Nodes, Six Edges Missing}\\
    \smallskip
    \noindent\underline{MOVE: $a_1(x_1, x_2, x_3)$}\\ \quad
        {\footnotesize Static: $\neg(x_1{=}x_2)$}\\
    \hspace*{1em} Pre: \texttt{(at $x_1$), (free $x_3$),}\\
    \hspace*{2.5em} \texttt{(connected $x_1$ $x_1$), (connected $x_2$ $x_2$), (connected $x_3$ $x_1$)}\\
    \hspace*{1em} Eff: \texttt{(not (at $x_1$)), (at $x_2$)}\\
    \smallskip
    \noindent\underline{UNLOAD: $a_2(x_1, x_2)$}\\ \quad
        {\footnotesize Static: $\neg(x_1{=}x_2)$}\\
    \hspace*{1em} Pre: \texttt{(at $x_1$), (at $x_2$), (loaded $x_1$), (connected $x_1$ $x_2$)}\\
    \hspace*{1em} Eff: \texttt{(not (loaded $x_1$)), (free $x_2$)}\\
    \smallskip
    \noindent\underline{LOAD: $a_3(x_1, x_2, x_3)$}\\ \quad
        {\footnotesize Static: $\neg(x_1{=}x_2),\ \neg(x_1{=}x_3),\ \neg(x_2{=}x_3)$}\\
    \hspace*{1em} Pre: \texttt{(at $x_2$), (at $x_3$), (free $x_1$), (free $x_3$), (connected $x_2$ $x_3$)}\\
    \hspace*{1em} Eff: \texttt{(not (free $x_3$)), (loaded $x_2$)}\\
    \smallskip
    \noindent\textbf{Objects:} $o_1, o_2, o_3, o_4$
    \textit{(2 pallets $\times$ 2 locations)}
    \end{mdframed}

\setlength{\baselineskip}{\normalbaselineskip}
\normalsize

\subsection{Forklift Pallet Management}
\label{sec:results_forklift}

Figure~\ref{fig:forklift_results} reports the primary quantitative
results on the real-world forklift evaluation.
The neuro-symbolic framework achieves a strong success rate of
$90\%$ with 30 demonstrations, thus
approaching the $98\%$ ceiling set by the hand-engineered classical
controller while requiring zero manual domain engineering.
And it also maintains a high success rate of
$86\%$ with only 10 demonstration.  
The graceful degradation from 30 to 5 demonstrations
($90\% \to 78\%$ strong SR) reflects the diminishing density of
training data in the ego-centric observation space.
 
\begin{figure*}[t]
    \centering
    \begin{subfigure}{0.32\textwidth}%
        \centering
        \includegraphics[width=\linewidth]{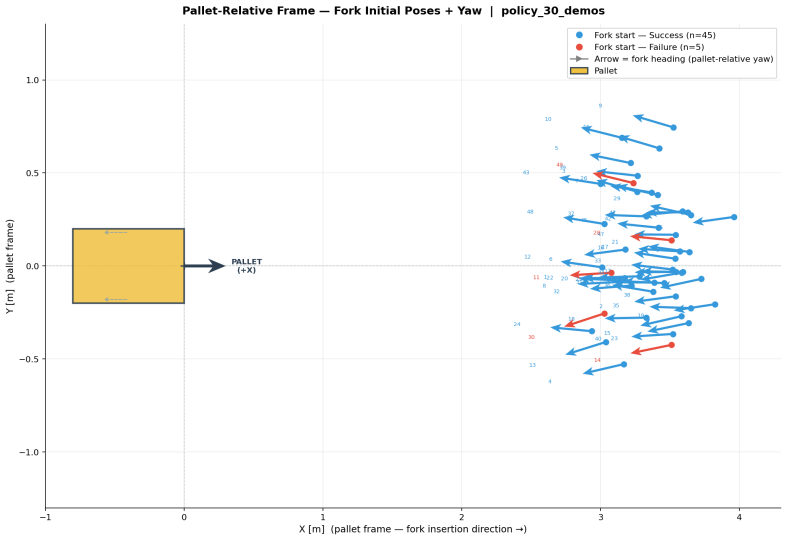}%
        \caption{30 demonstrations}
    \end{subfigure}%
    \hfill
    \begin{subfigure}{0.32\textwidth}%
        \centering
        \includegraphics[width=\linewidth]{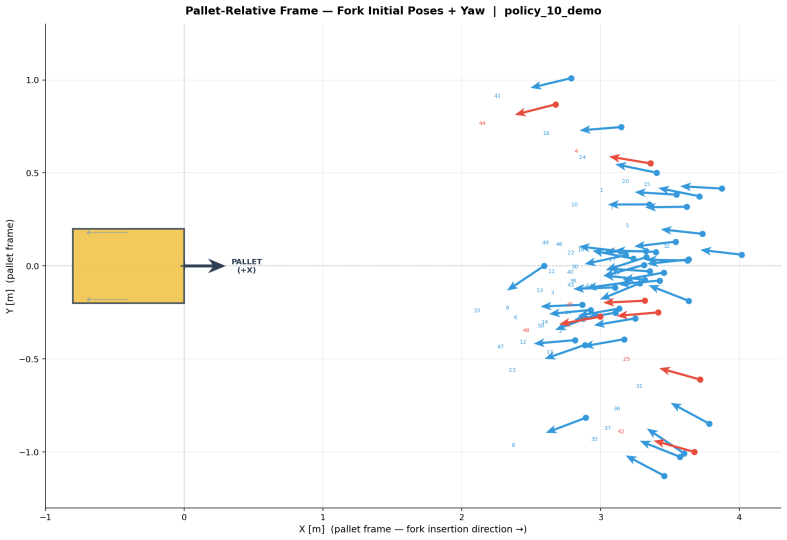}%
        \caption{10 demonstrations}
    \end{subfigure}%
    \hfill
    \begin{subfigure}{0.32\textwidth}%
        \centering
        \includegraphics[width=\linewidth]{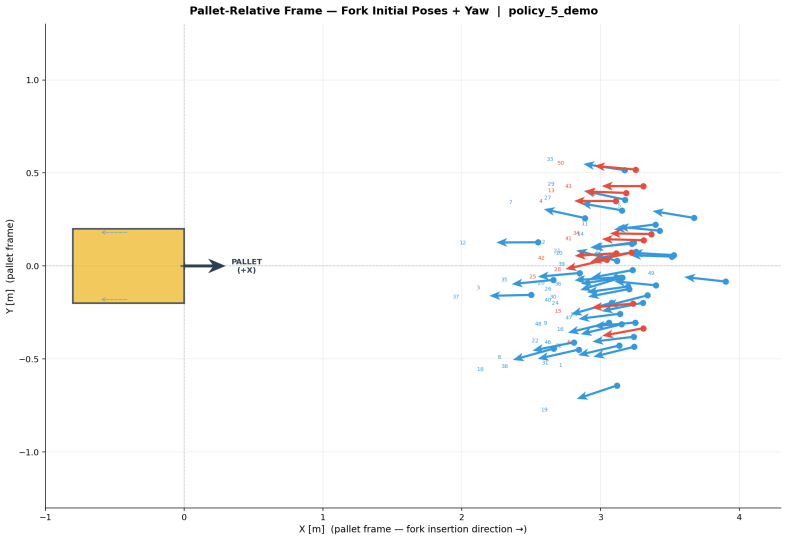}%
        \caption{5 demonstrations}
    \end{subfigure}
    \caption{
    Distribution of performance (success/failure) of the forklift pallet loading task over the forklift reset region of the state. Each episode reset pose is represented by an arrow parametrized with origin $(x_{\text{init}}, y_{\text{init}}, \textit{yaw}_{\text{init}})$ relative to the pallet location. \textcolor{blue}{Blue arrows} represent successes, while \textcolor{red}{Red arrows} represent failures. Reset configurations are sampled from a Gaussian distribution around a nominal pose.
    The three subfigures correspond to policies trained with 30, 10, and 5 demonstrations, respectively. No strong global correlation is observed between task success and initial pose. In the 5-demonstration setting, performance shows higher sensitivity to $x$, with a smooth success region of approximately $50\,\text{cm}$. Failures are primarily correlated with perception noise (lightning notably impacts the detections) and the number of demonstrations.}
    \label{fig:forklift_reset_distribution}
\end{figure*}

Three findings deserve emphasis.
\textbf{(i) The soft success rate remains consistently high.}
The gap between strong and soft success ($\approx 6$–$12$\,pp) is
attributable primarily to minor fork-tip contacts during the
alignment phase, not to planning failures or gross manipulation
errors.
This gap narrows as the number of demonstrations grows, consistent
with the diffusion policy better covering the multi-modal
distribution of approach trajectories.
\textbf{(ii) Classical control provides an engineering ceiling,
not a fair baseline.}
The classical controller achieves $98\%$ by relying on a precisely
hand-crafted, fully specified alignment controller, a task-specific
artefact that required months of expert engineering and encodes
detailed kinematic and geometric knowledge of the forklift-pallet
interaction.
Our framework treats the control layer as a \emph{prior}: low-level
controllers handle command-tracking and physical stabilization, freeing
the learned components to operate at a higher level of abstraction.
The Oracle grounds each skill to the relevant object poses, determining
\emph{where} to move, while diffusion policies capture \emph{how} to
move by learning continuous motion behavior from demonstrations.
At the top level, the symbolic planner determines \emph{what} skill to
execute and \emph{when}, chaining operators into coherent task-level
plans.
The human role is thereby reduced from \emph{engineering a complete
task-level controller} to \emph{demonstrating a skill}: rather than
explicitly specifying preconditions, alignment logic, and control gains
for each new task, the expert performs as few as one demonstration, and
the framework distributes that demonstrated intent across the planning
and policy layers automatically. A low-level motion controller remains
a prerequisite, but its development or learning is a strictly one-time effort,
once acquired, controls are shared across all tasks of the system. Importantly,
this one-time cost is of a fundamentally different nature than
task-level engineering: a motion controller need only capture the
platform's physical capabilities, not the semantics of any particular
task, object configuration, or alignment scenario. Every new skill,
pallet type, or environment is then handled purely through
demonstration, with no further control engineering required.
The $8$--$12$\,pp gap between the two approaches thus quantifies
the tradeoff between full expert specification and
demonstration-driven learning: performance against generality and
deployment speed.
For reference, informal human trials suggest that
operator performance on these tasks falls within the same range as
our agent trained on 30 demonstrations.
\textbf{(iii) The reset-pose distribution (Fig.~\ref{fig:forklift_reset_distribution})
reveals that failures are driven by the number of training samples and perception rather than pose geometry.}
No strong spatial correlation is observed between initial pose and task outcome within the demonstrated configuration space, suggesting robust generalization across the tested distribution. This diagnosis was made possible by the
modular transparent design of our architecture: because perception, planning, and
control are explicit, decoupled components, practitioners can isolate failure
modes to a specific module and reason about their root cause directly, without
resorting to black-box attribution methods. This interpretability has enormous practical value for
deployment in safety-critical robotic systems and represents a
distinct advantage over end-to-end learned policies, where such
failure attribution is generally intractable.

\subsection{Long-Horizon Composition and Rapid Adaptation}
\label{sec:results_composition}
 
The long-horizon two-pallet, two-zone experiment demonstrates that
compositional planning from learned operators incurs negligible
overhead: the MetricFF planner produces the correct eight-operator
sequence in under one second, and end-to-end task execution
proceeds without any additional human intervention.
Failures in this setting are exclusively attributable to low-level
execution errors (primarily perception and forks insertion within pallets collision), confirming that the symbolic layer correctly captures task
structure even in compositions never seen during training.
 
The single-demonstration truck-bed adaptation experiment is
particularly instructive.
The fact that a single new demonstration, is
sufficient to extend over an existing skill policy speaks to the modularity of the architecture.
End-to-end imitation learning approaches would
require collecting and labeling a new dataset for the entire
task, retraining the full policy, and re-evaluating across all
task configurations; our framework localizes the update to a
single affected operator and reuses all previously acquired
knowledge without modification.
 
\subsection{Cross-Platform Arm Experiments}
\label{sec:results_arm}
 
Results on the Kinova Gen3 benchmarks confirm the domain
independence of the pipeline.
The same framework, with no changes to symbolic abstraction,
planning, or learning components (the perception only is different), produces valid, executable
PDDL domains for both the stacking and kitchen environments.
It is all the more important to note that our framework has been trained with{\em only a \em single demonstration} in both the kitchen and stacking tasks. It utilized
this single demonstration to project and augment the data in the real world using existing controls, before abstracting automatically the planning domain (VLM annotation, graph construction, ASP solver) to solve the long-horizon problems.
Crucially, unlike end-to-end approaches, our framework
generalizes across novel task orderings and reuses the same controls across all tasks (inter-task
generalization) without any additional demonstrations, whereas
end-to-end methodologies require retraining for each new goal
configuration.
 
Also, note the ASP solver's ability to represent
non-spatial temporal operators (e.g., \texttt{wait})~\cite{Lorang_Lu_Huemer_Zips_Scheutz_2025},
yielding a structurally richer PDDL domain than clustering-based abstraction methods, which collapse all pre-conditions to purely spatial predicates.

\section{Discussion}
\label{sec:discussion}
The results raise several points worth examining beyond the quantitative performance figures.

\textbf{On the role of the classical controller.}
A recurring question in neuro-symbolic robotics is where the boundary between learned and engineered components should lie. Our results suggest that this boundary need not be fixed: by treating low-level controllers as fixed \emph{priors} rather than as objects of learning, the framework gains reliable, sample-efficient low-level execution and concentrates the learning budget on higher-level structure, namely operator sequencing, precondition abstraction, and motion direction, where demonstrations are most informative. The residual gap to the classical controller's performance is therefore not a failure of the learning approach but a quantification of what hand-engineering uniquely supplies: tightly-tuned kinematic geometry and fine-grained alignment logic. Whether this gap is worth closing through further demonstration collection or through targeted engineering is a deployment decision rather than a methodological one.

\textbf{On interpretability as an operational property.}
The modular decomposition into perception, planning, and control components produces interpretability as a practical side-effect rather than as a post-hoc explanation tool. In the forklift experiments, failure attribution required no gradient-based or surrogate-model analysis; failures could be traced directly to identifiable modules, specifically perception noise under varying illumination conditions or insufficient demonstration density in particular regions of the observation space. For safety-critical deployment, this diagnostic tractability is arguably as important as the success rate itself: an operator can act on a diagnosis (improve lighting, collect targeted demonstrations) in a way that is simply not possible when attribution relies on black-box methods.

\textbf{On failure recovery.}
The current framework operates in open-loop at the task level: the symbolic plan is executed sequentially without replanning on failure. While the Transformer termination predictor absorbs low-level execution noise, mid-task perception failures or unexpected environmental changes can cause silent, unrecoverable downstream errors. Closing this loop requires a reliable symbolic state estimator, one capable of determining at each step whether the preconditions for the next operator are genuinely satisfied. This is technically non-trivial: VLM-based state estimation introduces its own failure modes, including ambiguous scenes and partial occlusion, and replanning under incorrect state estimates risks compounding errors. We regard this as the most pressing open problem for the framework's transition from demonstration-driven learning to genuinely reactive autonomy.

\textbf{On data efficiency and its limits.}
The single-demonstration performance on the arm benchmarks and the graceful degradation curve on the forklift (from thirty to five demonstrations) both suggest that the framework's data efficiency stems from a combination of factors: data augmentation within existing control trajectories, the compressive effect of ego-centric observation filtering, and the ability of the symbolic planner to reuse acquired operators in novel compositions. However, data efficiency is not monotone in task complexity. Operators with multimodal approach distributions, where the same high-level action can be initiated from qualitatively different configurations, are underserved by small demonstration sets, as the diffusion policy cannot adequately cover the distribution's modes. This is consistent with the observed sensitivity to the number of training samples rather than to pose geometry, and it defines a natural criterion for deciding when additional demonstrations are warranted.
\section{Limitations}
\label{sec:limitations}

Beyond failure recovery, discussed separately in Section~\ref{sec:discussion}, it is important to point
out some additional limitations, notably ``scaling up'', a problem
that bedevils all current task and skill learning approaches. While
the proposed framework scales well for modest skill repertoires
($|\mathcal{O}| \leq 10$ in all experiments), but two bottlenecks
emerge as the library grows. First, the ASP solver's search
complexity increases with the number of graph edges, potentially
making domain synthesis expensive or even intractable for large task
libraries.  A second problem is related to the VLM-based bisimulation
compression which may conflate semantically distinct states whose
visual differences are subtle (e.g., two assembly states differing
only in a single bolt orientation), leading to unsolvable satisfiability problems, i.e., no working ASP output, or an under-specified PDDL
domain. Currently, human supervision over the graph construction is still necessary. Hierarchical domain construction, grouping operators into
composable sub-domains, is a natural mitigation we leave to future
work.  Additionally, skill segmentation relies on an automatic
change-point detector applied to the motion signal which may fail to
segment skills with smooth, overlapping velocity profiles (e.g., a
pick immediately followed by a place with no intermediate pause),
negatively impacting sub-policy training. While the Transformer
termination predictor partially absorbs this noise, fundamentally
unreliable segmentation degrades both graph construction and diffusion
policy training. A semantics-aware segmenter conditioned on
VLM-predicted event boundaries would be a natural direction to improve
robustness.

\section{Conclusion}
\label{sec:conclusion}
 
We have presented \textit{Build on Priors} --- leveraging foundation models,
classical controls, and perception as structural priors --- a scalable neuro-symbolic framework for data-efficient robot manipulation that
learns both a symbolic planning domain and a set of neural skill
policies from as few as one to thirty unannotated demonstrations
per skill, without manual domain engineering or symbolic
supervision.
The framework integrates four tightly coupled components:
VLM-driven automated annotation and bisimulation-based graph
construction, ASP-based synthesis of expressive PDDL domains,
oracle-guided ego-centric observation filtering, and
control-level imitation learning with real-world data augmentation.
 
Empirical validation on a real autonomous industrial forklift
and on a Kinova Gen3 robotic arm across stacking and kitchen
benchmarks demonstrate  that the proposed model produces
interpretable, generalizable, and data-efficient behavior across
qualitatively different robotic platforms and task domains.
%
%
On the forklift, the neuro-symbolic framework achieves a strong
success rate of 90\% with 30 demonstrations and 86\% with only
10, approaching the ceiling set by a fully hand-engineered
classical controller while requiring only a human-provided skill
vocabulary rather than full manual domain specification.
Statistical evaluation over a Gaussian distribution of reset poses
confirms that the framework maintains robust manipulation
performance across a wide range of initial conditions, degrading
gracefully toward the boundary of the feasibility region.
Long-horizon experiments with two pallets and two drop-off
zones demonstrate that compositional planning from learned
operators generalizes to task structures never seen in training,
and a single-demonstration truck-bed adaptation experiment
shows that the modular symbolic architecture localizes new skill
acquisition to the affected operator without disturbing previously
learned knowledge.
 
Taken together, these results support the broader claim: {\em grounding
control learning in symbolic reasoning, and grounding symbolic
reasoning in perceptual abstraction driven by large pre-trained
models, yields a practical path toward scalable, expert-free, and
interpretable robot learning in real-world settings.}
The proposed framework is neither a pure imitation learning method nor a
pure symbolic planning system, but rather a tightly integrated
symbolic-subsymbolic hierarchy in which each layer strengthens the others.
The symbolic planner decomposes the combinatorial task-planning
problem into independently learnable sub-problems; the oracle
reduces each sub-problem to its minimal ego-centric form; the
diffusion policy solves each sub-problem from a compact, smooth
learning target; and the VLM supplies the semantic supervision
that makes automatic domain construction possible.
 
Looking forward, several directions appear particularly promising.
First, closing the perception-to-planning loop via a reliable
symbolic state estimator would enable genuine failure recovery and
replanning, transforming the framework from an open-loop sequencer
into a fully reactive task-and-motion planning system.
Second, extending the oracle and data augmentation machinery to
mobile manipulation platforms would broaden the range of
deployable tasks significantly.
Third, integrating the neuro-symbolic planning layer with
large-scale VLA-style controllers for individual skills, using
the symbolic planner for task-level sequencing and a VLA for
rich visuomotor execution, represents a compelling hybrid that
would combine the data efficiency and interpretability of our
framework with the perceptual generality of foundation-model-based
controllers.

\clearpage     
\appendix
\section{Appendix}
\label{sec:appendix}

\subsection{VLM Graph Construction}

\begin{figure}[H]
    \centering
    \begin{subfigure}[t]{0.23\linewidth}
        \includegraphics[width=\linewidth, trim=0 0 0 20, clip]{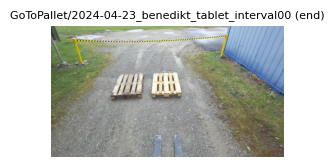}
    \end{subfigure}\hfill
    \begin{subfigure}[t]{0.23\linewidth}
        \includegraphics[width=\linewidth, trim=0 0 0 20, clip]{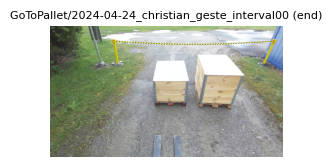}
    \end{subfigure}\hfill
    \begin{subfigure}[t]{0.23\linewidth}
        \includegraphics[width=\linewidth, trim=0 0 0 20, clip]{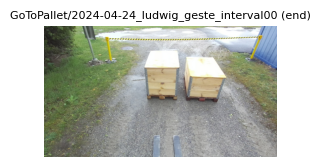}
    \end{subfigure}\hfill
    \begin{subfigure}[t]{0.23\linewidth}
        \includegraphics[width=\linewidth, trim=0 0 0 20, clip]{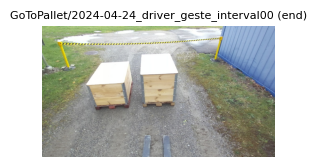}
    \end{subfigure}
    \vspace{0.5em}
    \\
    \noindent\textit{Node A: Forklift at loading bay, no pallet loaded, two pallets present in scene.}
    \vspace{1em}
    \begin{subfigure}[t]{0.23\linewidth}
        \includegraphics[width=\linewidth, trim=0 0 0 20, clip]{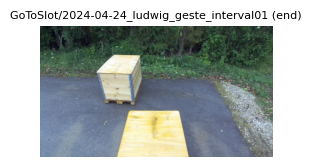}
    \end{subfigure}\hfill
    \begin{subfigure}[t]{0.23\linewidth}
        \includegraphics[width=\linewidth, trim=0 0 0 20, clip]{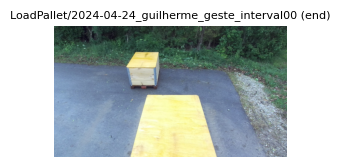}
    \end{subfigure}\hfill
    \begin{subfigure}[t]{0.23\linewidth}
        \includegraphics[width=\linewidth, trim=0 0 0 20, clip]{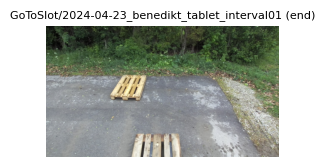}
    \end{subfigure}\hfill
    \begin{subfigure}[t]{0.23\linewidth}
        \includegraphics[width=\linewidth, trim=0 0 0 20, clip]{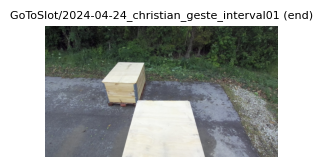}
    \end{subfigure}
    \vspace{0.5em}
    \\
    \noindent\textit{Node B: Forklift at unloading bay, pallet loaded on forks, one pallet remaining in scene.}
    \caption{Representative observations merged into a single graph node by
    Gemini~3~Flash, guided by a semantic domain hint defining high-level states
    in terms of pallet load status, number of pallets present in the scene, and
    forklift location. Each row shows four raw visual observations that the VLM
    abstracted into a single node despite significant viewpoint, lighting, and
    clutter variation. The domain hint provided to the VLM described high-level
    states as configurations differing in whether a pallet is currently loaded
    on the forklift, how many pallets remain in the scene, and the forklift's
    coarse location (loading bay vs.\ unloading bay). This abstraction capacity
    is key to reducing the number of nodes in the graph and ensuring that
    semantically equivalent observations are not split across multiple nodes.}
    \label{fig:gallery_merged}
\end{figure}

\begin{figure}[H]
    \centering
    \includegraphics[width=0.73\linewidth]{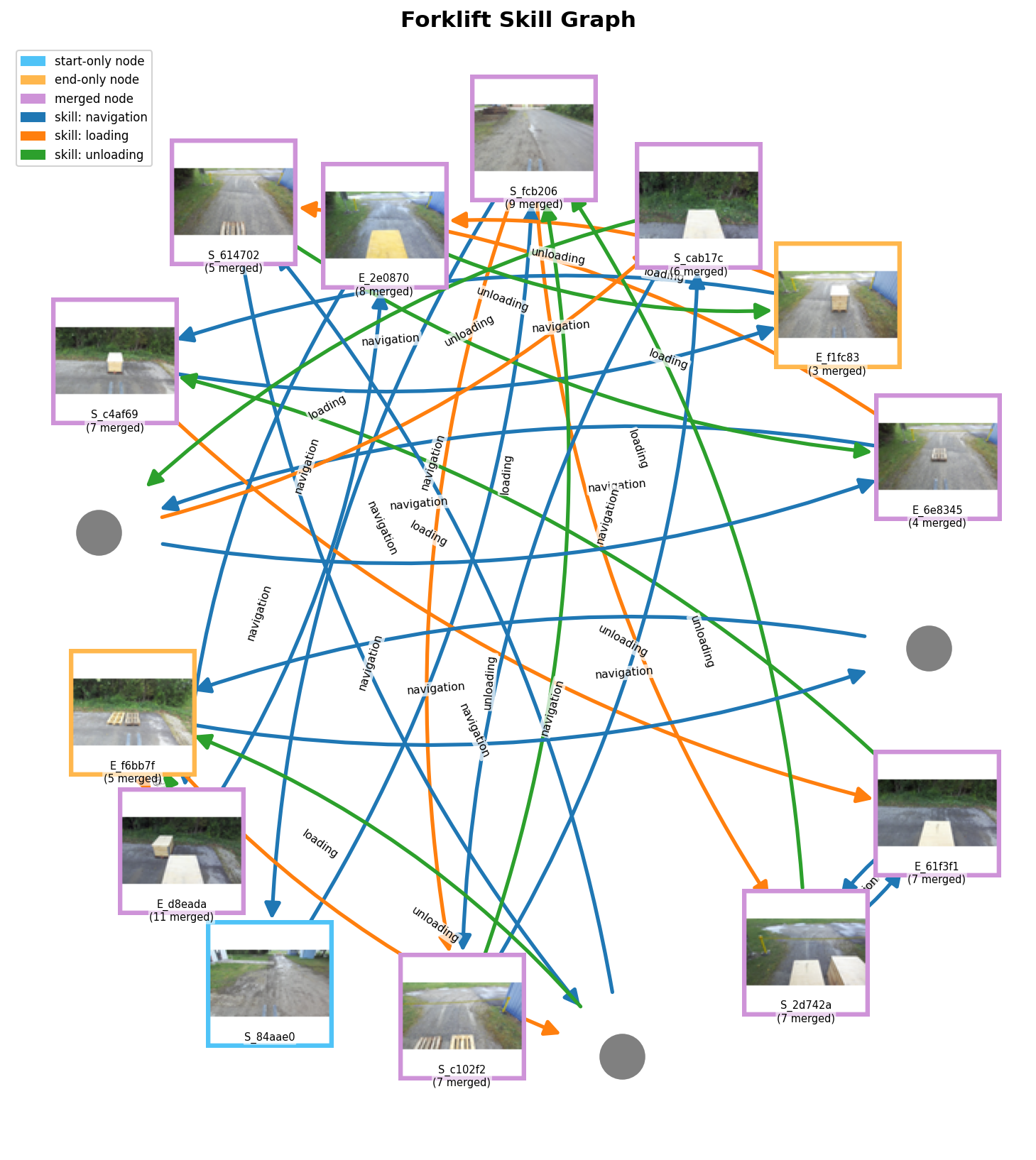}
    \caption{State-transition graph constructed by Gemini~3~Flash over 5
    demonstrations per skill. Nodes represent clusters of visually equivalent
    high-level states (boolean VLM matching); edges are annotated with
    single-word skill labels. Three nodes were manually completed to ensure
    full connectivity.}
    \label{fig:full}
\end{figure}

\begin{figure}[H]
    \centering
    \includegraphics[width=0.85\linewidth]{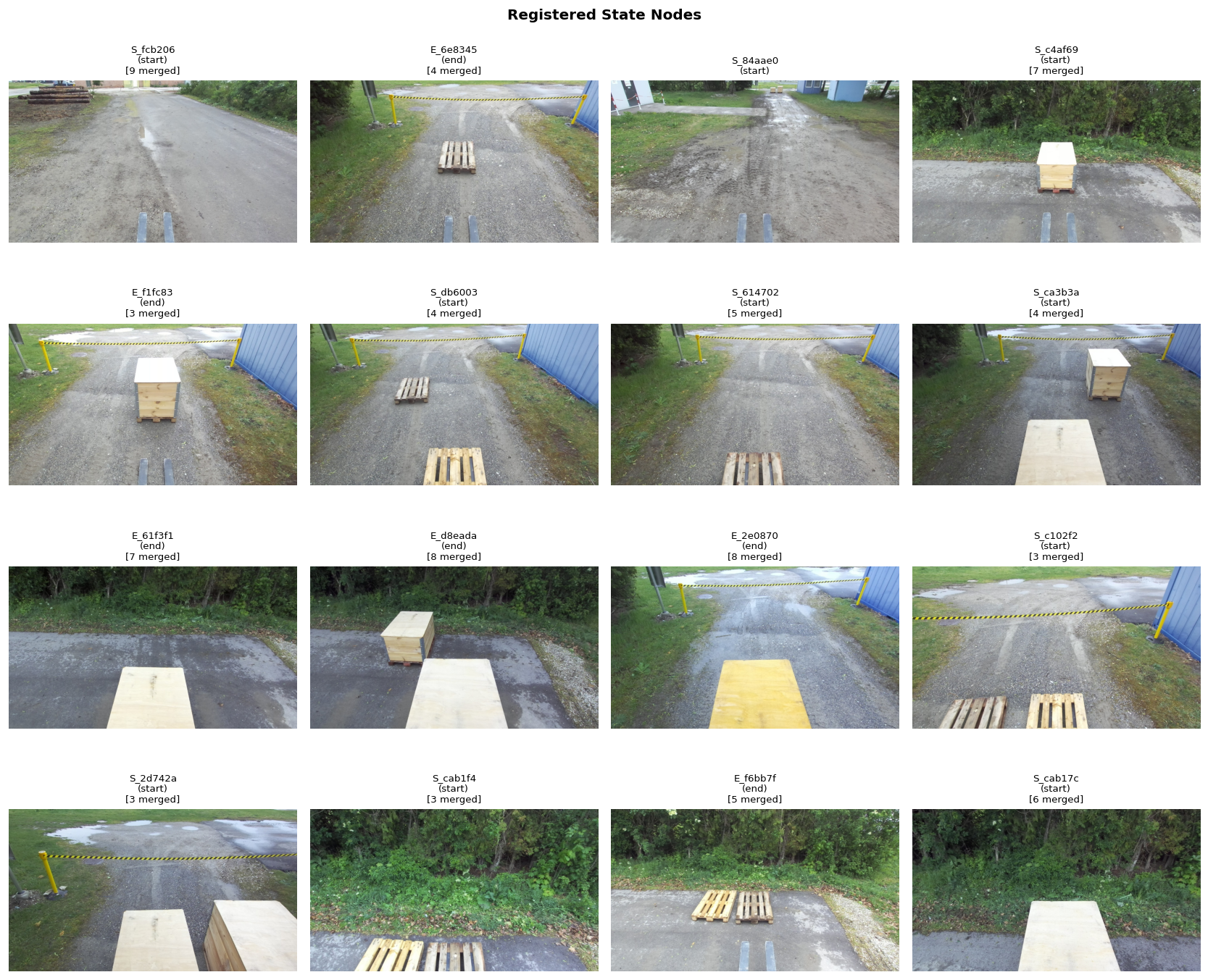}
    \caption{Gallery of raw visual observations grouped by node. Each cell
    shows the set of demonstration frames the VLM merged into a single abstract
    state, illustrating the degree of intra-node visual variability handled by
    the matching mechanism.}
    \label{fig:gallery}
\end{figure}

The following figures illustrate the VLM-based graph construction pipeline.
Fig.~\ref{fig:full} shows the full state-transition graph produced by
Gemini~3~Flash~\citep{google2025gemini3flash} over 5 demonstrations per skill
(\textit{navigate}, \textit{load}, \textit{unload}). Each node represents a
cluster of visually equivalent high-level states, determined by querying the
VLM with pairs of observations and collecting a boolean \texttt{true}/\texttt{false}
matching response. Edges are annotated with the single-word skill label
associated with the transition in the demonstration trajectory. Three nodes
were manually injected by a human operator to resolve ambiguities that the VLM
could not confidently resolve, ensuring full graph connectivity.
Figures~\ref{fig:gallery_merged} and~\ref{fig:gallery} provides a representative gallery of the raw visual
observations assigned to each node, illustrating the degree of intra-node
visual variability that the VLM is required to abstract over. 

\subsection{Gross ASP Outputs}

\noindent\makebox[\columnwidth]{\rule{0.175\columnwidth}{0.4pt}
\textbf{Forklift Multiple Pallets Storage Domain}
\rule{0.175\columnwidth}{0.4pt}}

\smallskip
The following action models are the raw symbolic outputs of the ASP solver
applied to the bisimulation-reduced state-transition graphs described above.
Three graph configurations are reported: the full graph, a graph with one node
and two edges missing, and a graph with two nodes and six edges missing. Small graph incompletion can induce action models becoming less
compact or noisier, requiring more action parameters, additional preconditions, and
finer-grained predicate structure, yet remain executable as planning domains.

\footnotesize
\setlength{\baselineskip}{1.3em}

\medskip
\noindent\begin{minipage}[t]{0.48\linewidth}
    \vspace{0pt}
    \includegraphics[width=\linewidth]{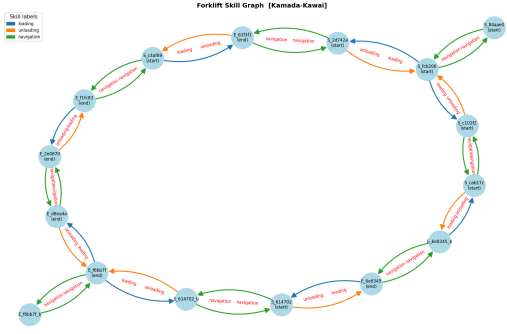}
    \captionof{figure}{Bisimulation reduction of the full state-transition
    graph fed to the ASP solver. Behaviourally equivalent nodes are merged
    into canonical representatives prior to induction.}
    \label{fig:bisim_full}
\end{minipage}
\hfill
\begin{minipage}[t]{0.48\linewidth}
    \vspace{0pt}
    \begin{mdframed}[linewidth=0.8pt, innerleftmargin=6pt, innerrightmargin=6pt,
                     innertopmargin=4pt, innerbottommargin=4pt]
    \textbf{Full Domain Graph}

    \smallskip
    \noindent\underline{MOVE: $a_1(x_1, x_2)$} \quad
        {\footnotesize Static: $\neg(x_1{=}x_2)$}\\
    \hspace*{1em} Pre: $p_2(x_1),\ p_3(x_2)$\\
    \hspace*{1em} Eff: $-\!p_2(x_1),\ -\!p_3(x_2),\ +p_3(x_1),\ +p_2(x_2)$

    \smallskip
    \noindent\underline{UNLOAD: $a_2(x_1, x_2)$}\\
    \hspace*{1em} Pre: $p_3(x_2),\ p_4(x_1)$\\
    \hspace*{1em} Eff: $-\!p_4(x_1),\ +p_1(),\ +p_5(x_1, x_2)$

    \smallskip
    \noindent\underline{LOAD: $a_3(x_1, x_2)$}\\
    \hspace*{1em} Pre: $p_1(),\ p_3(x_2),\ p_5(x_1, x_2)$\\
    \hspace*{1em} Eff: $-\!p_1(),\ -\!p_5(x_1, x_2),\ +p_4(x_1)$

    \smallskip
    \noindent\textbf{Objects:} $o_1, o_2, o_3, o_4$
    \textit{(2 pallets $\times$ 2 locations)}
    \end{mdframed}
\end{minipage}

\medskip
\begin{mdframed}[linewidth=0.8pt, innerleftmargin=6pt, innerrightmargin=6pt,
                 innertopmargin=4pt, innerbottommargin=4pt]

\textbf{One Node, Two Edges Missing}\\
\smallskip
\noindent\underline{MOVE: $a_1(x_1, x_2, x_3)$}\\ \quad
    {\footnotesize Static: $\neg(x_1{=}x_2),\ \neg(x_1{=}x_3)$}\\
\hspace*{1em} Pre: $p_3(x_2, x_1),\ p_3(x_2, x_2),\ p_4(x_3, x_1)$\\
\hspace*{1em} Eff: $-\!p_3(x_2, x_1),\ -\!p_3(x_2, x_2),\ +p_3(x_1, x_1),\ +p_3(x_1, x_2)$\\
\smallskip
\noindent\underline{UNLOAD: $a_2(x_1, x_2)$}\\ \quad
    {\footnotesize Static: $\neg(x_1{=}x_2)$}\\
\hspace*{1em} Pre: $p_2(x_1),\ p_3(x_1, x_1),\ p_3(x_2, x_2),\ p_4(x_2, x_2)$\\
\hspace*{1em} Eff: $-\!p_2(x_1),\ -\!p_3(x_2, x_2),\ +p_1(),\ +p_3(x_1, x_2)$\\
\smallskip
\noindent\underline{LOAD: $a_3(x_1, x_2)$}\\ \quad
    {\footnotesize Static: $\neg(x_1{=}x_2)$}\\
\hspace*{1em} Pre: $p_1(),\ p_3(x_2, x_1),\ p_3(x_2, x_2),\ p_4(x_1, x_1)$\\
\hspace*{1em} Eff: $-\!p_1(),\ -\!p_3(x_2, x_1),\ +p_2(x_2),\ +p_3(x_1, x_1)$\\
\smallskip
\noindent\textbf{Objects:} $o_1, o_2, o_3, o_4$
\textit{(2 pallets $\times$ 2 locations)}
\end{mdframed}

\medskip

\noindent\begin{minipage}[t]{0.38\linewidth}
    \vspace{0pt}
    \includegraphics[width=\linewidth]{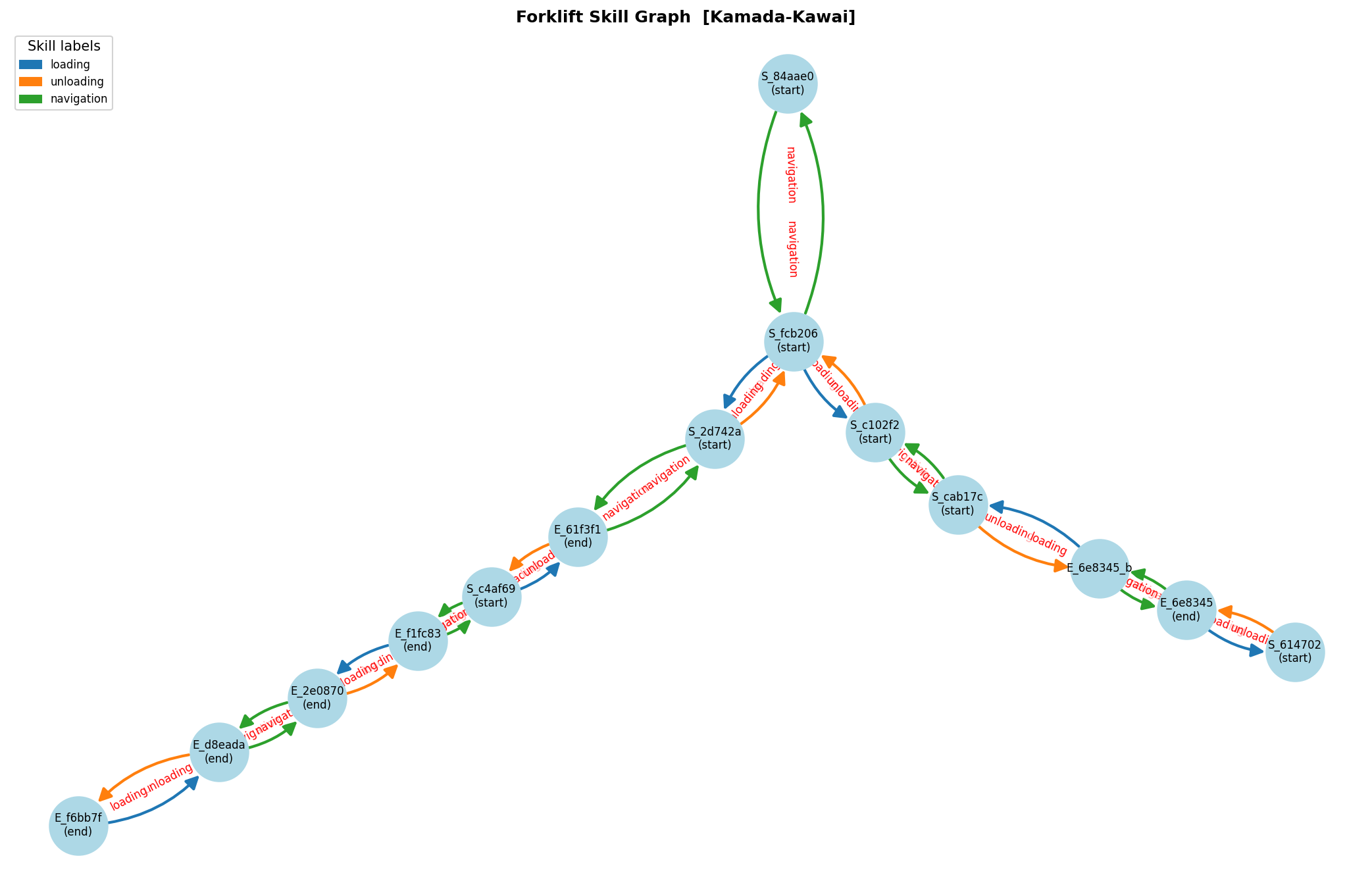}
    \captionof{figure}{Bisimulation reduction of the partial graph (two nodes,
    six edges missing). The reduced node set causes the solver to introduce
    additional parameters and finer-grained predicates.}
    \label{fig:bisim}
\end{minipage}
\hfill
\begin{minipage}[t]{0.59\linewidth}
    \vspace{0pt}
    \begin{mdframed}[linewidth=0.8pt, innerleftmargin=6pt, innerrightmargin=6pt,
                     innertopmargin=4pt, innerbottommargin=4pt]
    \textbf{Two Nodes, Six Edges Missing}\\
    \smallskip
    \noindent\underline{MOVE: $a_1(x_1, x_2, x_3)$}\\ \quad
        {\footnotesize Static: $\neg(x_1{=}x_2)$}\\
    \hspace*{1em} Pre: $p_1(x_1),\ p_2(x_3),$\\
    \hspace*{2.5em} $p_4(x_1, x_1),\ p_4(x_2, x_2),\ p_4(x_3, x_1)$\\
    \hspace*{1em} Eff: $-\!p_1(x_1),\ +p_1(x_2)$\\
    \smallskip
    \noindent\underline{UNLOAD: $a_2(x_1, x_2)$}\\ \quad
        {\footnotesize Static: $\neg(x_1{=}x_2)$}\\
    \hspace*{1em} Pre: $p_1(x_1),\ p_1(x_2),\ p_3(x_1),\ p_4(x_1, x_2)$\\
    \hspace*{1em} Eff: $-\!p_3(x_1),\ +p_2(x_2)$\\
    \smallskip
    \noindent\underline{LOAD: $a_3(x_1, x_2, x_3)$}\\ \quad
        {\footnotesize Static: $\neg(x_1{=}x_2),\ \neg(x_1{=}x_3),\ \neg(x_2{=}x_3)$}\\
    \hspace*{1em} Pre: $p_1(x_2),\ p_1(x_3),\ p_2(x_1),\ p_2(x_3),\ p_4(x_2, x_3)$\\
    \hspace*{1em} Eff: $-\!p_2(x_3),\ +p_3(x_2)$\\
    \smallskip
    \noindent\textbf{Objects:} $o_1, o_2, o_3, o_4$
    \textit{(2 pallets $\times$ 2 locations)}
    \end{mdframed}
\end{minipage}

\noindent\makebox[\columnwidth]{\rule{\columnwidth}{0.4pt}}

\setlength{\baselineskip}{\normalbaselineskip}
\normalsize
\bibliographystyle{plainnat} 
\bibliography{references}

@misc{Huemer2025ADAPT,
  author   = {Huemer, Johannes and Murschitz, Markus and Sch{\"o}rghuber, Matthias and Reisinger, Lukas and Kadiofsky, Thomas and Weidinger, Christoph and Niedermeyer, Mario and Widy, Benedikt and Zeilinger, Marcel and Beleznai, Csaba and Gl{\"u}ck, Tobias and Kugi, Andreas and Zips, Patrik},
  title    = {{{ADAPT}: An Autonomous Forklift for Construction Site Operation}},
  url      = {https://arxiv.org/abs/2503.14331},
  year     = {2025},
}

@article{hoffmann2003metric,
  title={The Metric-FF Planning System: Translating``Ignoring Delete Lists''to Numeric State Variables},
  author={Hoffmann, J{\"o}rg},
  journal={Journal of artificial intelligence research},
  volume={20},
  pages={291--341},
  year={2003}
}

@article{SUTTON1999181,
title = {Between MDPs and semi-MDPs: A framework for temporal abstraction in reinforcement learning},
journal = {Artificial Intelligence},
year = {1999},
issn = {0004-3702},
doi = {https://doi.org/10.1016/S0004-3702(99)00052-1},
author = {Richard S. Sutton and Doina Precup and Satinder Singh}
}

@inproceedings{peorl-Yang,
author = {Yang, Fangkai and Lyu, Daoming and Liu, Bo and Gustafson, Steven},
year = {2018},
month = {07},
pages = {4860-4866},
title = {PEORL: Integrating Symbolic Planning and Hierarchical Reinforcement Learning for Robust Decision-Making},
doi = {10.24963/ijcai.2018/675}
}

@InProceedings{Kokel2021,
title={RePReL: Integrating Relational Planning and Reinforcement Learning for Effective Abstraction},
booktitle={ICAPS},
author={Kokel, H and Manoharan, A and Natarajan, S and Ravindran, B and Tadepalli, P},
year={2021},
month={May}
}

@inproceedings{sarathy2021spotter,
  title={SPOTTER: Extending Symbolic Planning Operators through Targeted Reinforcement Learning},
  author={Sarathy, Vasanth and Kasenberg, Daniel and Goel, Shivam and Sinapov, Jivko and Scheutz, Matthias},
  booktitle={AAMAS},
  year={2021}
}

@inproceedings{goel2022rapidlearn,
  title={RAPid-Learn: A Framework for Learning to Recover for Handling Novelties in Open-World Environments},
  author={Goel, Shivam and Shukla, Yash and Sarathy, Vasanth and Scheutz, Matthias and Sinapov, Jivko},
  booktitle={IEEE ICDL},
  year={2022}
}

@misc{mcdermott_pddl_1998,
	title = {{PDDL} - {The} {Planning} {Domain} {Definition} {Language}},
	publisher = {The AIPS-98 Planning Competition Committee},
	author = {McDermott, Drew and Ghallab, Malik and Howe, Adele and Knoblock, Craig and Ram, Ashwin and Veloso, Manuela and Weld, Daniel and Wilkins, David},
	year = {1998}
}

@article{Konidaris_Kaelbling_Lozano-Perez_2018, title={From Skills to Symbols: Learning Symbolic Representations for Abstract High-Level Planning}, volume={61}, ISSN={1076-9757}, DOI={10.1613/jair.5575}, journal={Journal of Artificial Intelligence Research}, author={Konidaris, George and Kaelbling, Leslie Pack and Lozano-Perez, Tomas}, year={2018}, month=jan, pages={215–289}, language={en} }

@article{Icarte2020RewardME,
  title={Reward Machines: Exploiting Reward Function Structure in Reinforcement Learning},
  author={Rodrigo Toro Icarte and Toryn Q. Klassen and Richard Anthony Valenzano and Sheila A. McIlraith},
  journal={JAIR},
  year={2020},
  volume={73},
  pages={173-208}
}

@article{Zare_Kebria_Khosravi_Nahavandi_2024, title={A Survey of Imitation Learning: Algorithms, Recent Developments, and Challenges}, volume={54}, ISSN={2168-2275}, DOI={10.1109/TCYB.2024.3395626}, abstractNote={In recent years, the development of robotics and artificial intelligence (AI) systems has been nothing short of remarkable. As these systems continue to evolve, they are being utilized in increasingly complex and unstructured environments, such as autonomous driving, aerial robotics, and natural language processing. As a consequence, programming their behaviors manually or defining their behavior through the reward functions [as done in reinforcement learning (RL)] has become exceedingly difficult. This is because such environments require a high degree of flexibility and adaptability, making it challenging to specify an optimal set of rules or reward signals that can account for all the possible situations. In such environments, learning from an expert’s behavior through imitation is often more appealing. This is where imitation learning (IL) comes into play - a process where desired behavior is learned by imitating an expert’s behavior, which is provided through demonstrations.This article aims to provide an introduction to IL and an overview of its underlying assumptions and approaches. It also offers a detailed description of recent advances and emerging areas of research in the field. Additionally, this article discusses how researchers have addressed common challenges associated with IL and provides potential directions for future research. Overall, the goal of this article is to provide a comprehensive guide to the growing field of IL in robotics and AI.}, number={12}, journal={IEEE Transactions on Cybernetics}, author={Zare, Maryam and Kebria, Parham M. and Khosravi, Abbas and Nahavandi, Saeid}, year={2024}, month=dec, pages={7173–7186} }

@article{Osa_Pajarinen_Neumann_Bagnell_Abbeel_Peters_2018, title={An Algorithmic Perspective on Imitation Learning}, volume={7}, DOI={10.1561/2300000053}, abstractNote={As robots and other intelligent agents move from simple environments and problems to more complex, unstructured settings, manually programming their behavior has become increasingly challenging and expensive. Often, it is easier for a teacher to demonstrate a desired behavior rather than attempt to manually engineer it. This process of learning from demonstrations, and the study of algorithms to do so, is called imitation learning. This work provides an introduction to imitation learning. It covers the underlying assumptions, approaches, and how they relate; the rich set of algorithms developed to tackle the problem; and advice on effective tools and implementation. We intend this paper to serve two audiences. First, we want to familiarize machine learning experts with the challenges of imitation learning, particularly those arising in robotics, and the interesting theoretical and practical distinctions between it and more familiar frameworks like statistical supervised learning theory and reinforcement learning. Second, we want to give roboticists and experts in applied artificial intelligence a broader appreciation for the frameworks and tools available for imitation learning.}, journal={Foundations and Trends in Robotics}, author={Osa, Takayuki and Pajarinen, Joni and Neumann, Gerhard and Bagnell, J. and Abbeel, Pieter and Peters, Jan}, year={2018}, month=nov, pages={1–179} }

@article{Hussein_Gaber_Elyan_Jayne_2017, title={Imitation Learning: A Survey of Learning Methods}, volume={50}, ISSN={0360-0300}, DOI={10.1145/3054912}, abstractNote={Imitation learning techniques aim to mimic human behavior in a given task. An agent (a learning machine) is trained to perform a task from demonstrations by learning a mapping between observations and actions. The idea of teaching by imitation has been around for many years; however, the field is gaining attention recently due to advances in computing and sensing as well as rising demand for intelligent applications. The paradigm of learning by imitation is gaining popularity because it facilitates teaching complex tasks with minimal expert knowledge of the tasks. Generic imitation learning methods could potentially reduce the problem of teaching a task to that of providing demonstrations, without the need for explicit programming or designing reward functions specific to the task. Modern sensors are able to collect and transmit high volumes of data rapidly, and processors with high computational power allow fast processing that maps the sensory data to actions in a timely manner. This opens the door for many potential AI applications that require real-time perception and reaction such as humanoid robots, self-driving vehicles, human computer interaction, and computer games, to name a few. However, specialized algorithms are needed to effectively and robustly learn models as learning by imitation poses its own set of challenges. In this article, we survey imitation learning methods and present design options in different steps of the learning process. We introduce a background and motivation for the field as well as highlight challenges specific to the imitation problem. Methods for designing and evaluating imitation learning tasks are categorized and reviewed. Special attention is given to learning methods in robotics and games as these domains are the most popular in the literature and provide a wide array of problems and methodologies. We extensively discuss combining imitation learning approaches using different sources and methods, as well as incorporating other motion learning methods to enhance imitation. We also discuss the potential impact on industry, present major applications, and highlight current and future research directions.}, number={2}, journal={ACM Comput. Surv.}, author={Hussein, Ahmed and Gaber, Mohamed Medhat and Elyan, Eyad and Jayne, Chrisina}, year={2017}, month=apr, pages={21:1-21:35} }

@article{Manschitz_Kober_Gienger_Peters_2014, address={Chicago, IL, USA}, title={Learning to sequence movement primitives from demonstrations}, ISBN={9781479969340 9781479969319}, DOI={10.1109/IROS.2014.6943187}, abstractNote={We present an approach for learning sequential robot skills through kinesthetic teaching. The demonstrations are represented by a sequence graph. Finding the transitions between consecutive basic movements is treated as classification problem where both Support Vector Machines and Gaussian Mixture Models are evaluated as classifiers. We show how the observed primitive order of all demonstrations can help to improve the movement reproduction by restricting the classification outcome to the currently executed primitive and its possible successors in the graph. The approach is validated with an experiment in which a 7-DOF Barrett WAM robot learns to unscrew a light bulb.}, journal={2014 IEEE/RSJ International Conference on Intelligent Robots and Systems}, publisher={IEEE}, author={Manschitz, Simon and Kober, Jens and Gienger, Michael and Peters, Jan}, year={2014}, month=sep, pages={4414–4421} }

@inproceedings{Pertsch_Lee_Wu_Lim_2021, title={Demonstration-Guided Reinforcement Learning with Learned Skills}, abstractNote={Demonstration-guided reinforcement learning (RL) is a promising approach for learning complex behaviors by leveraging both reward feedback and a set of target task demonstrations. Prior approaches for demonstration-guided RL treat every new task as an independent learning problem and attempt to follow the provided demonstrations step-by-step, akin to a human trying to imitate a completely unseen behavior by following the demonstrator’s exact muscle movements. Naturally, such learning will be slow, but often new behaviors are not completely unseen: they share subtasks with behaviors we have previously learned. In this work, we aim to exploit this shared subtask structure to increase the efficiency of demonstration-guided RL. We first learn a set of reusable skills from large offline datasets of prior experience collected across many tasks. We then propose Skill-based Learning with Demonstrations (SkiLD), an algorithm for demonstration-guided RL that efficiently leverages the provided demonstrations by following the demonstrated skills instead of the primitive actions, resulting in substantial performance improvements over prior demonstration-guided RL approaches. We validate the effectiveness of our approach on long-horizon maze navigation and complex robot manipulation tasks.}, author={Pertsch, Karl and Lee, Youngwoon and Wu, Yue and Lim, Joseph J.}, year={2021}, month=jul }

@article{Tanwani_Yan_Lee_Calinon_Goldberg_2021, title={Sequential robot imitation learning from observations}, volume={40}, ISSN={0278-3649}, DOI={10.1177/02783649211032721}, abstractNote={This paper presents a framework to learn the sequential structure in the demonstrations for robot imitation learning. We first present a family of task-parameterized hidden semi-Markov models that extracts invariant segments (also called sub-goals or options) from demonstrated trajectories, and optimally follows the sampled sequence of states from the model with a linear quadratic tracking controller. We then extend the concept to learning invariant segments from visual observations that are sequenced together for robot imitation. We present Motion2Vec that learns a deep embedding space by minimizing a metric learning loss in a Siamese network: images from the same action segment are pulled together while being pushed away from randomly sampled images of other segments, and a time contrastive loss is used to preserve the temporal ordering of the images. The trained embeddings are segmented with a recurrent neural network, and subsequently used for decoding the end-effector pose of the robot. We first show its application to a pick-and-place task with the Baxter robot while avoiding a moving obstacle from four kinesthetic demonstrations only, followed by suturing task imitation from publicly available suturing videos of the JIGSAWS dataset with state-of-the-art 
85.5
% segmentation accuracy and 
0.94
 cm error in position per observation on the test set.}, number={10–11}, journal={The International Journal of Robotics Research}, publisher={SAGE Publications Ltd STM}, author={Tanwani, Ajay Kumar and Yan, Andy and Lee, Jonathan and Calinon, Sylvain and Goldberg, Ken}, year={2021}, month=sep, pages={1306–1325}, language={en} }

@article{Zhu_Stone_Zhu_2022, title={Bottom-Up Skill Discovery From Unsegmented Demonstrations for Long-Horizon Robot Manipulation}, volume={7}, rights={https://ieeexplore.ieee.org/Xplorehelp/downloads/license-information/IEEE.html}, ISSN={2377-3766, 2377-3774}, DOI={10.1109/LRA.2022.3146589}, abstractNote={We tackle real-world long-horizon robot manipulation tasks through skill discovery. We present a bottom-up approach to learning a library of reusable skills from unsegmented demonstrations and use these skills to synthesize prolonged robot behaviors. Our method starts with constructing a hierarchical task structure from each demonstration through agglomerative clustering. From the task structures of multi-task demonstrations, we identify skills based on the recurring patterns and train goal-conditioned sensorimotor policies with hierarchical imitation learning. Finally, we train a meta controller to compose these skills to solve long-horizon manipulation tasks. The entire model can be trained on a small set of human demonstrations collected within 30 minutes without further annotations, making it amendable to real-world deployment. We systematically evaluated our method in simulation environments and on a real robot. Our method has shown superior performance over state-of-the-art imitation learning methods in multi-stage manipulation tasks. Furthermore, skills discovered from multi-task demonstrations boost the average task success by 8% compared to those discovered from individual tasks.}, number={2}, journal={IEEE Robotics and Automation Letters}, author={Zhu, Yifeng and Stone, Peter and Zhu, Yuke}, year={2022}, month=apr, pages={4126–4133} }

@article{Wolfe_Marthi_Russell_2010, title={Combined Task and Motion Planning for Mobile Manipulation}, volume={20}, rights={Copyright (c) 2021 Proceedings of the International Conference on Automated Planning and Scheduling}, ISSN={2334-0843}, DOI={10.1609/icaps.v20i1.13436}, abstractNote={We present a hierarchical planning system and its application to robotic manipulation.&nbsp; The novel features of the system are: 1) it finds high-quality kinematic solutions to task-level problems; 2) it takes advantage of subtask-specific irrelevance information, reusing optimal solutions to state-abstracted subproblems across the search space.&nbsp; We briefly describe how the system handles uncertainty during plan execution, and present results on discrete problems as well as pick-and-place tasks for a mobile robot.}, journal={Proceedings of the International Conference on Automated Planning and Scheduling}, author={Wolfe, Jason and Marthi, Bhaskara and Russell, Stuart}, year={2010}, month=may, pages={254–257}, language={en} }

@inproceedings{Kaelbling_Lozano-Perez_2011, title={Hierarchical task and motion planning in the now}, ISSN={1050-4729}, url={https://ieeexplore.ieee.org/abstract/document/5980391/}, DOI={10.1109/ICRA.2011.5980391}, abstractNote={In this paper we outline an approach to the integration of task planning and motion planning that has the following key properties: It is aggressively hierarchical; it makes choices and commits to them in a top-down fashion in an attempt to limit the length of plans that need to be constructed, and thereby exponentially decrease the amount of search required. It operates on detailed, continuous geometric representations and does not require a-priori discretization of the state or action spaces.}, booktitle={2011 IEEE International Conference on Robotics and Automation}, author={Kaelbling, Leslie Pack and Lozano-Pérez, Tomás}, year={2011}, month=may, pages={1470–1477} }

@article{Garrett_Chitnis_Holladay_Kim_Silver_Kaelbling_Lozano-Perez_2021, title={Integrated Task and Motion Planning}, volume={4}, ISSN={2573-5144}, DOI={10.1146/annurev-control-091420-084139}, abstractNote={The problem of planning for a robot that operates in environments containing a large number of objects, taking actions to move itself through the world as well as to change the state of the objects, is known as task and motion planning (TAMP). TAMP problems contain elements of discrete task planning, discrete–continuous mathematical programming, and continuous motion planning and thus cannot be effectively addressed by any of these fields directly. In this article, we define a class of TAMP problems and survey algorithms for solving them, characterizing the solution methods in terms of their strategies for solving the continuous-space subproblems and their techniques for integrating the discrete and continuous components of the search.}, number={Volume 4, 2021}, journal={Annual Review of Control, Robotics, and Autonomous Systems}, publisher={Annual Reviews}, author={Garrett, Caelan Reed and Chitnis, Rohan and Holladay, Rachel and Kim, Beomjoon and Silver, Tom and Kaelbling, Leslie Pack and Lozano-Pérez, Tomás}, year={2021}, month=may, pages={265–293}, language={en} }

@inproceedings{Loula_Allen_Silver_Tenenbaum_2020,
  title     = {Learning Constraint-Based Planning Models from Demonstrations},
  booktitle = {2020 IEEE/RSJ International Conference on Intelligent Robots
               and Systems (IROS)},
  author    = {Loula, Joao and Allen, Kelsey and Silver, Tom and Tenenbaum, Josh},
  publisher = {IEEE},
  address   = {Las Vegas, NV, USA},
  pages     = {5410--5416},
  year      = {2020},
  url       = {https://ieeexplore.ieee.org/abstract/document/9341535/}
}

@inbook{Bonet_Geffner_2020,
  title     = {Learning First-Order Symbolic Representations for Planning
               from the Structure of the State Space},
  booktitle = {ECAI 2020 -- 24th European Conference on Artificial Intelligence},
  series    = {Frontiers in Artificial Intelligence and Applications},
  author    = {Bonet, Blai and Geffner, Hector},
  publisher = {IOS Press},
  address   = {Santiago de Compostela, Spain},
  pages     = {2322--2329},
  year      = {2020},
  doi       = {10.3233/FAIA200361},
  url       = {https://doi.org/10.3233/FAIA200361}
}

@inproceedings{kr2021rbrg,
    title     = {{Learning First-Order Representations for Planning from Black Box States: New Results}},
    author    = {Rodriguez, Ivan D. and Bonet, Blai and Romero, Javier and Geffner, Hector},
    booktitle = {{Proceedings of the 18th International Conference on Principles of Knowledge Representation and Reasoning}},
    pages     = {539--548},
    year      = {2021},
    month     = {11},
    doi       = {10.24963/kr.2021/51},
    url       = {https://doi.org/10.24963/kr.2021/51},
  }

@article{Ahmetoglu_Seker_Piater_Oztop_Ugur_2022, title={DeepSym: Deep Symbol Generation and Rule Learning from Unsupervised Continuous Robot Interaction for Planning}, volume={75}, ISSN={1076-9757}, DOI={10.1613/jair.1.13754}, abstractNote={Symbolic planning and reasoning are powerful tools for robots tackling complex tasks. However, the need to manually design the symbols restrict their applicability, especially for robots that are expected to act in open-ended environments. Therefore symbol formation and rule extraction should be considered part of robot learning, which, when done properly, will oﬀer scalability, ﬂexibility, and robustness. Towards this goal, we propose a novel general method that ﬁnds action-grounded, discrete object and eﬀect categories and builds probabilistic rules over them for non-trivial action planning. Our robot interacts with objects using an initial action repertoire that is assumed to be acquired earlier and observes the eﬀects it can create in the environment. To form action-grounded object, eﬀect, and relational categories, we employ a binary bottleneck layer in a predictive, deep encoderdecoder network that takes the image of the scene and the action applied as input, and generates the resulting eﬀects in the scene in pixel coordinates. After learning, the binary latent vector represents action-driven object categories based on the interaction experience of the robot. To distill the knowledge represented by the neural network into rules useful for symbolic reasoning, a decision tree is trained to reproduce its decoder function. Probabilistic rules are extracted from the decision paths of the tree and are represented in the Probabilistic Planning Domain Deﬁnition Language (PPDDL), allowing oﬀ-the-shelf planners to operate on the knowledge extracted from the sensorimotor experience of the robot. The deployment of the proposed approach for a simulated robotic manipulator enabled the discovery of discrete representations of object properties such as ‘rollable’ and ‘insertable’. In turn, the use of these representations as symbols allowed the generation of eﬀective plans for achieving goals, such as building towers of the desired height, demonstrating the eﬀectiveness of the approach for multi-step object manipulation. Finally, we demonstrate that the system is not only restricted to the robotics domain by assessing its applicability to the MNIST 8-puzzle domain in which learned symbols allow for the generation of plans that move the empty tile into any given position.}, note={arXiv:2012.02532 [cs]}, journal={Journal of Artificial Intelligence Research}, author={Ahmetoglu, Alper and Seker, M. Yunus and Piater, Justus and Oztop, Erhan and Ugur, Emre}, year={2022}, month=nov, pages={709–745}, language={en} }

@inproceedings{Silver_Chitnis_Kumar_McClinton_Lozano-Perez_Kaelbling_Tenenbaum_2022, title={Predicate Invention for Bilevel Planning}, url={http://arxiv.org/abs/2203.09634}, abstractNote={Efﬁcient planning in continuous state and action spaces is fundamentally hard, even when the transition model is deterministic and known. One way to alleviate this challenge is to perform bilevel planning with abstractions, where a highlevel search for abstract plans is used to guide planning in the original transition space. Previous work has shown that when state abstractions in the form of symbolic predicates are hand-designed, operators and samplers for bilevel planning can be learned from demonstrations. In this work, we propose an algorithm for learning predicates from demonstrations, eliminating the need for manually speciﬁed state abstractions. Our key idea is to learn predicates by optimizing a surrogate objective that is tractable but faithful to our real efﬁcient-planning objective. We use this surrogate objective in a hill-climbing search over predicate sets drawn from a grammar. Experimentally, we show across four robotic planning environments that our learned abstractions are able to quickly solve held-out tasks, outperforming six baselines.}, note={arXiv:2203.09634 [cs]}, number={arXiv:2203.09634}, publisher={arXiv}, author={Silver, Tom and Chitnis, Rohan and Kumar, Nishanth and McClinton, Willie and Lozano-Perez, Tomas and Kaelbling, Leslie Pack and Tenenbaum, Joshua}, year={2022}, month=nov, language={en} }

@inproceedings{Kumar_McClinton_Chitnis_Silver_Lozano-Pérez_Kaelbling_2023, title={Learning Efficient Abstract Planning Models that Choose What to Predict}, url={http://arxiv.org/abs/2208.07737}, abstractNote={An effective approach to solving long-horizon tasks in robotics domains with continuous state and action spaces is bilevel planning, wherein a highlevel search over an abstraction of an environment is used to guide low-level decision-making. Recent work has shown how to enable such bilevel planning by learning abstract models in the form of symbolic operators and neural samplers. In this work, we show that existing symbolic operator learning approaches fall short in many robotics domains where a robot’s actions tend to cause a large number of irrelevant changes in the abstract state. This is primarily because they attempt to learn operators that exactly predict all observed changes in the abstract state. To overcome this issue, we propose to learn operators that ‘choose what to predict’ by only modelling changes necessary for abstract planning to achieve specified goals. Experimentally, we show that our approach learns operators that lead to efficient planning across 10 different hybrid robotics domains, including 4 from the challenging BEHAVIOR-100 benchmark, while generalizing to novel initial states, goals, and objects.}, note={arXiv:2208.07737 [cs]}, number={arXiv:2208.07737}, publisher={arXiv}, author={Kumar, Nishanth and McClinton, Willie and Chitnis, Rohan and Silver, Tom and Lozano-Pérez, Tomás and Kaelbling, Leslie Pack}, year={2023}, month=sep, language={en} }

@inproceedings{Shah_Nagpal_Verma_Srivastava_2024, title={From Reals to Logic and Back: Inventing Symbolic Vocabularies, Actions, and Models for Planning from Raw Data}, url={http://arxiv.org/abs/2402.11871}, DOI={10.48550/arXiv.2402.11871}, abstractNote={Hand-crafted, logic-based state and action representations have been widely used to overcome the intractable computational complexity of long-horizon robot planning problems, including task and motion planning problems. However, creating such representations requires experts with strong intuitions and detailed knowledge about the robot and the tasks it may need to accomplish in a given setting. Removing this dependency on human intuition is a highly active research area.}, note={arXiv:2402.11871 [cs]}, number={arXiv:2402.11871}, publisher={arXiv}, author={Shah, Naman and Nagpal, Jayesh and Verma, Pulkit and Srivastava, Siddharth}, year={2024}, month=mar, language={en} }

@inproceedings{Umili_Antonioni_Riccio_Capobianco_Nardi_Giacomo, title={Learning a Symbolic Planning Domain through the Interaction with Continuous Environments}, author={Umili, Elena and Antonioni, Emanuele and Riccio, Francesco and Capobianco, Roberto and Nardi, Daniele and Giacomo, Giuseppe De}, language={en}, year={2021} }

@article{Cheng_Xu_2023, title={LEAGUE: Guided Skill Learning and Abstraction for Long-Horizon Manipulation}, volume={8}, ISSN={2377-3766}, DOI={10.1109/LRA.2023.3308061}, abstractNote={To assist with everyday human activities, robots must solve complex long-horizon tasks and generalize to new settings. Recent deep reinforcement learning (RL) methods show promise in fully autonomous learning, but they struggle to reach long-term goals in large environments. On the other hand, Task and Motion Planning (TAMP) approaches excel at solving and generalizing across long-horizon tasks, thanks to their powerful state and action abstractions. But they assume predefined skill sets, which limits their real-world applications. In this work, we combine the benefits of these two paradigms and propose an integrated task planning and skill learning framework named LEAGUE (Learning and Abstraction with Guidance). LEAGUE leverages the symbolic interface of a task planner to guide RL-based skill learning and creates abstract state space to enable skill reuse. More importantly, LEAGUE learns manipulation skills in-situ of the task planning system, continuously growing its capability and the set of tasks that it can solve. We evaluate LEAGUE on four challenging simulated task domains and show that LEAGUE outperforms baselines by large margins. We also show that the learned skills can be reused to accelerate learning in new tasks domains and transfer to a physical robot platform.}, number={10}, journal={IEEE Robotics and Automation Letters}, author={Cheng, Shuo and Xu, Danfei}, year={2023}, month=oct, pages={6451–6458} }

@inproceedings{Silver_Athalye_Tenenbaum_Lozano-Perez_Kaelbling_2023, title={Learning Neuro-Symbolic Skills for Bilevel Planning}, ISSN={2640-3498}, url={https://proceedings.mlr.press/v205/silver23a.html}, abstractNote={Decision-making is challenging in robotics environments with continuous object-centric states, continuous actions, long horizons, and sparse feedback. Hierarchical approaches, such as task and motion planning (TAMP), address these challenges by decomposing decision-making into two or more levels of abstraction. In a setting where demonstrations and symbolic predicates are given, prior work has shown how to learn symbolic operators and neural samplers for TAMP with manually designed parameterized policies. Our main contribution is a method for learning parameterized polices in combination with operators and samplers. These components are packaged into modular neuro-symbolic skills and sequenced together with search-then-sample TAMP to solve new tasks. In experiments in four robotics domains, we show that our approach — bilevel planning with neuro-symbolic skills — can solve a wide range of tasks with varying initial states, goals, and objects, outperforming six baselines and ablations.}, booktitle={Proceedings of The 6th Conference on Robot Learning}, publisher={PMLR},address={Auckland, New Zealand}, author={Silver, Tom and Athalye, Ashay and Tenenbaum, Joshua B. and Lozano-Pérez, Tomás and Kaelbling, Leslie Pack}, year={2023}, month=mar, pages={701–714}, language={en} }

@article{Illanes_Yan_Icarte_McIlraith_2020, title={Symbolic Plans as High-Level Instructions for Reinforcement Learning}, volume={30}, rights={Copyright (c) 2020 Association for the Advancement of Artificial Intelligence}, ISSN={2334-0843}, DOI={10.1609/icaps.v30i1.6750}, abstractNote={Reinforcement learning (RL) agents seek to maximize the cumulative reward obtained when interacting with their environment. Users define tasks or goals for RL agents by designing specialized reward functions such that maximization aligns with task satisfaction. This work explores the use of high-level symbolic action models as a framework for defining final-state goal tasks and automatically producing their corresponding reward functions. We also show how automated planning can be used to synthesize high-level plans that can guide hierarchical RL (HRL) techniques towards efficiently learning adequate policies. We provide a formal characterization of taskable RL environments and describe sufficient conditions that guarantee we can satisfy various notions of optimality (e.g., minimize total cost, maximize probability of reaching the goal). In addition, we do an empirical evaluation that shows that our approach converges to near-optimal solutions faster than standard RL and HRL methods and that it provides an effective framework for transferring learned skills across multiple tasks in a given environment.}, journal={Proceedings of the International Conference on Automated Planning and Scheduling}, author={Illanes, León and Yan, Xi and Icarte, Rodrigo Toro and McIlraith, Sheila A.}, year={2020}, month=jun, pages={540–550}, language={en} }

@inproceedings{Lorang_Goel_Shukla_Zips_Scheutz_2024, title={A Framework for Neurosymbolic Goal-Conditioned Continual Learning in Open World Environments}, ISSN={2153-0866}, url={https://ieeexplore.ieee.org/document/10801627}, DOI={10.1109/IROS58592.2024.10801627}, abstractNote={In dynamic open-world environments, agents continually face new challenges due to sudden and unpredictable novelties, hindering Task and Motion Planning (TAMP) in autonomous systems. We introduce a novel TAMP architecture that integrates symbolic planning with reinforcement learning to enable autonomous adaptation in such environments, operating without human guidance. Our approach employs symbolic goal representation within a goal-oriented learning framework, coupled with planner-guided goal identification, effectively managing abrupt changes where traditional reinforcement learning, re-planning, and hybrid methods fall short. Through sequential novelty injections in our experiments, we assess our method’s adaptability to continual learning scenarios. Extensive simulations conducted in a robotics domain corroborate the superiority of our approach, demonstrating faster convergence to higher performance compared to traditional methods. The success of our framework in navigating diverse novelty scenarios within a continuous domain underscores its potential for critical real-world applications.}, booktitle={2024 IEEE/RSJ International Conference on Intelligent Robots and Systems (IROS)}, author={Lorang, Pierrick and Goel, Shivam and Shukla, Yash and Zips, Patrik and Scheutz, Matthias}, year={2024}, month=oct, pages={12070–12077} }

@inproceedings{Lorang_Horvath_Kietreiber_Zips_Heitzinger_Scheutz_2024, title={Adapting to the “Open World”: The Utility of Hybrid Hierarchical Reinforcement Learning and Symbolic Planning}, url={https://ieeexplore.ieee.org/document/10611594}, DOI={10.1109/ICRA57147.2024.10611594}, abstractNote={Open-world robotic tasks such as autonomous driving pose significant challenges to robot control due to unknown and unpredictable events that disrupt task performance. Neural network-based reinforcement learning (RL) techniques (like DQN, PPO, SAC, etc.) struggle to adapt in large domains and suffer from catastrophic forgetting. Hybrid planning and RL approaches have shown some promise in handling environmental changes but lack efficiency in accommodation speed. To address this limitation, we propose an enhanced hybrid system with a nested hierarchical action abstraction that can utilize previously acquired skills to effectively tackle unexpected novelties. We show that it can adapt faster and generalize better compared to state-of-the-art RL and hybrid approaches, significantly improving robustness when multiple environmental changes occur at the same time.}, booktitle={2024 IEEE International Conference on Robotics and Automation (ICRA)}, author={Lorang, Pierrick and Horvath, Helmut and Kietreiber, Tobias and Zips, Patrik and Heitzinger, Clemens and Scheutz, Matthias}, year={2024}, month=may, pages={508–514} }

@inproceedings{Guan_Sreedharan_Kambhampati, title={Leveraging Approximate Symbolic Models for Reinforcement Learning via Skill Diversity}, abstractNote={Creating reinforcement learning (RL) agents that are capable of accepting and leveraging taskspecific knowledge from humans has been long identified as a possible strategy for developing scalable approaches for solving long-horizon problems. While previous works have looked at the possibility of using symbolic models along with RL approaches, they tend to assume that the high-level action models are executable at low level and the fluents can exclusively characterize all desirable MDP states. Symbolic models of real world tasks are however often incomplete. To this end, we introduce Approximate Symbolic-Model Guided Reinforcement Learning, wherein we will formalize the relationship between the symbolic model and the underlying MDP that will allow us to characterize the incompleteness of the symbolic model. We will use these models to extract highlevel landmarks that will be used to decompose the task. At the low level, we learn a set of diverse policies for each possible task subgoal identified by the landmark, which are then stitched together. We evaluate our system by testing on three different benchmark domains and show how even with incomplete symbolic model information, our approach is able to discover the task structure and efficiently guide the RL agent towards the goal.}, author={Guan, Lin and Sreedharan, Sarath and Kambhampati, Subbarao}, language={en}, year={2022} }

@inproceedings{Rajaraman_Yang_Jiao_Ramachandran_2020, title={Toward the Fundamental Limits of Imitation Learning}, url={http://arxiv.org/abs/2009.05990}, DOI={10.48550/arXiv.2009.05990}, abstractNote={Imitation learning (IL) aims to mimic the behavior of an expert policy in a sequential decision-making problem given only demonstrations. In this paper, we focus on understanding the minimax statistical limits of IL in episodic Markov Decision Processes (MDPs). We first consider the setting where the learner is provided a dataset of $N$ expert trajectories ahead of time, and cannot interact with the MDP. Here, we show that the policy which mimics the expert whenever possible is in expectation $lesssim frac{|mathcal{S}| H^2 log (N)}{N}$ suboptimal compared to the value of the expert, even when the expert follows an arbitrary stochastic policy. Here $mathcal{S}$ is the state space, and $H$ is the length of the episode. Furthermore, we establish a suboptimality lower bound of $gtrsim |mathcal{S}| H^2 / N$ which applies even if the expert is constrained to be deterministic, or if the learner is allowed to actively query the expert at visited states while interacting with the MDP for $N$ episodes. To our knowledge, this is the first algorithm with suboptimality having no dependence on the number of actions, under no additional assumptions. We then propose a novel algorithm based on minimum-distance functionals in the setting where the transition model is given and the expert is deterministic. The algorithm is suboptimal by $lesssim min { H sqrt{|mathcal{S}| / N} , |mathcal{S}| H^{3/2} / N }$, showing that knowledge of transition improves the minimax rate by at least a $sqrt{H}$ factor.}, note={arXiv:2009.05990 [cs]}, number={arXiv:2009.05990}, publisher={arXiv}, author={Rajaraman, Nived and Yang, Lin F. and Jiao, Jiantao and Ramachandran, Kannan}, year={2020}, month=sep }

@inproceedings{Xu_Li_Yu_2020, title={Error Bounds of Imitating Policies and Environments}, url={http://arxiv.org/abs/2010.11876}, DOI={10.48550/arXiv.2010.11876}, abstractNote={Imitation learning trains a policy by mimicking expert demonstrations. Various imitation methods were proposed and empirically evaluated, meanwhile, their theoretical understanding needs further studies. In this paper, we firstly analyze the value gap between the expert policy and imitated policies by two imitation methods, behavioral cloning and generative adversarial imitation. The results support that generative adversarial imitation can reduce the compounding errors compared to behavioral cloning, and thus has a better sample complexity. Noticed that by considering the environment transition model as a dual agent, imitation learning can also be used to learn the environment model. Therefore, based on the bounds of imitating policies, we further analyze the performance of imitating environments. The results show that environment models can be more effectively imitated by generative adversarial imitation than behavioral cloning, suggesting a novel application of adversarial imitation for model-based reinforcement learning. We hope these results could inspire future advances in imitation learning and model-based reinforcement learning.}, note={arXiv:2010.11876 [cs]}, number={arXiv:2010.11876}, publisher={arXiv}, author={Xu, Tian and Li, Ziniu and Yu, Yang}, year={2020}, month=oct }

@article{Teng_Chen_Ai_Zhou_Xuanyuan_Hu_2023, title={Hierarchical Interpretable Imitation Learning for End-to-End Autonomous Driving}, volume={8}, ISSN={2379-8904}, DOI={10.1109/TIV.2022.3225340}, abstractNote={End-to-end autonomous driving provides a simple and efficient framework for autonomous driving systems, which can directly obtain control commands from raw perception data. However, it fails to address stability and interpretability problems in complex urban scenarios. In this paper, we construct a two-stage end-to-end autonomous driving model for complex urban scenarios, named HIIL (Hierarchical Interpretable Imitation Learning), which integrates interpretable BEV mask and steering angle to solve the problems shown above. In Stage One, we propose a pretrained Bird’s Eye View (BEV) model which leverages a BEV mask to present an interpretation of the surrounding environment. In Stage Two, we construct an Interpretable Imitation Learning (IIL) model that fuses BEV latent feature from Stage One with an additional steering angle from Pure-Pursuit (PP) algorithm. In the HIIL model, visual information is converted to semantic images by the semantic segmentation network, and the semantic images are encoded to extract the BEV latent feature, which are decoded to predict BEV masks and fed to the IIL as perception data. In this way, the BEV latent feature bridges the BEV and IIL models. Visual information can be supplemented by the calculated steering angle for PP algorithm, speed vector, and location information, thus it could have better performance in complex and terrible scenarios. Our HIIL model meets an urgent requirement for interpretability and robustness of autonomous driving. We validate the proposed model in the CARLA simulator with extensive experiments which show remarkable interpretability, generalization, and robustness capability in unknown scenarios for navigation tasks.}, number={1}, journal={IEEE Transactions on Intelligent Vehicles}, author={Teng, Siyu and Chen, Long and Ai, Yunfeng and Zhou, Yuanye and Xuanyuan, Zhe and Hu, Xuemin}, year={2023}, month=jan, pages={673–683} }

@inproceedings{Le_Jiang_Agarwal_Dudik_Yue_Hal_Daume_2018,
  title     = {Hierarchical Imitation and Reinforcement Learning},
  booktitle = {Proceedings of the 35th International Conference on Machine Learning},
  author    = {Le, Hoang and Jiang, Nan and Agarwal, Alekh and Dudik, Miroslav
               and Yue, Yisong and {Hal Daum\'{e} III}},
  publisher = {PMLR},
  address   = {Stockholm, Sweden},
  pages     = {2917--2926},
  year      = {2018},
  month     = jul,
  issn      = {2640-3498},
  url       = {https://proceedings.mlr.press/v80/le18a.html}
}

@article{Gehring_Asai_Chitnis_Silver_Kaelbling_Sohrabi_Katz_2022, title={Reinforcement Learning for Classical Planning: Viewing Heuristics as Dense Reward Generators}, volume={32}, DOI={10.1609/icaps.v32i1.19846}, number={1}, journal={Proceedings of the International Conference on Automated Planning and Scheduling}, author={Gehring, Clement and Asai, Masataro and Chitnis, Rohan and Silver, Tom and Kaelbling, Leslie and Sohrabi, Shirin and Katz, Michael}, year={2022}, month=jun, pages={588–596} }

@article{Mitchener_Tuckey_Crosby_Russo_2022, title={Detect, Understand, Act: A Neuro-symbolic Hierarchical Reinforcement Learning Framework}, volume={111}, ISSN={1573-0565}, DOI={10.1007/s10994-022-06142-7}, abstractNote={In this paper we introduce Detect, Understand, Act (DUA), a neuro-symbolic reinforcement learning framework. The Detect component is composed of a traditional computer vision object detector and tracker. The Act component houses a set of options, high-level actions enacted by pre-trained deep reinforcement learning (DRL) policies. The Understand component provides a novel answer set programming (ASP) paradigm for symbolically implementing a meta-policy over options and effectively learning it using inductive logic programming (ILP). We evaluate our framework on the Animal-AI (AAI) competition testbed, a set of physical cognitive reasoning problems. Given a set of pre-trained DRL policies, DUA requires only a few examples to learn a meta-policy that allows it to improve the state-of-the-art on multiple of the most challenging categories from the testbed. DUA constitutes the first holistic hybrid integration of computer vision, ILP and DRL applied to an AAI-like environment and sets the foundations for further use of ILP in complex DRL challenges.}, number={4}, journal={Machine Learning}, author={Mitchener, Ludovico and Tuckey, David and Crosby, Matthew and Russo, Alessandra}, year={2022}, month=apr, pages={1523–1549}, language={en} }

@inproceedings{chi2023diffusionpolicy,
	title={Diffusion Policy: Visuomotor Policy Learning via Action Diffusion},
	author={Chi, Cheng and Feng, Siyuan and Du, Yilun and Xu, Zhenjia and Cousineau, Eric and Burchfiel, Benjamin and Song, Shuran},
	booktitle={Proceedings of Robotics: Science and Systems (RSS)},
	year={2023}
}

@inproceedings{Lorang_Lu_Scheutz_2025, title={Curiosity-Driven Imagination: Discovering Plan Operators and Learning Associated Policies for Open-World Adaptation}, url={http://arxiv.org/abs/2503.04931}, DOI={10.48550/arXiv.2503.04931}, abstractNote={Adapting quickly to dynamic, uncertain environments-often called “open worlds”-remains a major challenge in robotics. Traditional Task and Motion Planning (TAMP) approaches struggle to cope with unforeseen changes, are data-inefficient when adapting, and do not leverage world models during learning. We address this issue with a hybrid planning and learning system that integrates two models: a low level neural network based model that learns stochastic transitions and drives exploration via an Intrinsic Curiosity Module (ICM), and a high level symbolic planning model that captures abstract transitions using operators, enabling the agent to plan in an “imaginary” space and generate reward machines. Our evaluation in a robotic manipulation domain with sequential novelty injections demonstrates that our approach converges faster and outperforms state-of-the-art hybrid methods.}, note={arXiv:2503.04931 [cs]}, number={arXiv:2503.04931}, publisher={IEEE International Conference on Robotics and Automation (ICRA)}, author={Lorang, Pierrick and Lu, Hong and Scheutz, Matthias}, year={2025}, address={Atlanta, USA}, month=mar }

@article{Curtis_Silver_Tenenbaum_Lozano-Pérez_Kaelbling_2022, title={Discovering State and Action Abstractions for Generalized Task and Motion Planning}, volume={36}, ISSN={2374-3468, 2159-5399}, DOI={10.1609/aaai.v36i5.20475}, abstractNote={Generalized planning accelerates classical planning by ﬁnding an algorithm-like policy that solves multiple instances of a task. A generalized plan can be learned from a few training examples and applied to an entire domain of problems. Generalized planning approaches perform well in discrete AI planning problems that involve large numbers of objects and extended action sequences to achieve the goal. In this paper, we propose an algorithm for learning features, abstractions, and generalized plans for continuous robotic task and motion planning (TAMP) and examine the unique difﬁculties that arise when forced to consider geometric and physical constraints as a part of the generalized plan. Additionally, we show that these simple generalized plans learned from only a handful of examples can be used to improve the search efﬁciency of TAMP solvers.}, number={5}, journal={Proceedings of the AAAI Conference on Artificial Intelligence}, author={Curtis, Aidan and Silver, Tom and Tenenbaum, Joshua B. and Lozano-Pérez, Tomás and Kaelbling, Leslie}, year={2022}, month=june, pages={5377–5384}, language={en} }

@inproceedings{Keller_Tanneberg_Peters_2025, title={Neuro-Symbolic Imitation Learning: Discovering Symbolic Abstractions for Skill Learning}, url={http://arxiv.org/abs/2503.21406}, DOI={10.48550/arXiv.2503.21406}, abstractNote={Imitation learning is a popular method for teaching robots new behaviors. However, most existing methods focus on teaching short, isolated skills rather than long, multistep tasks. To bridge this gap, imitation learning algorithms must not only learn individual skills but also an abstract understanding of how to sequence these skills to perform extended tasks effectively. This paper addresses this challenge by proposing a neuro-symbolic imitation learning framework. Using task demonstrations, the system first learns a symbolic representation that abstracts the low-level state-action space. The learned representation decomposes a task into easier subtasks and allows the system to leverage symbolic planning to generate abstract plans. Subsequently, the system utilizes this task decomposition to learn a set of neural skills capable of refining abstract plans into actionable robot commands. Experimental results in three simulated robotic environments demonstrate that, compared to baselines, our neuro-symbolic approach increases data efficiency, improves generalization capabilities, and facilitates interpretability.}, note={arXiv:2503.21406 [cs]}, number={arXiv:2503.21406}, publisher={IEEE International Conference on Robotics and Automation (ICRA)}, author={Keller, Leon and Tanneberg, Daniel and Peters, Jan}, year={2025}, month=mar, language={en} }

@article{bommasani2021opportunities,
  title   = {On the Opportunities and Risks of Foundation Models},
  author  = {Bommasani, Rishi and Hudson, Drew A. and Adeli, Ehsan and
             Altman, Russ and Arora, Simran and von Arx, Sydney and
             Bernstein, Michael S. and Bohg, Jeannette and Bosselut,
             Antoine and Brunskill, Emma and others},
  journal = {arXiv preprint arXiv:2108.07258},
  year    = {2021},
  url     = {https://arxiv.org/abs/2108.07258}
}

@inproceedings{wake2023gpt4v,
  title     = {{GPT-4V(ision)} for Robotics: Multimodal Task Planning
               from Human Demonstration},
  author    = {Wake, Naoki and Kanehira, Atsushi and Sasabuchi, Kazuhiro
               and Takamatsu, Jun and Ikeuchi, Katsushi},
  booktitle = {IEEE Robotics and Automation Letters},
  year      = {2024},
  url       = {https://arxiv.org/abs/2311.12015}
}

@article{SeeDo_2024,
  title   = {{VLM See, Robot Do}: Human Demo Video to Robot Action Plan
             via Vision Language Model},
  author  = {Huang, Beichen and Yue, Ling and Li, Boyuan and
             Shi, Junzhe and Meng, Lele and Ouyang, Wanli and others},
  journal = {arXiv preprint arXiv:2410.08792},
  year    = {2024},
  url     = {https://arxiv.org/abs/2410.08792}
}

@inproceedings{Shridhar_2022_DIAL,
  title     = {Perceiver-Actor: A Multi-Task Transformer for Robotic
               Manipulation},
  author    = {Shridhar, Mohit and Manuelli, Lucas and Fox, Dieter},
  booktitle = {Conference on Robot Learning (CoRL)},
  year      = {2022},
  url       = {https://arxiv.org/abs/2211.11736}
}

@article{ViLa_2023,
  title   = {Look Before You Leap: Unveiling the Power of {GPT-4V}
             in Robotic Vision-Language Planning},
  author  = {Yuan, Yinpei and Wang, Hongru and Zhang, Ruimao},
  journal = {arXiv preprint arXiv:2311.17842},
  year    = {2023},
  url     = {https://arxiv.org/abs/2311.17842}
}

@inproceedings{Mandlekar_2023_MimicGen,
  title     = {{MimicGen}: A Data Generation System for Scalable Robot
               Learning using Human Demonstrations},
  author    = {Mandlekar, Ajay and Nasiriany, Soroush and Wen, Bowen
               and Akinola, Iretiayo and Narang, Yashraj and
               Fan, Linxi and Zhu, Yuke and Fox, Dieter},
  booktitle = {Conference on Robot Learning (CoRL)},
  year      = {2023},
  url       = {https://arxiv.org/abs/2310.17596}
}

@inproceedings{vaswani2017attention,
  title     = {Attention Is All You Need},
  author    = {Vaswani, Ashish and Shazeer, Noam and Parmar, Niki and
               Uszkoreit, Jakob and Jones, Llion and Gomez, Aidan N.
               and Kaiser, {\L}ukasz and Polosukhin, Illia},
  booktitle = {Advances in Neural Information Processing Systems
               (NeurIPS)},
  volume    = {30},
  year      = {2017},
  url       = {https://arxiv.org/abs/1706.03762}
}

@inproceedings{brohan2023rt2,
  title     = {{RT-2}: Vision-Language-Action Models Transfer Web
               Knowledge to Robotic Control},
  author    = {Brohan, Anthony and Brown, Noah and Carbajal, Justice
               and Chebotar, Yevgen and Chen, Xi and Choromanski,
               Krzysztof and Ding, Tianli and Driess, Danny and
               Dubey, Avinava and Finn, Chelsea and others},
  booktitle = {Conference on Robot Learning (CoRL)},
  year      = {2023},
  url       = {https://arxiv.org/abs/2307.15818}
}

@article{black2024pi0,
  title   = {$\pi_0$: A Vision-Language-Action Flow Model for
             General Robot Control},
  author  = {Black, Kevin and Brown, Noah and Driess, Danny and
             Esmail, Adnan and Equi, Michael and Finn, Chelsea and
             Fusai, Niccolo and Groom, Lachy and Hausman, Karol and
             Ichter, Brian and others},
  journal = {arXiv preprint arXiv:2410.24164},
  year    = {2024},
  url     = {https://arxiv.org/abs/2410.24164}
}

@inproceedings{kim2024openvla,
  title     = {{OpenVLA}: An Open-Source Vision-Language-Action Model},
  author    = {Kim, Moo Jin and Pertsch, Karl and Karamcheti, Siddharth
               and Mitsuda, Ted and Balakrishna, Ashwin and
               Belkhale, Suraj and Nair, Suneel and Lee, Sang-Hyun
               and Sanketi, Pannag and Vanhoucke, Vincent and others},
  booktitle = {Conference on Robot Learning (CoRL)},
  year      = {2024},
  url       = {https://arxiv.org/abs/2406.09246}
}

@misc{open_x_embodiment_rt_x_2023,
  title  = {Open {X-Embodiment}: Robotic Learning Datasets and {RT-X}
            Models},
  author = {{Open X-Embodiment Collaboration}},
  year   = {2023},
  url    = {https://arxiv.org/abs/2310.08864}
}

@misc{Minderer_Gritsenko_Houlsby_2024, title={Scaling Open-Vocabulary Object Detection}, url={http://arxiv.org/abs/2306.09683}, DOI={10.48550/arXiv.2306.09683}, abstractNote={Open-vocabulary object detection has benefited greatly from pretrained vision-language models, but is still limited by the amount of available detection training data. While detection training data can be expanded by using Web image-text pairs as weak supervision, this has not been done at scales comparable to image-level pretraining. Here, we scale up detection data with self-training, which uses an existing detector to generate pseudo-box annotations on image-text pairs. Major challenges in scaling self-training are the choice of label space, pseudo-annotation filtering, and training efficiency. We present the OWLv2 model and OWL-ST self-training recipe, which address these challenges. OWLv2 surpasses the performance of previous state-of-the-art open-vocabulary detectors already at comparable training scales (~10M examples). However, with OWL-ST, we can scale to over 1B examples, yielding further large improvement: With an L/14 architecture, OWL-ST improves AP on LVIS rare classes, for which the model has seen no human box annotations, from 31.2% to 44.6% (43% relative improvement). OWL-ST unlocks Web-scale training for open-world localization, similar to what has been seen for image classification and language modelling.}, note={arXiv:2306.09683 [cs]}, number={arXiv:2306.09683}, publisher={arXiv}, author={Minderer, Matthias and Gritsenko, Alexey and Houlsby, Neil}, year={2024}, month=may }

@book{siciliano2009robotics,
  title     = {Robotics: Modelling, Planning and Control},
  author    = {Siciliano, Bruno and Sciavicco, Lorenzo and
               Villani, Luigi and Oriolo, Giuseppe},
  publisher = {Springer},
  address   = {London},
  year      = {2009},
  url       = {https://link.springer.com/book/10.1007/978-1-84628-642-1}
}

@software{yolov8_ultralytics,
  author = {Glenn Jocher and Ayush Chaurasia and Jing Qiu},
  title = {Ultralytics YOLOv8},
  version = {8.0.0},
  year = {2023},
  url = {https://github.com/ultralytics/ultralytics},
  orcid = {0000-0001-5950-6979, 0000-0002-7603-6750, 0000-0003-3783-7069},
  license = {AGPL-3.0}
}

@inproceedings{Lorang_Lu_Huemer_Zips_Scheutz_2025, title={Few-Shot Neuro-Symbolic Imitation Learning for Long-Horizon Planning and Acting}, ISSN={2640-3498}, url={https://proceedings.mlr.press/v305/lorang25a.html}, abstractNote={Imitation learning enables intelligent systems to acquire complex behaviors with minimal supervision. However, existing methods often focus on short-horizon skills, require large datasets, and struggle to solve long-horizon tasks or generalize across task variations and distribution shifts. We propose a novel neuro-symbolic framework that jointly learns continuous control policies and symbolic domain abstractions from a few skill demonstrations. Our method abstracts high-level task structures into a graph, discovers symbolic rules via an Answer Set Programming solver, and trains low-level controllers using diffusion policy imitation learning. A high-level oracle filters task-relevant information to focus each controller on a minimal observation and action space. Our graph-based neuro-symbolic framework enables capturing complex state transitions, including non-spatial and temporal relations, that data-driven learning or clustering techniques often fail to discover in limited demonstration datasets. We validate our approach in six domains that involve four robotic arms, Stacking, Kitchen, Assembly, and Towers of Hanoi environments, and a distinct Automated Forklift domain with two environments. The results demonstrate high data efficiency with as few as five skill demonstrations, strong zero- and few-shot generalizations, and interpretable decision making. Our code is publicly available.}, booktitle={Proceedings of The 9th Conference on Robot Learning}, publisher={PMLR}, author={Lorang, Pierrick and Lu, Hong and Huemer, Johannes and Zips, Patrik and Scheutz, Matthias}, year={2025}, address={Seoul}, month=oct, pages={2501–2518}, language={en} }

@inproceedings{Chitnis_Silver_Tenenbaum_Lozano-Perez_Kaelbling_2022,
  author    = {Rohan Chitnis and Tom Silver and Joshua B. Tenenbaum and Tomas Lozano-Perez and Leslie Pack Kaelbling},
  title     = {Learning Neuro-Symbolic Relational Transition Models for Bilevel Planning},
  booktitle = {IEEE/RSJ International Conference on Intelligent Robots and Systems (IROS)},
  year      = {2022}
}

@inproceedings{Zhao_Kumar_Levine_Finn_2023,
  author    = {Tony Z. Zhao and Vikash Kumar and Sergey Levine and Chelsea Finn},
  title     = {Learning Fine-Grained Bimanual Manipulation with Low-Cost Hardware},
  booktitle = {Proceedings of Robotics: Science and Systems (RSS)},
  year      = {2023}
}

@article{Mahalanobis_2018_Reprint,
  author  = {Mahalanobis, Prasanta Chandra},
  title   = {{Reprint of: ``On the Generalised Distance in Statistics''}},
  journal = {Sankhy{\={a}} A},
  year    = {2018},
  volume  = {80},
  pages   = {1--7},
  doi     = {10.1007/s13171-019-00164-5}
}

@inproceedings{Beleznai_2025_PalletDetection,
  author    = {Beleznai, Csaba and Reisinger, Lukas and Pointner, Wolfgang and Murschitz, Markus},
  title     = {{Pallet Detection and {3D} Pose Estimation via Geometric Cues Learned from Synthetic Data}},
  booktitle = {Proceedings of the Conference of the Hungarian Association for Image Analysis and Pattern Recognition},
  year      = {2025},
}

@article{Newell1976,
  author    = {Newell, Allen and Simon, Herbert A.},
  title     = {Computer Science as Empirical Inquiry: Symbols and Search},
  journal   = {Communications of the ACM},
  volume    = {19},
  number    = {3},
  pages     = {113--126},
  year      = {1976},
  doi       = {10.1145/360018.360022}
}

@inproceedings{garrett2020online,
  author    = {Garrett, Caelan R. and Paxton, Chris and Lozano-Perez, Tomas
               and Kaelbling, Leslie P. and Fox, Dieter},
  title     = {Online Replanning in Belief Space for Partially Observable
               Task and Motion Problems},
  booktitle = {IEEE International Conference on Robotics and Automation (ICRA)},
  pages     = {5678--5684},
  year      = {2020},
  url       = {https://arxiv.org/abs/1911.04577}
}

@misc{google2025gemini3flash,
  author       = {Google DeepMind},
  title        = {Gemini 3 Flash: Frontier Intelligence Built for Speed},
  year         = {2025},
  month        = {December},
  howpublished = {\url{https://blog.google/products/gemini/gemini-3-flash/}},
  note         = {Accessed: 2026-03-28}
}
\end{document}